\documentclass[numbers]{assets/matterlab}

\usepackage{enumitem}
\usepackage{titletoc}
\usepackage[dvipsnames]{xcolor}
\usepackage[nolist]{acronym}
\usepackage{cinzel}

\usepackage{microtype}
\usepackage{graphicx}
\usepackage{booktabs}
\usepackage{float}
\usepackage{xurl}
\usepackage{hyperref}
\usepackage{titlesec}

\usepackage{amsmath}
\usepackage{amssymb}
\usepackage{mathtools}
\usepackage{amsthm}
\usepackage{bm}

\usepackage{amsmath,amsfonts,bm}

\def\1{\bm{1}}

\DeclareMathAlphabet{\mathsfit}{\encodingdefault}{\sfdefault}{m}{sl}
\SetMathAlphabet{\mathsfit}{bold}{\encodingdefault}{\sfdefault}{bx}{n}

\usepackage{orcidlink}
\usepackage{comment}
\usepackage[version=4]{mhchem}
\usepackage{longtable}
\usepackage{array}
\renewcommand{\arraystretch}{1.7}
\usepackage{multirow}           %
\usepackage{amsfonts}           %
\usepackage{nicefrac}
\usepackage{duckuments}         %
\usepackage{enumitem}
\usepackage{xcolor,colortbl}
\usepackage{tikz}
\usepackage{tikz-cd}
\usepackage{caption}
\usepackage{subcaption}
\usepackage{listings}
\usepackage{gensymb}
\usepackage{textcomp} %
\usepackage[inkscapelatex=false]{svg}

\usepackage[noabbrev,nameinlink]{cleveref}

\newcommand{\addressCHEM}{Department of Chemistry, University of Toronto, Lash Miller Chemical Laboratories, 80 St. George Street, ON M5S 3H6, Toronto, Canada}
\newcommand{\addressAC}{Acceleration Consortium, 700 University Ave., M7A 2S4, Toronto, Canada}
\newcommand{\addressCS}{Department of Computer Science, University of Toronto, Sandford Fleming Building, 10 King’s College Road, ON M5S 3G4, Toronto, Canada}
\newcommand{\addressVECTOR}{Vector Institute for Artificial Intelligence, 661 University Ave. Suite 710, ON M5G 1M1, Toronto, Canada}
\newcommand{\addressMSE}{Department of Materials Science \& Engineering, University of Toronto, 184 College St., M5S 3E4, Toronto, Canada}
\newcommand{\addressCHEMENG}{Department of Chemical Engineering \& Applied Chemistry, University of Toronto, 200 College St. ON M5S 3E5, Toronto, Canada}
\newcommand{\addressMEDICALSCI}{Institute of Medical Science, 1 King's College Circle, Medical Sciences Building, Room 2374, Toronto, ON M5S 1A8, Canada}
\newcommand{\addressCIFAR}{Senior Fellow, Canadian Institute for Advanced Research (CIFAR), 661 University Ave., M5G 1M1, Toronto, Canada}
\newcommand{\addressNVIDIA}{NVIDIA, 431 King St W \#6th, M5V 1K4, Toronto, Canada}

\newcommand{\acknowAC}{This research is part of the University of Toronto’s Acceleration Consortium, which receives funding from the CFREF-2022-00042 Canada First Research Excellence Fund.}

\providecommand{\doi}[1]{\href{https://doi.org/#1}{doi:#1}}

\crefname{section}{Sec.}{Secs.}
\crefname{figure}{Fig.}{Figs.}
\crefname{table}{Tab.}{Tabs.}
\crefname{equation}{Eq.}{Eqs.}
\Crefname{section}{Section}{Sections}
\Crefname{figure}{Figure}{Figures}
\Crefname{table}{Table}{Tables}
\Crefname{equation}{Equation}{Equations}

\makeatletter
\patchcmd{\AC@verridelabel}
  {\ltx@label{#1}}{\cref@old@label{#1}}
  {}{\GenericError{}{Could not patch acronym cross-reference labels}{}{}}
\makeatother

\newcommand{\farm}{{\cinzel F\'armaco}}
\newcommand{\farmaco}{{\cinzel El Agente F\'armaco}}
\newcommand{\farmacoatlas}{{\cinzel F\'armaco Atlas}}
\newcommand{\elagente}{{\cinzel El Agente}}

\newcommand{\elagenteG}{{\cinzel El Agente Gr\'afico}}

\renewcommand\affiliationformat[2][]{{$^{#1}$#2}}
\renewcommand\contributionformat[2][]{{$^{#1}$#2}}
\patchcmd{\mymaketitle}
  {\affiliationlist\par}
  {{\footnotesize\affiliationlist\par}}
  {}{\GenericError{}{Could not patch affiliation list formatting}{}{}}
\patchcmd{\mymaketitle}
  {\contributionlist\par}
  {{\footnotesize\contributionlist\par}}
  {}{\GenericError{}{Could not patch contribution list formatting}{}{}}

\title{El Agente Fármaco: A Scalable AI Scientist for Drug Discovery}
\author[1,6]{Marcel Müller}
\author[1,2,6]{Mohammad Ghazi Vakili}
\author[1]{Yoshihiro Ogura}
\author[1]{Ella Ashoori}
\author[2]{Shimeng Wang}
\author[1,2,10]{Alessandro S. Nascimento}
\author[11]{Eddie Calleja}
\author[11]{Yutong Zhao}
\author[2,7]{Yimeng Liu}
\author[1,6]{Ella Miray Rajaonson}
\author[2]{Aiwei Yin}
\author[1,6]{Jiaru Bai}
\author[1,6]{Yunheng Zou}
\author[11]{Abraham C. Stern}
\author[1,2,7,*]{Varinia Bernales}
\author[1,2,3,4,5,6,7,8,9,*]{Al\'an Aspuru-Guzik}

\affiliation[1]{\addressCHEM}
\affiliation[2]{\addressCS}
\affiliation[3]{\addressMSE}
\affiliation[4]{\addressCHEMENG}
\affiliation[5]{\addressMEDICALSCI}
\affiliation[6]{\addressVECTOR}
\affiliation[7]{\addressAC}
\affiliation[8]{\addressCIFAR}
\affiliation[9]{\addressNVIDIA}
\affiliation[10]{São Carlos Institute of Physics, University of São Paulo, São Carlos, Brazil}
\affiliation[11]{NVIDIA, 2788 San Tomas Expressway, Santa Clara, CA 95051, USA}

\abstract{
Computational drug discovery offers a growing range of tools for proposing and evaluating molecules, but combining them and deciding what to do next still requires substantial expert effort.
Here, we introduce \farmaco{}, an AI scientist that uses modern \ac{llm} technology to turn discovery goals expressed in everyday language into complex, scalable computational screening workflows.
The agentic harness selects scientific tools, reviews their results, and adjusts subsequent steps to the target and the user's objectives.
Our approach centers on a shared scientific memory: \farmacoatlas{}, which stores molecules, structures, and results together with their provenance---a record of how they were obtained.
The agent uses targeted database queries to retrieve the evidence needed for each decision.
A single request can launch calculations for an entire collection of molecules, without requiring a separate \ac{llm} decision for every compound.
This design limits the information the model must process as screening collections grow, while keeping the underlying results available for inspection and further analysis.
The integrated tools support molecule generation, prediction of binding and molecular properties, and searches for compounds available for purchase.
In a retrospective comparison with an expert-led campaign, the agent proposed chemically distinct candidates with competitive docking scores and higher calculated drug-likeness under explicitly imposed selection criteria.
In a 22-task benchmark against plain tool calls and Codex, the agent achieved the highest observed mean model-judged score, with lower median cumulative token use and peak context.
\farm{} provides a foundation for flexible, scalable drug discovery workflows that connect scientific reasoning with an enduring, traceable record of computational work.

}

\date{\today}
\correspondence{\email{varinia@bernales.org} and \email{alan@aspuru.com}}

\begin{document}

\acrodef{si}[SI]{Supporting Information}
\acrodef{llm}[LLM]{large language model}
\acrodef{ai}[AI]{artificial intelligence}
\acrodef{ml}[ML]{machine learning}
\acrodef{bo}[BO]{Bayesian optimization}
\acrodef{al}[AL]{active learning}
\acrodef{mcp}[MCP]{Model Context Protocol}
\acrodef{dora}[DORA]{Draft Outline Research Assistant}
\acrodef{msa}[MSA]{multiple sequence alignment}

\acrodef{qc}[QC]{quantum chemistry}
\acrodef{mof}[MOF]{metal-organic framework}
\acrodef{pdb}[PDB]{Protein Data Bank}
\acrodef{spr}[SPR]{surface plasmon resonance}
\acrodef{sdf}[SDF]{structure data format}
\acrodef{mmcif}[mmCIF]{macromolecular crystallographic information file}
\acrodef{smiles}[SMILES]{simplified molecular-input line-entry system}
\acrodef{admet}[ADMET]{absorption, distribution, metabolism, excretion, and toxicity}
\acrodef{dili}[DILI]{drug-induced liver injury}
\acrodef{fep}[FEP]{free-energy perturbation}
\acrodef{rbfe}[RBFE]{relative binding free energy}
\acrodef{abfe}[ABFE]{absolute binding free energy}
\acrodef{lomap}[LOMAP]{Lead Optimization Mapper}
\acrodef{plip}[PLIP]{Protein--Ligand Interaction Profiler}
\acrodef{tps}[TPS]{thiamine phosphate}
\acrodef{atp}[ATP]{adenosine triphosphate}
\acrodef{qed}[QED]{quantitative estimate of drug-likeness}
\acrodef{pains}[PAINS]{pan-assay interference compounds}
\acrodef{smarts}[SMARTS]{SMILES arbitrary target specification}
\acrodef{plddt}[pLDDT]{predicted local-distance difference test}
\acrodef{rmsd}[RMSD]{root-mean-square deviation}
\acrodef{mcs}[MCS]{maximum common substructure}
\acrodef{sa}[SA]{synthetic accessibility}
\acrodef{tpsa}[TPSA]{topological polar surface area}
\acrodef{hbd}[HBD]{hydrogen-bond donors}
\acrodef{hba}[HBA]{hydrogen-bond acceptors}
\acrodef{iptm}[ipTM]{interface predicted template modeling score}
\acrodef{nih}[NIH]{National Institutes of Health}
\acrodef{darpa}[DARPA]{Defense Advanced Research Projects Agency}

\acrodef{nim}[NIM]{NVIDIA inference microservice}
\acrodef{api}[API]{application programming interface}
\acrodef{cli}[CLI]{command-line interface}
\acrodef{io}[I/O]{input/output}
\acrodef{sql}[SQL]{structured query language}
\acrodef{http}[HTTP]{Hypertext Transfer Protocol}
\acrodef{json}[JSON]{JavaScript Object Notation}
\acrodef{csv}[CSV]{comma-separated values}
\acrodef{rcsb}[RCSB]{Research Collaboratory for Structural Bioinformatics}
\acrodef{3d}[3D]{three-dimensional}
\acrodef{gpu}[GPU]{graphics processing unit}
\acrodef{utc}[UTC]{Coordinated Universal Time}

\maketitle
\acresetall

\section{Introduction}

Early-stage drug discovery involves identifying and evaluating candidate therapeutics, including small molecules, peptides, and proteins \cite{hughesPrinciplesEarlyDrug2011}, as well as other modalities such as messenger RNA (mRNA) vaccines.
Over the past decade, the computational toolkit for drug discovery has broadened considerably.
Established approaches such as molecular docking and binding free energy calculations \cite{paggiArtScienceMolecular2024a,leeAlchemicalBindingFree2020,courniaRelativeBindingFree2017,baumannLargescaleCollaborativeAssessment2025} have been joined by a growing range of data-driven tools, including AlphaFold for protein structure prediction \cite{jumperHighlyAccurateProtein2021} and Boltz-2 for predicting protein--ligand complexes and binding affinities \cite{passaroBoltz2AccurateEfficient2025,papidochaDiscoveryNanomolarCDK2}.
The growing range of available tools creates an opportunity to combine complementary evidence when proposing and prioritizing compounds for experimental testing.
However, advances in individual methods do not automatically translate into more efficient drug development: declining pharmaceutical research productivity \cite{scannellDiagnosingDeclinePharmaceutical2012} and high clinical attrition \cite{sunWhy90Clinical2022} highlight the continuing difficulty of selecting successful candidates.

Integrating computational tools into discovery workflows requires deciding which calculations to perform, how to interpret their outputs, and when to revise a candidate set.
Medicinal chemistry optimization requires balancing multiple molecular properties through successive hit-to-lead and lead-optimization cycles \cite{raczChangingLandscapeMedicinal2025}.
Screening funnels organize this process through successive selection and filtering steps \cite{mayrNovelTrendsHighthroughput2009}.
In a predefined pipeline, the order of calculations and filtering steps is specified in advance; changing that sequence in response to intermediate results requires explicitly implemented decision logic or manual intervention.
The challenge is to combine screening efficiency with the ability to revise the workflow around the target, available evidence, and discovery objectives.

\Ac{ai} scientists are usually {\sl harnesses} that package a complex array of scientific tools alongside modern \acp{llm} to achieve scientific goals.
These systems (harnesses) are often built on agentic architectures that are familiar to everyday users.
A great example of a ``viral'' agent is \href{https://openclaw.ai/}{OpenClaw}, which attracted over 100,000 GitHub stars within approximately two months of its creation \cite{steinbergerIntroducingOpenClaw2026}.
OpenClaw supports tasks such as communicating with users through Telegram and managing their email and schedules.
In parallel, coding agents such as \href{https://openai.com/codex/}{Codex}, \href{https://claude.com/product/claude-code}{Claude Code}, or \href{https://pi.dev/}{Pi} have taken the world by storm, enabling programmers to develop complex software \cite{barbasteHarnessEngineeringAnatomy2026}.
In a similar fashion, \ac{ai} scientists are agentic systems that are specialized to carry out scientific tasks instead (and often alongside) coding \cite{fengAgenticAISystems2026}.

\Ac{ai} agentic systems offer an opportunity to make these workflows more adaptive by connecting scientific objectives expressed in natural language to the selection and execution of computational tools.
An agent can use intermediate results to choose subsequent calculations, revise candidate selections, and retrieve additional evidence as a discovery task develops.
This exploits the \ac{llm}'s stochastic inference as a centrepiece for flexible coordination, with specialized models and simulation tools performing the underlying scientific calculations.
Previous work in our group has explored this approach in quantum chemistry \cite{zouAgenteAutonomousAgent2025,perez-sanchezAgenteQunturResearch2026a}, quantum simulations \cite{gustinAgenteCuanticoAutomating2026a}, solid-state simulations \cite{kumarAgenteSolidoNew2026}, atomistic simulations driven by machine-learning interatomic potentials \cite{koAgentePotenteHighThroughput2026}, and molecular structure editing \cite{choiAgenteEstructuralArtificially2026}.
Related efforts have addressed structured scientific execution \cite{baiAgenteGraficoStructured2026}, task-driven generation and reuse of scientific tools \cite{zhangAgenteForjadorTaskDriven2026}, laboratory safety planning and evaluation \cite{kangAgenteSeguroAgent}, and the orchestration of optimization campaigns in self-driving laboratories \cite{mullerAgenteOptimaAgentic2026}.

In drug discovery and related fields, existing systems apply this approach at different levels, from assisting interactions with individual scientific tools to coordinating multiple stages of a discovery workflow.
PyMolAI \cite{ravishar313Ravishar313PyMolAI2026} integrates an agentic assistant into PyMOL for molecular visualization and tool execution.
BioChemAIgent \cite{yousefiBioChemAIgentAIdrivenProtein2025} extends the scope to structure-based drug discovery through a chat interface connected to tools for molecular analysis, protein modeling, and docking.
AgentD \cite{ockLargeLanguageModel2026} organizes biomedical retrieval, molecular generation, property prediction, iterative refinement, and protein--ligand structure generation into a staged pipeline.\footnote{
      Both BioChemAIgent and AgentD expose capabilities through \ac{mcp} interfaces.
      BioChemAIgent's \href{https://github.com/imsb-uke/bcai/blob/main/mcp/server.py}{tool registry} includes library preparation, screening, and workflow storage and execution.
      AgentD's \href{https://github.com/hoon-ock/AgentD\#pipeline-configuration}{documented pipeline} uses configurable refinement iterations and candidate-selection thresholds, saves results in run directories, and optionally generates complexes with Boltz-2.
}
Related scientific agents illustrate the breadth of capabilities that can be coordinated in this way: MatSciAgent routes natural-language requests to specialized agents for data retrieval, structure generation, and physical simulation \cite{chaudhariModularLargeLanguage2026}, while TxAgent selects among 211 biomedical tools and iteratively retrieves evidence during therapeutic reasoning \cite{gaoTxAgentAI2025}.
Related commercial platforms and industry initiatives extend agentic approaches across molecular design, scientific analysis, and research support.
\href{https://boltz.bio/boltzlab}{Boltz Lab} provides agents for small-molecule hit discovery and protein design, while \href{https://www.anthropic.com/news/claude-science-ai-workbench}{Claude Science} provides a broader scientific workbench connecting agents to research tools, data sources, and computing resources, including \href{https://boltz.bio/boltz-claude}{access to the Boltz \ac{api}}.
Insilico Medicine's \href{https://insilico.com/news/kv5e80yc41-insilico-open-sources-dora-on-github-sup}{\ac{dora}} supports literature synthesis and the drafting of scientific documents.
At the level of research infrastructure, the \href{https://nvidianews.nvidia.com/news/nvidia-and-lilly-announce-co-innovation-lab-to-reinvent-drug-discovery-in-the-age-of-ai}{Lilly--NVIDIA co-innovation initiative} aims to connect computational modeling with automated experimentation in a continuous learning system.

Across these approaches, connecting tools raises a shared architectural question: how should scientific entities, results, and provenance be represented and accessed as workflows grow or change?
An agent needs to retain the relationships between candidate compounds, calculations, and observations across successive decisions, while managing tool descriptions and retrieved evidence within a finite context window.
For example, a complete \ac{pdb} file can occupy substantial context even when the agent needs only a summary of protein--ligand interactions or a comparison between experimental and predicted structures.
Separating persistent scientific state from the \ac{llm}'s working context allows detailed records to remain available while the agent retrieves the evidence needed for the current decision.

\begin{figure}
      \centering
      \includegraphics[width=\linewidth,height=0.80\textheight,keepaspectratio]{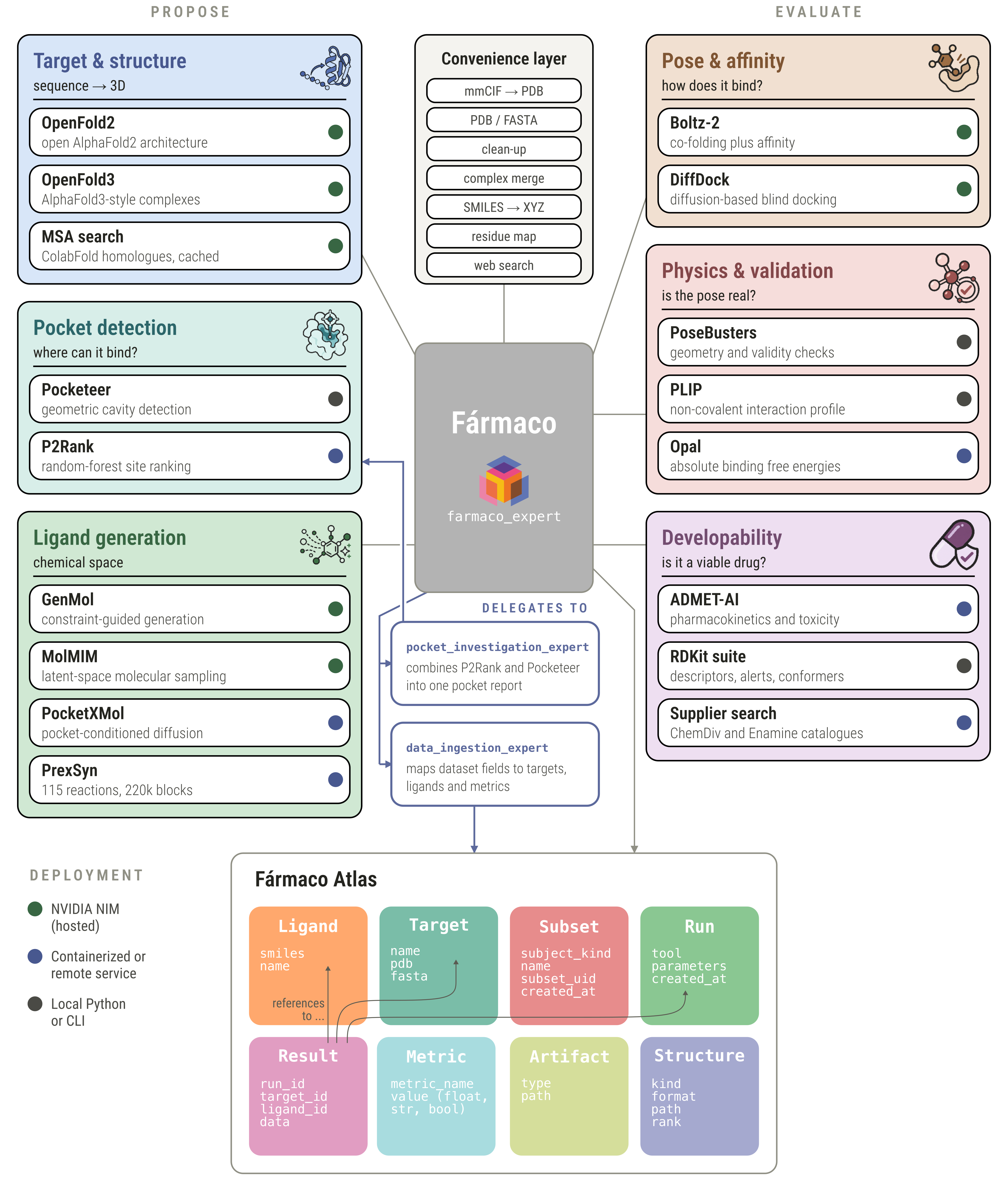}
      \caption{
            Capability overview of \farm{}.
            \farm{} (\textit{centre}) coordinates workflows and delegates pocket investigation and dataset ingestion to specialist agents.
            Subset-based workflows select inputs through \acs*{sql} queries, dispatch tool calls, and return run summaries.
            Its capabilities include format conversion and retrieval utilities (\textit{above}), six categories of scientific tools (\textit{flanking}), and persistent storage in \farmacoatlas{} (\textit{below}), shown with selected fields from its core records.
            The left and right columns group tools by their roles in proposal and evaluation, respectively.
            Marker colour indicates the deployment modes shown: hosted \ac{nim}, containerized or remote services, and local Python or \acs*{cli} tools.
      }
      \label{fig:agent_network}
\end{figure}

Here, we present \farmaco{}, an agentic system that combines drug-discovery capabilities spanning ligand generation to free energy calculation with \farmacoatlas{}, a persistent database of scientific entities and workflow provenance (\cref{fig:agent_network}).
Scientific tools operate on database subsets, and the agent receives compact run summaries from which it can request detailed results as needed.
This design connects workflow planning, execution, and subsequent analysis through a shared scientific state, while selectively exposing data and tool parameters to the \ac{llm}.
We describe the database and tool-execution architecture and evaluate the system through a 22-task benchmark and five computational case studies.
These cover retrospective ligand prioritization, adaptive and prospective candidate selection, binding-pose assessment, and relative binding free energy calculations and analogue design.

\section{A drug design-native database accessible to an agent}

At the center of \farmaco{} is the \farmacoatlas{}, a SQLModel-based \cite{ramirez_sqlmodel_2026} persistent database that stores all scientific entities, results, artifacts, and workflow state in a unified relational schema.

This design builds on earlier approaches to storing scientific state outside the conversation.
\elagente{} persisted information in files \cite{zouAgenteAutonomousAgent2025}, while \elagenteG{} used a knowledge-graph-backed Python state for structured access during agent-steered workflows \cite{baiAgenteGraficoStructured2026}.
Files can store structured data, but a shared schema, query interface, and explicit provenance relationships must be provided by the surrounding system.
\farmacoatlas{} supplies these through a relational database that persists beyond individual chats and restarts, with large artifacts kept in registered external files.\footnote{
      Complementary memory mechanisms include semantic retrieval of embedded documents or summaries \cite{lewisRetrievalAugmentedGenerationKnowledgeIntensive2021} and event logs recording actions and observations.
      These mechanisms support different access patterns: semantic retrieval helps locate relevant text, while event logs preserve execution history; neither alone defines the relational schema used here for scientific entities and metrics.
}

The core database records comprise:
\begin{itemize}
      \item \textbf{Protein and ligand entities:}
            Protein targets and ligands defined by \ac{smiles} strings define the primary biological and chemical objects represented in the database.

      \item \textbf{Structures:}
            Materialized structures capture concrete three-dimensional objects, including predicted protein structures, docked poses, protein--ligand complexes, conformers, and derived structural models.

      \item \textbf{Workflow provenance:}
            Tool workflow runs record input parameters, detailed results, and generated artifacts, showing which computational action was executed, which inputs it used, and what outputs it produced.

      \item \textbf{Metrics:}
            Normalized metrics store heterogeneous tool outputs as queryable scientific observations, including scalar, categorical, and Boolean values.

      \item \textbf{Subsets:}
            Persistent subsets define reusable collections of ligands, targets, structures, or protein--ligand pairs, allowing agents to carry curated selections across multi-step workflows.
\end{itemize}

\paragraph{Data ingestion.}
User-provided \ac{csv} datasets can enter \farmacoatlas{} through a dedicated \texttt{data\_ingestion\_expert}, which inspects columns and sample rows before mapping selected fields to target and ligand records and typed metrics.
The import records its source filename in a run, creates reusable subsets of targets, ligands, and protein--ligand pairs, and commits the batch transactionally.
Separately, externally supplied protein or ligand structure files can be registered with existing target or ligand records, allowing downstream tools to use experimental or refined structures without requiring a public \ac{pdb} accession.
Mapping and validation details are provided in \cref{sec:si-ingestion}.

\paragraph{Subset-based execution.}
In \farm{}, scientific workflows operate on predefined database subsets (see \cref{fig:database_workflow}).
For example, a workflow screening 200 ligands for \ac{admet} properties receives a single \texttt{subset\_id} referencing the collection, without requiring the agent to include 200 \ac{smiles} strings in the tool call.
The same mechanism applies to targets, structures, and protein--ligand pairs.

\subsection{Progressive disclosure of context and tools}

Progressive disclosure treats context retrieval as a query-planning problem: the agent requests the information needed for the current reasoning step.\footnote{
      Related patterns include incremental retrieval through \href{https://graphql.org/learn/pagination/}{GraphQL cursor connections}; field selection, relationship expansion, filtering, and pagination through \href{https://learn.microsoft.com/en-us/odata/concepts/queryoptions-overview}{OData query options}; and field-restricted previews of top-ranked records within result groups using \href{https://www.elastic.co/docs/reference/aggregations/search-aggregations-metrics-top-hits-aggregation}{Elasticsearch top hits aggregations}.
      For agents, Anthropic describes retaining lightweight identifiers and loading details on demand as a just-in-time context strategy using \href{https://www.anthropic.com/engineering/effective-context-engineering-for-ai-agents}{progressive disclosure}.
}
In \farm{}, this principle governs both access to scientific data and exposure of tool parameters.

\paragraph{Disclosure of scientific data.}
Subset identifiers reference the collections selected for processing, while run summaries identify the stored results and fields available for querying.
Generated metrics, structures, and artifacts are stored in \farmacoatlas{} or in files registered in the database.
The agent retrieves detailed records through targeted \ac{sql} queries and accesses registered artifacts as needed for subsequent reasoning.
The query interface enforces read-only access and bounded row previews, rejects broad \texttt{SELECT *} projections, and reports when results have been truncated.
Oversized previews are shortened before being returned, while file-reading tools likewise limit the text exposed to the agent (\cref{sec:si-execution}).
\farmacoatlas{} retains the persistent scientific state while the working context contains the selected evidence needed for the current task.

\begin{figure}
      \centering
      \includegraphics[width=1\linewidth]{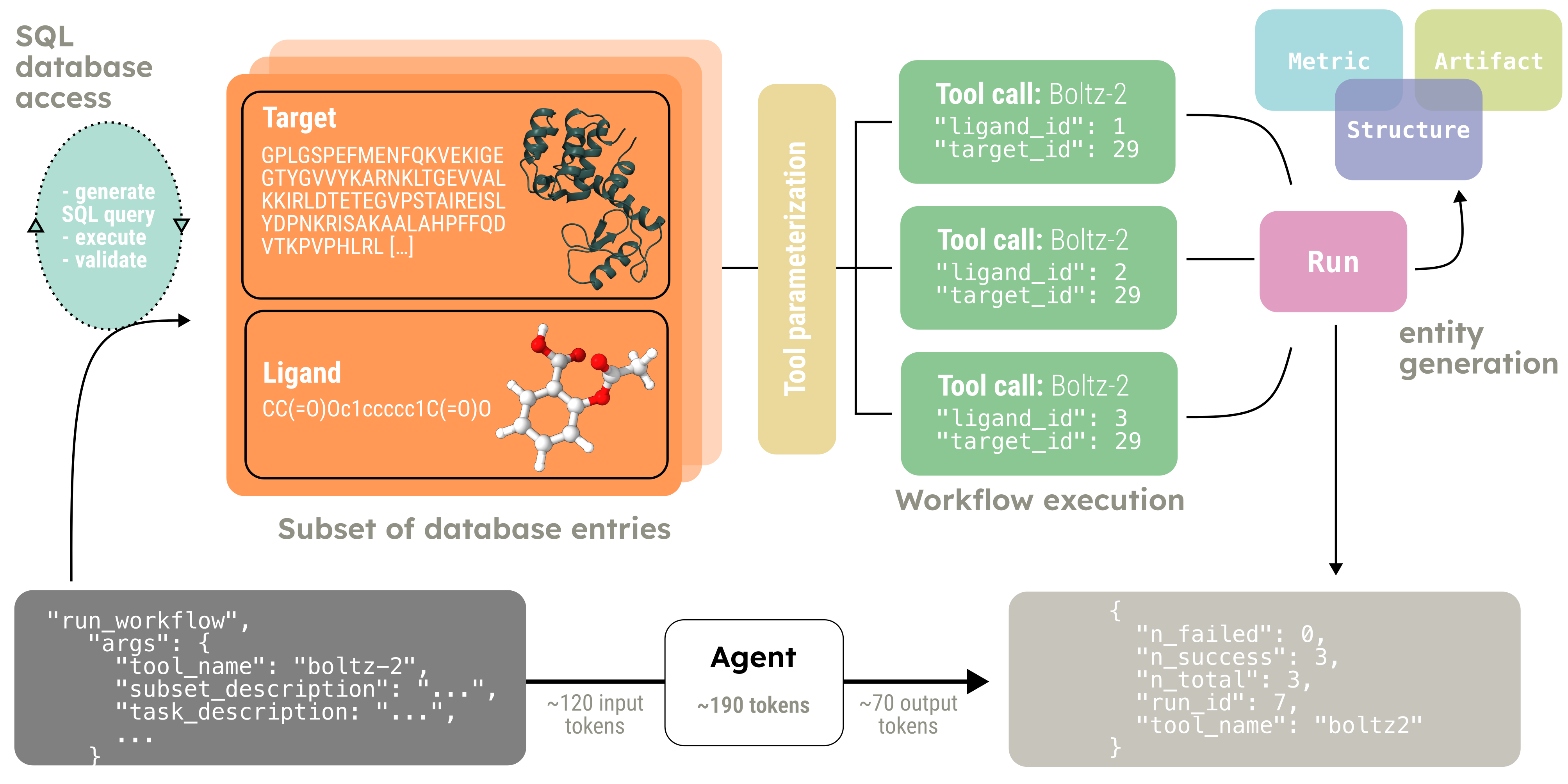}
      \caption{
            Database-driven tool execution.
            The agent requests a tool using high-level task and subset descriptions.
            The workflow translates the subset description into a validated \acs*{sql} query, creates a \texttt{subset\_id}, and configures the selected tool.
            The \acs*{sql} database workflow resolves the subset entity into a list of targets, ligands, or target-ligand pairs (\textit{orange}), sets up job specifications per entity, fans out parallelized workflows (\textit{green}), and summarizes information into a run entity (\textit{pink}).
            The agent only receives a summary of the run; the resulting metrics, structures, or artifacts are stored in the database (\textit{top-right corner}) and are only accessed upon a targeted \acs*{sql} request.
      }
      \label{fig:database_workflow}
\end{figure}

\paragraph{Disclosure of tool parameters.}
\farm{} receives a list of available tools together with high-level descriptions of their capabilities.
Detailed tool parameters are disclosed to a separate configuration node only after a tool has been selected.
Separating tool configuration introduces an additional \ac{llm} call; this design anticipates a substantially larger tool ecosystem than the one currently implemented.
A tool-execution workflow takes only the requested tool name, a high-level description of the task, such as ``\textit{predict aqueous solubilities},'' and a description of the target entities, such as ``\textit{the 50 ligands with the highest binding affinity}.''
If the requested tool requires database entities as input, the first step is a separate subgraph.
This subgraph starts with an \ac{llm}-backed agent node that translates the high-level entity description into an \ac{sql} query command (see \cref{fig:database_workflow,fig:si-database-workflow,fig:si-subset-selection}).
After the \ac{sql} command executes, the workflow verifies that only subset-native entity IDs are returned, then creates a subset entity from the resulting ID collection.
The tool name and high-level task description are then passed to another \ac{llm}-backed agent node.
At this stage, the \ac{llm} receives the relevant calculation settings and fills in run-specific parameters.
The workflow then resolves each subset into the corresponding database entities, for example, 50 ligands, and dispatches their tool calls with bounded concurrency without further \ac{llm} interaction.
After all tool calls have completed, the workflow records their outputs and returns a run summary, completing the data-disclosure cycle described above.
Together, these mechanisms limit the tool specifications and scientific data entering the \ac{llm} context, reducing token usage and keeping the available evidence focused on the scientific decision at hand.
Detailed execution and failure policies are described in \cref{sec:si-execution}.

\subsection{Drug discovery tools}

The integrated tools cover target and structure preparation, pocket detection, ligand generation and comparison, pose and affinity prediction, structural analysis and validation, free energy calculations, developability assessment, and compound sourcing (\cref{fig:agent_network}).
Data-retrieval and format-conversion utilities connect these capabilities.
Individual tool descriptions and integration details are provided in \cref{sec:si-tools}.

Tools are deployed through hosted \acp{api}, containerized services, or local Python and command-line interfaces.
Hosted \href{https://developer.nvidia.com/blog/nvidia-nim-offers-optimized-inference-microservices-for-deploying-ai-models-at-scale/}{\acp{nim}} provide remote model inference, while dedicated Docker containers expose other tools through FastAPI services to isolate dependencies.
Tools with manageable dependencies run in the agent environment, with Python wrappers providing access to command-line programs.
For remote compute jobs, a shared client manages submission, input and output artifacts, waiting, and cancellation (\cref{sec:si-compute}).

\farm{} consists of one main agent and two specialist subagents.
The main agent, \texttt{farmaco\_expert}, serves as the user entry point and coordinates structure prediction, docking, ligand generation, and developability workflows.
It delegates pocket characterization to \texttt{pocket\_investigation\_expert}, which combines complementary pocket-finding methods into a single report, and dataset import to \texttt{data\_ingestion\_expert}, which inspects user-provided datasets and maps selected fields to targets, ligands, and typed metrics.
Broad workflow planning thus remains centralized while these specialized tasks are delegated to the two subagents.
\section{Multi-task benchmark}
\label{sec:benchmark}

We compared \farmaco{} with its database and workflow interface, plain tool calls, and Codex across 22 tasks covering molecular filtering, conformer generation, sequence and structure analysis, ligand generation, binding-pose assessment, and synthesis-route exploration.
Each configuration ran three repetitions per task with the same scientific instructions and supplied input files, using \texttt{gpt-6-sol} throughout.
Each result was assessed in three judge sessions using \texttt{gpt-6.1-sol}, giving 198 agent runs and 594 judge assessments.

\begin{figure}
      \centering
      \includegraphics[width=\linewidth]{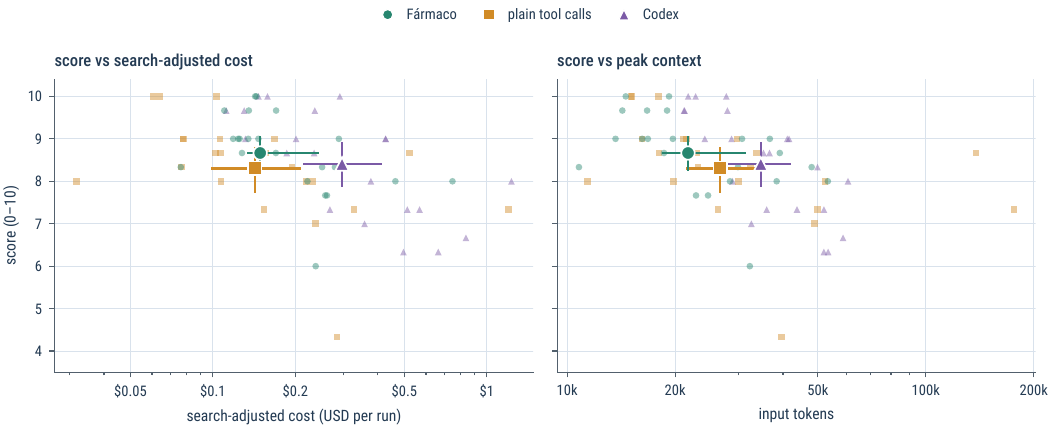}
      \caption{\textbf{Benchmark quality versus model cost and peak context.}
            Small markers show task-level results.
            Large markers summarize each configuration.
            Lines indicate 95\% confidence intervals.
            Metric definitions and statistical methods are provided in \cref{sec:si-benchmark}.}
      \label{fig:benchmark-summary}
\end{figure}

\Cref{fig:benchmark-summary} highlights the balance between judged task quality, context use, and model cost.
\farm{} achieved the highest mean judge score observed.
Its smaller median peak context and lower cumulative token use are consistent with the design goal of keeping detailed scientific state outside the conversation while retrieving selected evidence for each decision.
Its median model cost was lower than Codex.
Plain tool calls had the lowest median total model cost and latency, highlighting that reduced context use does not necessarily yield the cheapest or fastest execution.
Median cached input shares were 90.6\% for plain tool calls and 91.8\% for Codex, compared with 82.2\% for \farm{} (\cref{tab:si-benchmark-summary}).
High cache reuse moderates the cost of repeatedly processing conversation context and helps explain why Codex's substantially higher token use did not lead to a proportional increase in cost.
We selected manageable tasks for repeated execution and assessment, leaving long-running, large-library discovery campaigns outside the benchmark's scope.
This task selection likely underestimates the benefits of subset-based execution for screening thousands or hundreds of thousands of ligands, where per-ligand outputs accumulate in context.
An additional experiment extended the STK17B\footnote{Serine/threonine kinase 17B.} ranking task from 48 to 256 ligands to examine execution at a larger scale (\cref{sec:si-stk17b}).

These results suggest a continuing role for domain-specific agents alongside general-purpose coding agents.
A further consideration, only partly captured by numerical benchmarks, is the reliability supported by established scientific tools and tested workflows.
Established implementations with documented methods and tested behaviour provide a stronger basis for trust than ad hoc scripts generated during an individual run.
Task definitions, judging methods, total model costs, and per-task results are provided in \cref{sec:si-benchmark}.
A worked example of shape and pharmacophore comparison is provided in \cref{sec:si-shape}.

\section{Case studies}

The following case studies evaluate \farmaco{} through discovery-oriented campaigns and technical workflow benchmarks.
For TBXT\footnote{T-box transcription factor T, also known as brachyury.}, compounds proposed by the agent are compared with those selected and purchased in an independent expert-led campaign using a common docking protocol and molecular property assessments.
For PGK2, with PGK1 as a counter-target\footnote{PGK2 and PGK1 denote phosphoglycerate kinase 2 and 1, respectively.}, we examine a workflow that iterates generation and evaluation until a specified number of candidates meet criteria for molecular properties, pocket interactions, and predicted score differences between targets.
Further case studies examine hit enrichment and pose recovery for AmpC\footnote{AmpC $\beta$-lactamase.} and candidate selection for KpThiL\footnote{Thiamine-monophosphate kinase from \textit{Klebsiella pneumoniae}.}.
The final case study examines free energy calculations and analogue design for TYK2\footnote{Tyrosine kinase 2.}.

\subsection{Retrospective TBXT ligand prioritization}
\label{sec:tbxt}

TBXT is a lineage-survival driver in chordoma and contains a shallow pocket, denoted A$'$, that has recently been shown to accommodate small-molecule ligands \cite{newmanStructuralInsightsHuman2025}.
Before this study, team members independently completed a structure-based campaign against this pocket and purchased 40 compounds for experimental evaluation.
Structures were available for 39 of these compounds; one compound was no longer listed in the \href{https://www.chemdiv.com/}{ChemDiv} catalogue and was therefore excluded from further analysis.
These 39 compounds provided a computational reference set selected without \farm{}.

The initial prompt asked \farm{} to prepare the target and pocket, generate 700--1,000 molecular samples, and prioritize candidates for catalogue matching (\cref{sec:si-tbxt}).
The workflow produced 705 ligand records.

RDKit returned molecular descriptors for 693 of the 705 generated ligand records; the remaining 12 were flagged as invalid and excluded from downstream selection.
The agent combined molecular descriptors, structural-alert screening, and \ac{admet} predictions to filter compounds by physicochemical properties, \ac{pains} and Brenk alerts, and predicted hERG and Ames risks, retaining 59 compounds.
\farm{} then organized these compounds into 20 descriptor-based clusters and used Boltz-2 predictions to select up to two compounds per cluster, yielding 34 representatives.
Finally, it searched the ChemDiv and \href{https://enamine.net/}{Enamine} catalogues, returning matches for 20 and 24 of the 34 representatives, respectively; the 0.7 similarity cutoff was specified for ChemDiv.

\begin{figure}
      \centering
      \includegraphics[width=\linewidth]{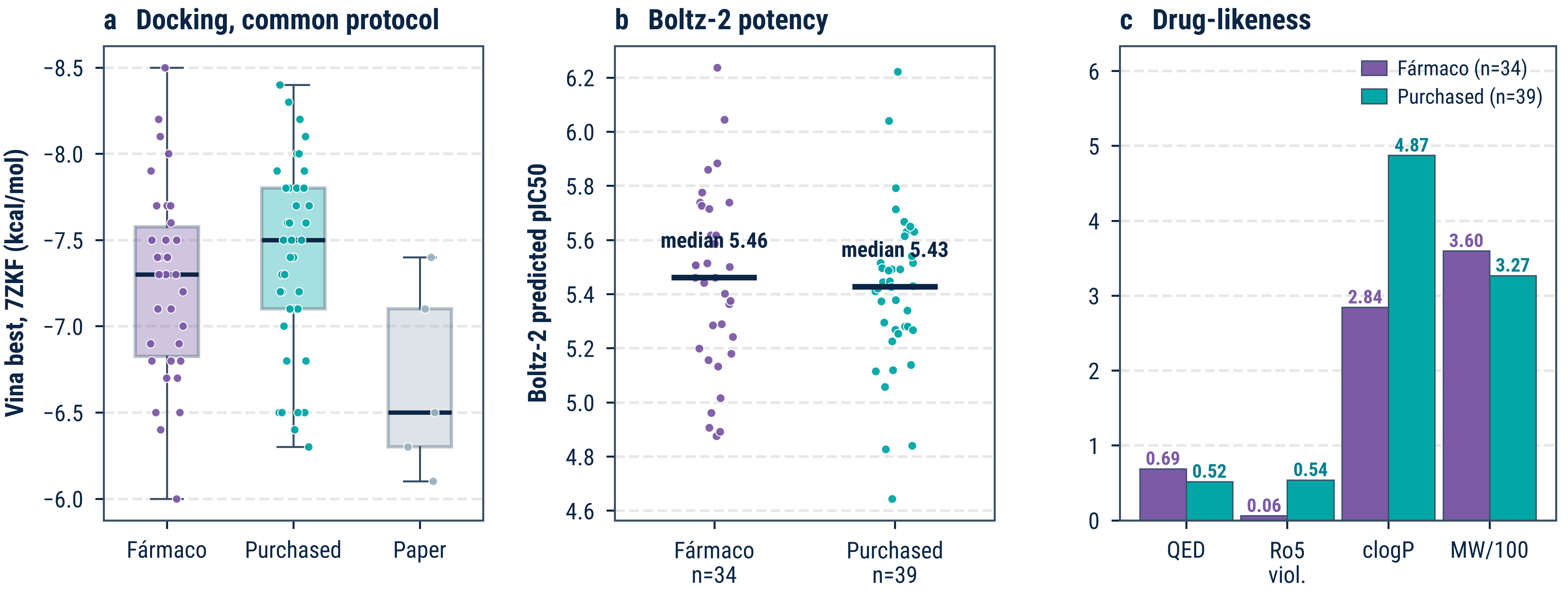}
      \caption{\textbf{Retrospective computational comparison of \farm{} and an independently completed expert-led TBXT campaign.}
            Both campaigns targeted pocket A$'$; panels (a--c) compare 34 generated compounds selected by the agent with 39 compounds purchased in the expert-led campaign.
            \textbf{(a)} Common-protocol docking scores against \acs*{pdb} 7ZKF, including five published reference ligands.
            \textbf{(b)} Boltz-2 converted affinity scores ($6-\texttt{affinity\_pred\_value}$; see \cref{sec:si-tbxt-methods}).
            \textbf{(c)} Cohort means of \acs*{qed} (higher is better), rule-of-five violations, calculated \(\log P\) and molecular weight divided by 100.
            Calculated \(\log P\) is the base-10 logarithm of the predicted octanol--water partition coefficient.
            Docking and Boltz-2 scores are predictions; experimental affinities are available only for the five reference ligands (\cref{sec:si-reference-ligands}).}
      \label{fig:tbxt}
\end{figure}

For a common retrospective comparison, the 34 generated representatives selected by the agent (not the catalogue matches), the 39 purchased compounds, and five published reference ligands were evaluated with the same docking procedure against \ac{pdb} 7ZKF.
This redocking was carried out outside \farm{} using AutoDock Vina \cite{trottAutoDockVinaImproving2010}, applied uniformly to all three sets so that no set inherited a score from its own pipeline.\footnote{
      AutoDock Vina is not among the tools integrated into \farm{}; the comparison therefore evaluates the compounds \farm{} proposed, not a docking capability of the system itself.
}
Preparation settings and software versions are given in \cref{sec:si-tbxt}.
\Cref{fig:tbxt} compares the compounds generated by \farm{} with those selected through the earlier human-led campaign.
Panel~(a) presents the docking-score distributions for the agent-selected and purchased compounds, together with the experimentally validated ligands from the previous study as a reference.
Both the agent-selected and purchased compounds achieved more favourable docking scores than the reference ligands.
Panel~(b) compares the Boltz-2 predictions.
The median predicted \(\mathrm{pIC}_{50}\) was 5.46 for the agent-selected compounds and 5.43 for the purchased set.\footnote{
      Boltz-2 scores were obtained through different services and scoring batches, so their comparison is descriptive (\cref{sec:si-boltz-runs}).
}
The agent-selected compounds showed more favourable developability profiles.
Compared with the purchased set, they had a higher mean \ac{qed} (0.69 versus 0.52), fewer mean Lipinski rule-of-five violations (0.06 versus 0.54), and a lower mean calculated \(\log P\) (2.84 versus 4.87).
This comparison primarily demonstrates \farm{}'s enforcement of the developability constraints explicitly provided to the agent.

\Cref{tab:tbxt} summarizes the selection counts, docking performance, Boltz-2 scores, developability descriptors, and scaffold diversity of the two campaigns.
The 34 agent-selected compounds represented 34 distinct Murcko scaffold groups, whereas the 39 purchased compounds represented 30 groups.
The two final sets also occupied distinct regions of chemical fingerprint space.
The maximum Tanimoto similarity between an agent-selected compound and the purchased set was 0.30, and our structure-level comparison identified neither identical molecules nor shared Bemis--Murcko scaffolds.

Overall, this single-target, retrospective analysis shows that \farm{} executed a multi-stage, constraint-aware computational prioritization workflow and proposed compounds that were computationally competitive under one common docking protocol while being chemically distinct from the expert-selected set.

\begin{table}
      \centering
      \small
      \caption{\textbf{Retrospective computational comparison on TBXT pocket A$'$.}
            Docking scores were calculated using one common protocol for both sets; lower values indicate more favorable predicted binding.
            Boltz-2 converted affinity scores are $6-\texttt{affinity\_pred\_value}$.
            Descriptor values were calculated consistently for the 34 compounds selected by \farm{} and the 39 compounds with retrievable structures among the 40 purchased in the expert-led campaign.
            Property measures include \acs*{qed} and calculated \(\log P\), the base-10 logarithm of the predicted octanol--water partition coefficient.
            Filtering covers \acs*{admet} and structural alerts.
            The developability comparison is descriptive because the two campaigns used different selection constraints.}
      \label{tab:tbxt}
      \begin{tabular}{lrr}
            \toprule
                                                             & \textbf{\farm{}}     & \textbf{Expert-led campaign}                \\
            \midrule
            Generated $\rightarrow$ retained for comparison  & $705 \rightarrow 34$ & $\sim$600+ $\rightarrow 39$ of 40 purchased \\
            Best docking score (kcal\,mol$^{-1}$)            & $-8.5$               & $-8.4$                                      \\
            Median docking score (kcal\,mol$^{-1}$)          & $-7.3$               & $-7.5$                                      \\
            Median Boltz-2 converted affinity score          & 5.46                 & 5.43                                        \\
            \addlinespace
            Mean \acs*{qed}                                  & 0.69                 & 0.52                                        \\
            Mean rule-of-five violations                     & 0.06                 & 0.54                                        \\
            Mean calculated \(\log P\)                       & 2.84                 & 4.87                                        \\
            Explicit \acs*{admet}/structural-alert filtering & yes                  & no                                          \\
            Murcko scaffold groups / compounds               & 34 / 34              & 30 / 39                                     \\
            \bottomrule
      \end{tabular}
\end{table}

\subsection{Adaptive multistage selection of PGK2 ligand candidates using PGK1 as a counter-target}
\label{sec:pgk2}

We evaluated an adaptive candidate-selection workflow targeting human PGK2 with PGK1 as a counter-target (\cref{sec:si-pgk2}).
The prompt specified sequential criteria and an adaptive rule that repeated generation and evaluation within predefined limits until at least five candidates were obtained.
Across two search cycles, \farm{} conducted four PocketXMol generations and registered 369 molecules before meeting the stopping condition.
In cycle 1, three candidates were selected from 181 generated molecules according to the prespecified relative-score criterion; cycle 2 added two candidates from another 188 molecules.
Thus, the adaptive iteration rule determined the search scale rather than a fixed initial library size.

\begin{figure}
      \centering
      \includegraphics[width=\linewidth]{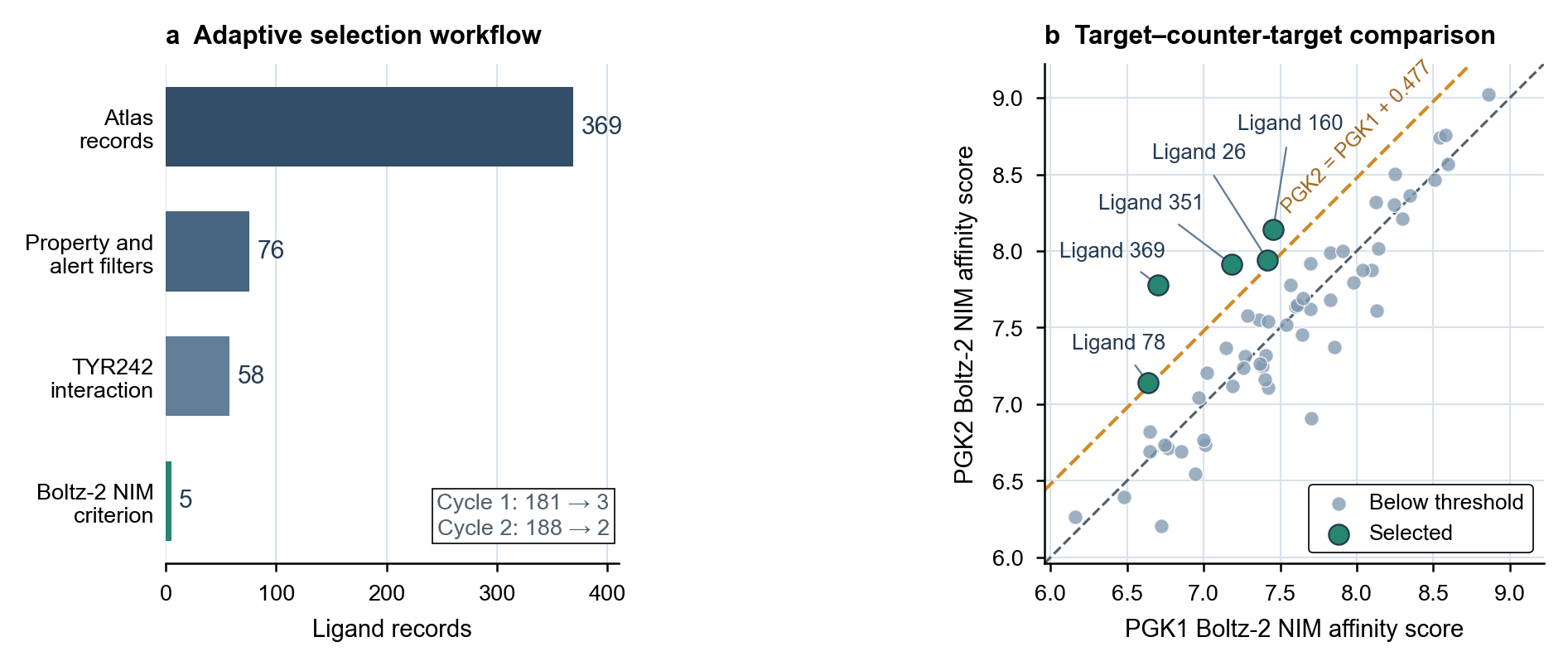}
      \caption{\textbf{Adaptive multistage selection of PGK2 ligand candidates using PGK1 as a counter-target.}
            \textbf{(a)} Candidate counts retained at successive stages of the workflow.
            Across two adaptive search cycles, 369 compounds were registered in \farmacoatlas{}, of which 76 passed the property and structural-alert filters, 58 met the prespecified PGK2 TYR242 interaction criterion, and 5 met the Boltz-2 target--counter-target score criterion.
            \textbf{(b)} PGK2 and PGK1 Boltz-2 affinity scores for the 58 ligands passing Stage~1.
            The grey dashed diagonal denotes equal PGK2 and PGK1 scores, whereas the orange dashed line denotes the prespecified selection threshold, PGK2 = PGK1 + 0.477.
            TYR242 denotes tyrosine residue 242.
            Green markers indicate the five selected candidates, labelled individually; grey markers indicate the remaining compounds.}
      \label{fig:pgk2}
\end{figure}

In Stage 1, PocketXMol \cite{pengUnifiedModeling3D2026} generated molecules using the specified PGK2 pocket center and radius.
\Ac{pains} screening, \ac{qed} and synthetic accessibility thresholds, Lipinski’s rule-of-five, and related filters were then applied, yielding 76 of 369 molecules that met these criteria.
The passing molecules were docked against PGK2 using DiffDock \cite{corsoDiffDockDiffusionSteps2023,corsoDeepConfidentSteps2024}, and protein–ligand complexes were constructed from the rank-1 poses.
\ac{plip} \cite{schakePLIP2025Introducing2025} was subsequently used to evaluate hydrogen-bond or hydrophobic interactions with the prespecified pocket residue tyrosine (TYR) 242.
Of the 76 molecules, 58 satisfied the TYR242 interaction criterion and advanced to Stage 2 (\cref{fig:pgk2}a).

In Stage 2, the 58 ligands passing Stage 1 were independently evaluated against PGK2 and PGK1 by Boltz-2 co-folding, together with DiffDock calculations against PGK1.
The prespecified threshold required the PGK2 minus PGK1 Boltz-2 affinity-score difference to be at least 0.477.
Five molecules met this threshold, with model-derived differences ranging from 0.501 to 1.076 (\cref{fig:pgk2}b and \cref{tab:pgk2}).
In their Stage 1 PGK2 poses, three of these molecules formed hydrogen bonds with TYR242 and two formed hydrophobic interactions.
\acs*{admet}-\acs*{ai} predictions were subsequently obtained for these five molecules.
For the five selected candidates, predicted PGK2 binding probabilities ranged from 0.172 to 0.371.

\begin{table}
      \centering
      \small
      \caption{\textbf{Computational results for the five PGK2 ligand candidates selected using PGK1 as a counter-target.}
            Scores are the archived Boltz-2 \texttt{affinity\_pic50} outputs (\cref{sec:si-pgk2}).
            Higher scores indicate stronger predicted binding, and $\Delta$ score is the PGK2 score minus the PGK1 score.
            TYR242 denotes tyrosine residue 242; its interactions were identified by \acs*{plip} analysis of the PGK2 DiffDock rank-1 poses.}
      \label{tab:pgk2}
      \begin{tabular}{rlrrr}
            \toprule
            \textbf{Ligand} & \textbf{TYR242 interaction} & \textbf{\shortstack{Boltz-2 affinity\\score (PGK2)}} & \textbf{\shortstack{Boltz-2 affinity\\score (PGK1)}} & \textbf{$\Delta$ score} \\
            \midrule
            26              & Hydrophobic                 & 7.943                                                & 7.417                                                & 0.526                   \\
            78              & Hydrogen bond               & 7.140                                                & 6.639                                                & 0.501                   \\
            160             & Hydrogen bond               & 8.138                                                & 7.454                                                & 0.684                   \\
            351             & Hydrophobic                 & 7.914                                                & 7.182                                                & 0.731                   \\
            369             & Hydrogen bond               & 7.779                                                & 6.703                                                & 1.076                   \\
            \bottomrule
      \end{tabular}
\end{table}

The adaptive workflow reduced 369 generated molecules to five traceable candidates meeting the reported property, interaction, and Boltz-2 affinity-score criteria.
We plan to evaluate these candidates in future experimental binding and selectivity studies.

\subsection{Retrospective AmpC \texorpdfstring{$\beta$}{beta}-lactamase virtual-screening benchmark}
\label{sec:ampc}

The two preceding case studies evaluate \farm{} against an expert-led campaign and in a prospective pilot, in both cases without experimental outcomes for the proposed compounds.
We therefore also evaluated a retrospective workflow using published screening outcomes and experimental structures.
Liu et al. docked a 1.7-billion-molecule make-on-demand library against AmpC $\beta$-lactamase, tested 1{,}521 compounds experimentally, and solved co-crystal structures for four hits \cite{liuImpactLibrarySize2025}.
The released dataset contains molecular structures, original DOCK scores and global ranks, similarity coefficients, compound-selection metadata, and experimental outcomes, enabling assessment of enrichment and pose recovery against experimental references.
Eight scripted prompts and two follow-up questions guided \farm{} through dataset analysis, retrospective compound selection, pocket detection, structural comparison, \ac{admet} triage, and molecular generation in a single conversation.

For the docking task, \farm{} selected the 50 compounds with the best original global docking ranks from the 1{,}521 experimentally tested compounds.
The four available co-crystal structures provided experimental references for assessing binding-pocket identification and ligand-pose recovery.
\farm{} delegated pocket investigation to its specialized subagent, which used P2Rank \cite{krivakP2RankMachineLearning2018} and Pocketeer \cite{leguillouxFpocketOpenSource2009} to locate the catalytic cleft in the receptor and co-crystal structures.
It then docked the four co-crystal ligands into the 1L2S-derived receptor with DiffDock and compared their poses and interaction patterns with the experimental structures.
Docking did not consistently recover the experimental poses or interaction patterns (\cref{tab:ampc}).

In a follow-up prompt, \farm{} used \acs*{admet}-\acs*{ai} \cite{swansonADMETAIMachineLearning2024} to assess developability across all 298 compounds with quantified experimental $K_\mathrm{i}$ values, which we count as hits.
Of the 298 hits, 215 passed the \ac{pains}/Brenk structural-alert screen, and 71 returned an exact match in the queried Enamine REAL catalogue.
Predicted \ac{dili} risk was high for most hits and remained a concern even among compounds combining activity, clean structural-alert profiles, and catalogue availability.

\begin{figure}
      \centering
      \includegraphics[width=\linewidth]{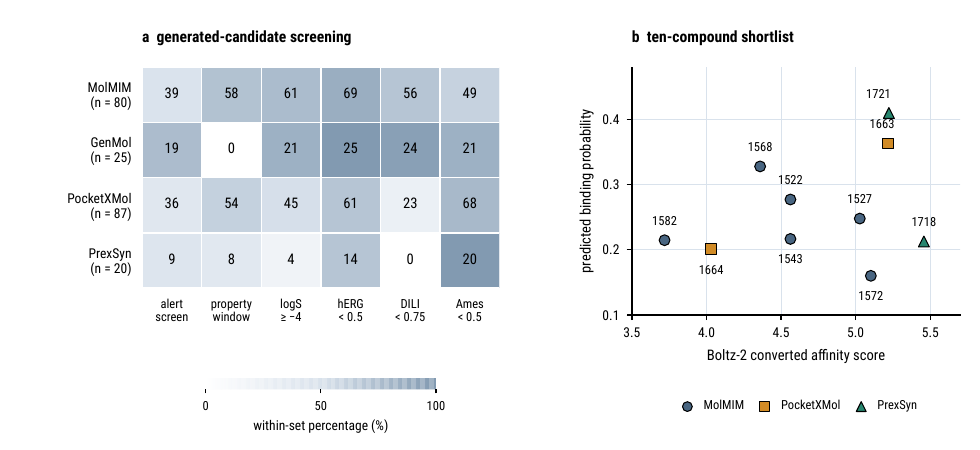}
      \caption{\textbf{Computational prioritization of newly generated AmpC candidates.}
            \textbf{(a)} Screening profiles of the 212 candidates from four generation and analogue-search methods.
            Cells give counts meeting each criterion, with colour indicating the percentage within each set.
            Each column applies its criterion separately.
            \textbf{(b)} Boltz-2 converted affinity scores and predicted binding probabilities for the ten shortlisted candidates; labels identify compounds.
            Higher values are better on both axes.}
      \label{fig:ampc}
\end{figure}

The next prompt asked \farm{} to propose analogues of the strongest experimental hits, aiming to retain similarity to active compounds while improving predicted developability.
The resulting candidates were assessed using molecular properties, structural alerts, and \ac{admet} predictions.
The comparison included 80 MolMIM candidates \cite{reidenbachImprovingSmallMolecule2023}, 25 GenMol candidates \cite{leeGenMolDrugDiscovery2025}, 87 PocketXMol candidates \cite{pengUnifiedModeling3D2026}, and 20 distinct PrexSyn analogues \cite{luoEfficientProgrammableExploration2025}.
\Cref{fig:ampc}a summarizes the properties and predicted toxicity profiles of the 212 candidates analyzed.

To assess predicted binding, a subsequent request directed \farm{} to submit a ten-compound shortlist drawn from the generated sets for Boltz-2 co-folding and affinity prediction against the 1L2S sequence \cite{passaroBoltz2AccurateEfficient2025}; all ten predictions were completed.
The predicted $\mathrm{pIC}_{50}$ values span 3.72 to 5.46, with the two PrexSyn analogues and one PocketXMol sample ranked highest.
The returned binder probabilities range from 0.160 to 0.410 (\cref{fig:ampc}b).
Six of the ten compounds come from MolMIM, two from PrexSyn, and two from PocketXMol.

To assess the contribution of database-supported execution, we repeated the same ten prompts using the plain-tool-call configuration (\cref{sec:benchmark}).
Including the successful Boltz-2 step in both versions, the plain-tool-call configuration was less expensive (USD 14.74 against 17.66) and faster (1.43 against 2.97~h).
For the original ten turns, the plain-tool-call configuration issued 103 model calls against 303: the database configuration spends a separate call on each query, workflow invocation, and schema or subset description.
The per-turn comparison shows where the database reduced or increased costs.
The database configuration was cheaper on four of the original ten turns.
Three of these involved bulk tabular data: retrospective selection over 1{,}521 compounds (USD 0.78 against 2.03), pocket detection (1.07 against 1.22), and \ac{admet} triage over 298 hits (2.42 against 2.62).
It was more expensive on the turns dominated by many small exploratory steps, most clearly the analogue-search follow-up (3.76 against 1.09, 63 model calls against five).
Holding tabular state in a queried store rather than in context therefore pays whenever a turn reasons over a large record set, and costs whenever a turn is a long sequence of individually small operations.
Detailed methods, score definitions, and per-turn results are provided in \cref{sec:si-ampc}.

\subsection{Dimeric KpThiL Prospective Ligand Selection}

KpThiL is a ThiL enzyme from \textit{Klebsiella pneumoniae}, represented here by a newly refined experimental structure not yet deposited in the \ac{pdb}.
The enzyme is a homodimer comprising chains~A and~B, each containing one non-covalently bound \ac{tps} molecule.
The second substrate, \ac{atp}, was not resolved in the experimental density.
Inhibitors of \textit{Pseudomonas aeruginosa} ThiL have been reported \cite{kimThiLEnzymeValid2020,liThiadiazolylideneMorpholineCompound2026}.
No ligand-screening dataset currently exists for this target.

In this context, \farm{} proposed a \emph{ranked, prospective test set of purchasable small molecules} for KpThiL.
Because no prior binding data are available, this represents a genuinely \emph{ab initio} ligand-selection problem.
Accordingly, \farm{} reasoned directly from the experimental structure, first defining the substrate-binding pocket and then selecting candidate compounds.
\farm{} compared the \ac{tps}-contacting residues in the two sites and selected chain~A as representative, retaining the full dimer to account for partner-chain contacts.
P2Rank and Pocketeer identified pockets overlapping the \ac{tps}-binding region.

Starting from substrate-analogue seeds comprising thiamine, thiamine monophosphate, pyrithiamine, and oxythiamine, \farm{} queried the Enamine and ChemDiv catalogues by similarity, yielding 189 raw hits.
It then expanded the chemical space using a MolMIM generative sampler to produce 40 \ac{qed}-optimised molecules, whose Enamine similarity searches returned another 736 hits, bringing the aggregate to 925 raw supplier hits across the searches.
From the pooled candidates, \farm{} selected a ranked, prospective test set of 50 purchasable Enamine-catalogue compounds with distinct recorded \ac{smiles}.
All 50 selected compounds passed \ac{pains} structural-alert screening and had valid RDKit descriptors.

In a follow-up session, \farm{} generated ten molecules with PocketXMol using the experimentally defined \ac{tps} pocket.
All ten passed \ac{pains} screening and PoseBusters molecular-geometry checks and had valid RDKit descriptors.
PrexSyn searches succeeded for nine generated molecules, returning 45 analogue hits with proposed synthesis routes.

A subsequent standalone follow-up reassessed all 105 candidates (the 50 catalogue compounds, ten PocketXMol molecules, and 45 PrexSyn hits) using Boltz-2 and \acs*{admet}-\acs*{ai}.
The agent aligned the predicted monomeric complexes to experimental chain~A and assessed ligand placement against the retained dimer.
Nine poses met the \ac{tps}-pocket proximity/contact heuristic, three of which had more than three ligand heavy atoms in potential clash with the experimental protein (distances below 2~\AA{}).
The highest converted affinity score, $\mathrm{pIC}_{50}=6.66$, belonged to a pose whose ligand centroid lay approximately 27~\AA{} from the \ac{tps} centroid, illustrating why candidate prioritization considered both predicted affinity and placement in the \ac{tps}-binding pocket.
The resulting heuristic ranking placed a \ac{tps}-like catalogue reference first; the highest-ranked novel generated molecule had predicted $\mathrm{pIC}_{50}=5.50$, but a binder probability of 0.152 and high predicted hERG and \ac{dili} scores.
The agent recommended a \ac{tps}-like catalogue compound and a related non-phosphorylated comparator for initial biochemical testing.
For follow-up, it proposed a nitrogen-rich polycyclic lactam generated by PocketXMol and, optionally, a PrexSyn analogue containing quinoline and piperazine rings, subject to synthesis or confirmed availability.
Detailed methods and results are provided in \cref{sec:si-kpthil}.

\subsection{Relative binding free energies for TYK2}
\label{subsec:tyk2}

We evaluated \farm{} on \ac{rbfe} calculations and analogue design for a congeneric TYK2 series of 4-aminopyridine benzamides \cite{liangLeadIdentificationNovel2013,liangLeadOptimization4Aminopyridine2013}.
The first benchmark session tested a fluorine-to-chlorine substitution on a \textit{cis}-cyclopropane, and the second examined network design for 13 ligands.
Both sessions used the protein structure and bound ligand poses of a public free-energy benchmark set \cite{hahnBestPracticesConstructing2022,baumannLargescaleCollaborativeAssessment2025}.
For these sessions, published $K_\mathrm{i}$ values were withheld from the agent and are used here only for retrospective comparison.
Calculations were performed on the Azulene Studio platform provided by \href{https://www.azulenelabs.com/}{Azulene Labs}.
The agent submitted calculations as jobs, recorded them in \farmacoatlas{}, and retrieved the results in later user turns (\cref{sec:si-tools}).

\subsubsection{Network design and initial calculations}
\label{subsec:tyk2-network}

\farm{} was asked to plan an \ac{rbfe} network for 13 ligands.
The agent computed pairwise fingerprint similarities, maximum common substructures, common-core deviations between the supplied poses, and protein--ligand clashes.
From these, it proposed a 12-edge spanning tree with seven additional cycle-closure edges.
The proposal separated computed metrics from qualitative judgement.
The agent identified four potentially challenging transformation types: the bridge between acyclic and cyclopropyl amides, ring-size changes, polar linker changes to urea and carbamate, and the stereochemistry of substituted cyclopropanes.
Nine of its 19 edges also appear in an 18-edge \ac{lomap} reference network \cite{liuLeadOptimizationMapper2013} that was not provided to the agent.
Asked to submit ``the first 6 most interesting calculations'', the agent submitted the first six edges of its table, a star around EJM42.
Each calculation used 10~ps of production per replica.
A single collection call in the next turn returned five results and one failure (\cref{tab:tyk2}).
Against the experiment, the five completed edges have a mean unsigned error of 1.12~kcal\,mol$^{-1}$ and four of five correct signs.

\begin{table}
      \centering
      \small
      \caption{\textbf{First six \acs*{rbfe} edges of the agent-designed TYK2 network.}
            Ligand labels abbreviate the benchmark identifiers (\cref{sec:si-tyk2-benchmark}).
            $\Delta\Delta G$ values (kcal\,mol$^{-1}$) are relative to EJM42; negative values favour the candidate.
            Calculated values are Azulene Studio estimates with 5~ps equilibration and 10~ps production per replica and three repeats; the uncertainty is the one Azulene Studio reports.
            Experimental values are $RT \ln(K_{\mathrm{i},B}/K_{\mathrm{i},A})$ at 298.15~K from published $K_\mathrm{i}$ \cite{liangLeadIdentificationNovel2013}, which were not shown to the agent.
            A and B denote the reference and candidate ligands, respectively.}
      \label{tab:tyk2}
      \begin{tabular}{llrr}
            \toprule
            \textbf{Candidate} & \textbf{Transformation from EJM42} & \textbf{Calculated $\Delta\Delta G$} & \textbf{Experimental $\Delta\Delta G$} \\
            \midrule
            EJM31              & ethyl $\to$ methyl                 & $-1.06 \pm 0.87$                     & $+0.24 \pm 0.25$                       \\
            EJM43              & ethyl $\to$ isopropyl              & $+2.20 \pm 1.22$                     & $+1.53 \pm 0.25$                       \\
            EJM50              & ethyl $\to$ hydroxymethyl          & $+0.50 \pm 0.72$                     & $+0.81 \pm 0.25$                       \\
            EJM55              & propanamide $\to$ methyl carbamate & failed                               & $+0.58 \pm 0.25$                       \\
            EJM54              & propanamide $\to$ ethyl urea       & $-0.05 \pm 0.81$                     & $-0.75 \pm 0.25$                       \\
            EJM45              & ethyl $\to$ cyclopropylmethyl      & $+2.86 \pm 0.65$                     & $+0.23 \pm 0.25$                       \\
            \bottomrule
      \end{tabular}
\end{table}

\subsubsection{Interactive analogue design and \acs*{rbfe} follow-up}
\label{subsec:tyk2-design}

We also tested an interactive TYK2 design campaign starting from a supplied receptor and one posed 4-aminopyridine benzamide seed from a public free-energy benchmark \cite{hahnBestPracticesConstructing2022,baumannLargescaleCollaborativeAssessment2025}.
The user requested independent analogue design that balances predicted potency, molecular properties, and synthetic feasibility, excludes published TYK2 compounds and activity data, and defers free-energy calculations.
The campaign used \texttt{gpt-6-sol}; its complete execution record is described in \cref{sec:si-tyk2-design}.

The agent first designed eight analogues and shortlisted four, then used MolMIM, GenMol, PocketXMol, and PrexSyn in response to follow-up requests.
It retained 11 distinct candidates for investigation, excluding duplicate outputs and GenMol fragments that did not preserve the intended scaffold.
Docking, molecular-property predictions, structural alerts, and interaction analysis supported initial triage.
Seed-templated conformer generation supplied alternative poses and resolved a protein-distance warning for MolMIM-generated ligand~15, which replaces one ortho chlorine of the seed with fluorine.
The agent selected five candidates for further evaluation based on their predicted binding poses.

The agent obtained a protonated, water-free receptor using Rowan \cite{rowanScientific2026} with PDBFixer and terminal caps.
Two seed-to-candidate \ac{rbfe} calculations were submitted for ligands~2 and~5, with 0.5~ns equilibration, 5~ns production per replica, and three repeats.
After these initial submissions, Boltz-2 screening covered the seed and all 11 candidates.
Both initial \ac{rbfe} calculations completed and favoured the seed (\cref{tab:tyk2-design}).
Ligand~15 was the only candidate whose reported affinity score and binding probability both exceeded the seed's, motivating a third \ac{rbfe} calculation with the same protocol.
Its $\Delta\Delta G = +0.018 \pm 0.099$~kcal\,mol$^{-1}$ suggested similar binding to the seed, whereas Boltz-2 had suggested improved potency.

\begin{table}
      \centering
      \small
      \caption{\textbf{TYK2 seed-to-candidate follow-up.}
            Ligand identifiers refer to the archived database.
            Azulene Studio reports $\Delta\Delta G = \Delta G(B)-\Delta G(\mathrm{seed})$ in kcal\,mol$^{-1}$; positive values favour the seed.
            All three edges used the same water-free receptor, 0.5~ns equilibration, 5~ns production per replica, and three repeats.
            Uncertainties are reported by Azulene Studio.
            Predicted $\mathrm{pIC}_{50}$ values were normalized from the archived Boltz-2 outputs as described in \cref{sec:si-tyk2-design}.}
      \label{tab:tyk2-design}
      \begin{tabular}{llrrr}
            \toprule
            \textbf{Ligand} & \textbf{Benzoyl-ring substitution} & \textbf{Boltz-2 $\mathrm{pIC}_{50}$} & \textbf{\shortstack{Boltz-2\\binding probability}} & \textbf{\shortstack{\acs*{rbfe}\\$\Delta\Delta G$}} \\
            \midrule
            Seed            & 2,6-dichloro                       & 6.500                                & 0.805                                              & ---                                                 \\
            2               & 2-chloro-4-fluoro                  & 6.289                                & 0.730                                              & $+1.710 \pm 0.114$                                  \\
            5               & 2,6-difluoro                       & 6.169                                & 0.742                                              & $+0.807 \pm 0.205$                                  \\
            15              & 2-chloro-6-fluoro                  & 6.785                                & 0.828                                              & $+0.018 \pm 0.099$                                  \\
            \bottomrule
      \end{tabular}
\end{table}

\section{Discussion}

\farmaco{} brings together adaptive workflow coordination and persistent scientific state for early-stage drug discovery.
In the 22-task benchmark, the database-backed configuration combined the smallest median peak context with model costs close to those of plain tool calls and about half those of Codex.
The case studies showed how the agent combined scientific tools with persistent scientific state in \farmacoatlas{} to address early-stage drug discovery tasks.
Together, these results illustrate how a domain-specific harness can support token-efficient scientific workflows while reliably storing scientific state with traceable provenance.

The capabilities of an agentic \ac{ai} system depend on three factors, each shaped by different stakeholders:
\textit{i)} the capability of the underlying \ac{llm} as a reasoning and decision-making engine;
\textit{ii)} access to computational methods and software, together with their capabilities and limitations, such as compound generation and the accuracy of co-folding and docking for early screening; and
\textit{iii)} the support provided by the harness in which the \ac{llm} operates.
For the first factor, an open question is whether fine-tuning or training models from scratch for domains such as drug discovery offers advantages over general-purpose frontier models with broad training data and cross-domain reasoning capabilities.
The second factor largely depends on progress in open-source initiatives and models.

For the third factor, we expect future systems to combine the flexibility of agent platforms such as \href{https://openclaw.ai/}{OpenClaw} and coding agents such as \href{https://claude.com/product/claude-code}{Claude Code} and \href{https://openai.com/codex/}{Codex} with domain-specific infrastructure such as \farm{}.
This infrastructure can improve reliability through established scientific workflows and persistent, structured storage of scientific state and its provenance.
Tool interfaces should also expose assumptions, limitations, and informative error feedback, as illustrated by the AmpC example (\cref{sec:si-ampc}).
Opaque interfaces can restrict an agent's ability to investigate and assess results.
In such cases, more flexible architectures that let agents inspect code, data, and execution environments may better diagnose and resolve problems.
The limiting factor may be access to computational context, such as why a protein was prepared in a particular protonation state, even when the model has the scientific knowledge needed to assess its implications.
This is consistent with the broader argument that agent performance can be constrained by the execution harness and access to essential information, even when models have the capabilities needed to solve the task \cite{zhangStopComparingLLM2026,trinhHiLBenchHumaninLoopBenchmark2026}.
We view current architectures, including \farm{}, as intermediate steps towards this broader integration.

Agentic \ac{ai} will make conventional workflows and data pipelines in fields such as early-stage drug discovery significantly more adaptive and customizable.
However, integrating a broad range of \textit{in silico} tools to identify potential hit or lead compounds is only the first step.
Agentic solutions will realize their full value only when they also incorporate on-demand experimental data.
In drug discovery, this means coupling \farm{} to automated robots for chemical synthesis and biological assays.
The reasoning agent could then assess the correlation between \textit{in silico} predictions and experimental data and suggest modifications to candidate compounds.
Approaches such as Bayesian optimization \cite{mcdonaldBayesianOptimizationMultiple2025,mullerAgenteOptimaAgentic2026} or active learning \cite{toshBayesianActiveLearning2025} could use these experimental results to guide compound selection for subsequent testing.
Together with virtual screening, this would close the loop between computational predictions and experimental feedback.
As experimental data accumulate, they could refine the predictive models guiding these methods, creating a data flywheel that supports better compound proposals in subsequent rounds.

Closing this loop could, in turn, change how medicinal and computational chemists contribute to discovery.
Medicinal chemists could focus on overarching design decisions and provide the system with additional domain- or entity-specific constraints and guidance.
They would no longer need to do the hands-on work of virtually screening a class of compounds using docking and co-folding methods, checking whether binding poses align, or ensuring compounds meet structural requirements such as passing \ac{pains} filters.
In our longer-term vision, identifying candidate compounds for animal or human studies would become increasingly automatable, shifting the main bottlenecks towards preclinical and clinical evaluation and regulatory approval.
The roles of medicinal and computational chemists in drug discovery would merge, with a deep understanding of interaction mechanisms and downstream constraints on molecular structures becoming much more relevant than conducting simulations.

As increasingly capable agentic systems automate computational execution, human expertise will centre on framing questions, defining domain-specific requirements, and interpreting results.
The human-in-the-loop would act as a verifier, assessing whether the results support the conclusions drawn and what scientific or practical consequences follow.

Our team is working towards the release of the entire family of El Agente tools in a scalable platform for the general public to employ.
We hope that this action will help democratize access as well as help many people make advances towards the cure of diseases that plague humanity, especially those in the neglected disease category \cite{deryckerAntitrypanosomatidDrugDiscovery2023,gyapongCurrentStateFuture2025}.

\section*{Data availability}

Supporting data, workflow instructions, execution records, and analysis scripts for the case studies, multi-task benchmark, and additional STK17B experiment are available at \url{https://github.com/marcelmbn/farmaco-artifacts}.
Package contents and release limitations are documented in the corresponding \texttt{README.md} files.
The \ac{si} describes the database and tool integrations, evaluation methods, and detailed computational results.

\section*{Acknowledgments}
We gratefully acknowledge the longstanding contributions of the Matter Lab’s current and past group members (\url{https://matter.toronto.edu}), in particular, the El Agente subgroup.
We thank Azulene Labs (\url{https://azulenelabs.com}), particularly Nicolas P.~D. Sawaya and Anugraha Thyagatur, for providing access to Azulene Studio for binding free-energy calculations.
M.M. was partially supported through a collaborative partnership with Merck KGaA, Darmstadt, Germany.
Y.O. thanks Daiichi Sankyo Co., Ltd. for financial support.
A.A.-G. thanks Anders~G.~Fr{\o}seth for his generous support.
A.A.-G. also acknowledges the generous support of Natural Resources Canada and the Canada 150 Research Chairs program.
\acknowAC{}
This work was also supported by the \ac{darpa} under Agreement No.~HR0011262E022 and the AI2050 program of Schmidt Sciences.

\clearpage

{
  \small
  \raggedright %
  \interlinepenalty=10000 %
  \bibliography{references}
  \bibliographystyle{assets/plainnat}
}

\clearpage

\appendix
\pdfbookmark[0]{Supporting Information}{supporting-information}
{\Huge\sffamily\bfseries Supporting Information\par}
\startcontents[appendices]
\section*{Contents}
\printcontents[appendices]{l}{1}{\setcounter{tocdepth}{2}}

\section{Data and workflow infrastructure}
\label{app:related}
\label{sec:si-infrastructure}

\subsection{Dataset and structure ingestion}
\label{sec:si-ingestion}

The ingestion specialist constructs a mapping from database field names to selected \ac{csv} columns after inspecting the file schema and representative rows.
Supported entity fields include target names, \ac{pdb} accessions, protein sequences, ligand names, and ligand \ac{smiles}; metric mappings specify floating-point, string, or Boolean values.
The importer validates mappings and values before writing, creates a \texttt{csv\_import} run recording the source filename, and associates imported metrics with that run and the relevant entities.
It creates ordered subsets of referenced targets, ligands, and protein--ligand pairs, and commits the import in one database transaction; an exception rolls back the batch.
The returned summary reports inserted entities and metrics together with the subset identifiers.

The structure-import interface registers an existing workspace file as a \texttt{Structure} record with a kind, format, and path, linked to an existing target, ligand, or both.
This makes externally supplied or refined structures available to workflows consuming persisted structures.
Such imports do not receive a fabricated computational source run.

\subsection{Validation, bounded retrieval, and execution policies}
\label{sec:si-execution}

\begin{figure}
      \centering
      \includegraphics[width=\textwidth]{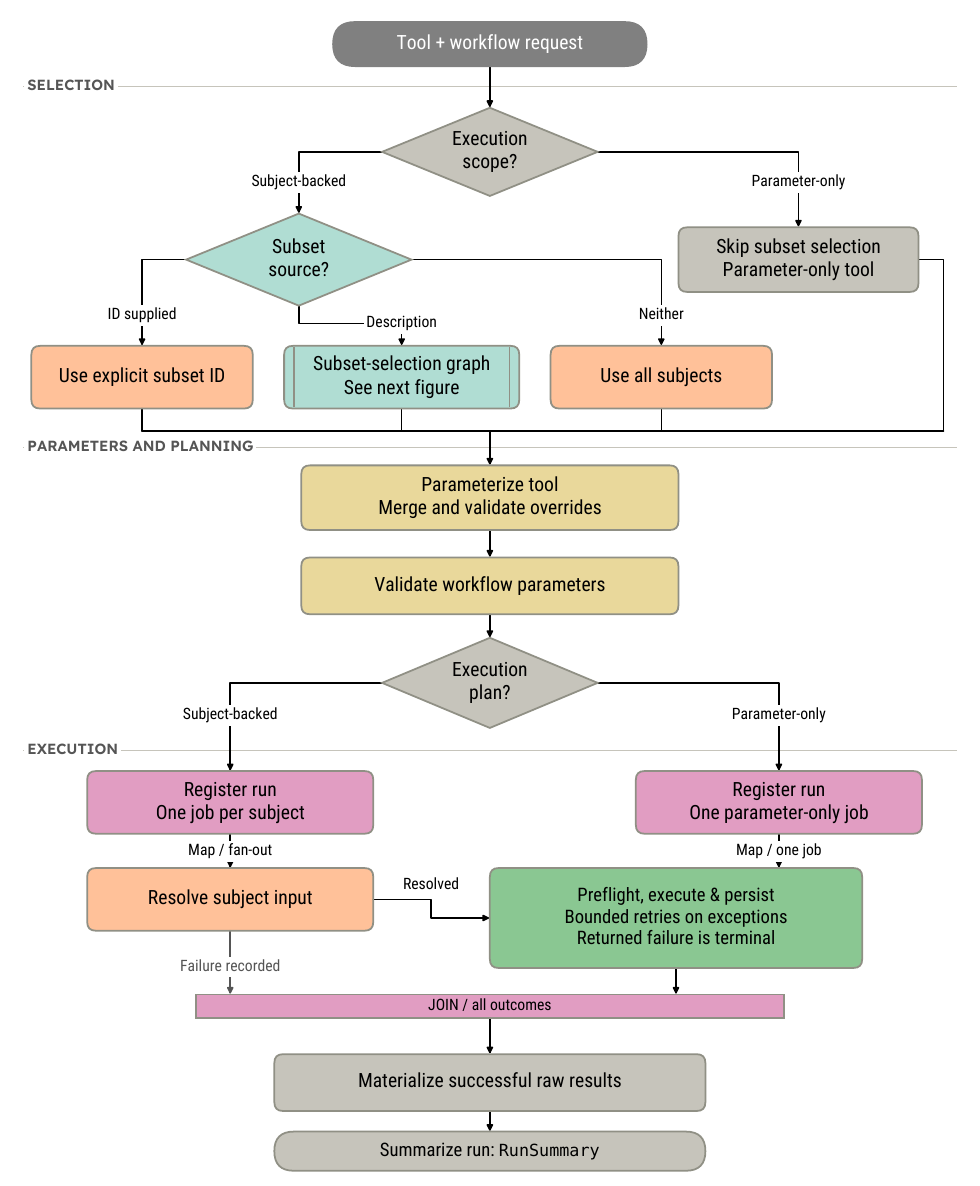}
      \caption{\textbf{Database workflow.}
      Subject jobs fan out through input resolution; parameter-only tools dispatch one job directly to execution.
      Outcomes join before materialization and summary generation.
      Diamonds mark decisions; the double-bordered box links to \cref{fig:si-subset-selection}.
      Colours follow \cref{fig:database_workflow}.}
      \label{fig:si-database-workflow}
\end{figure}

\begin{figure}
      \centering
      \includegraphics[width=\textwidth]{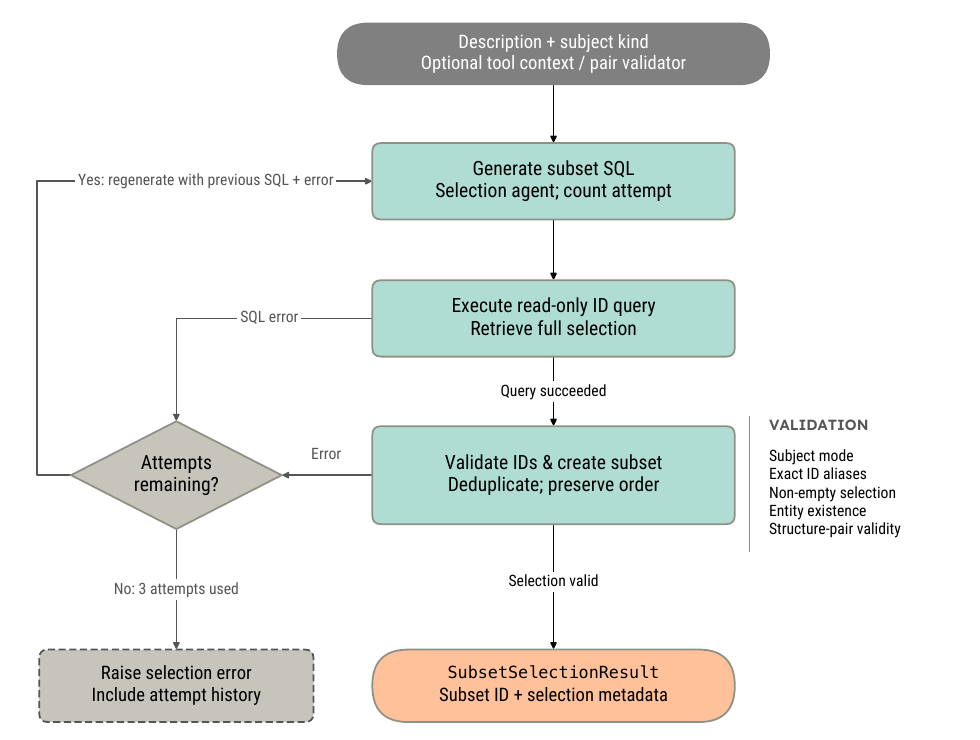}
      \caption{\textbf{Subset-selection graph.}
      A generated read-only \acs*{sql} query produces identifiers for validation and subset creation.
      Errors trigger query regeneration with feedback; three failed attempts lead to an error with attempt history.}
      \label{fig:si-subset-selection}
\end{figure}

The workflow execution graph is implemented with Pydantic Graph.
It implements selection, parameterization, dispatch, execution, result persistence, and summary generation as explicit steps.
\Cref{fig:si-database-workflow,fig:si-subset-selection} show its execution paths and the nested subset-selection graph.
Its subset-selection subgraph checks the requested subject category, exact ID-column aliases, entity existence, and, where applicable, compatibility of ordered structure pairs.
\Ac{sql} execution or selection-validation errors are returned to the selection agent, with a maximum of three selection attempts.
Tool parameters are generated against a Pydantic model and validated before dispatch, including tool-specific semantic checks.
Resolved inputs undergo a further preflight check before retryable execution.

Agent-facing \ac{sql} queries execute in read-only mode and return previews with selected column names, row counts, and truncation metadata.
The default preview contains at most 20 rows, with a maximum requested limit of 100, regardless of whether the query itself contains a \texttt{LIMIT} clause.
Broad \texttt{SELECT *} projections are rejected.
The public query wrapper further removes preview rows when the serialized response exceeds 20,000 characters and advises narrower retrieval.
These preview limits do not truncate the internal ID query used to construct a workflow subset.
File-reading tools return bounded text, and the general file reader rejects protein-structure formats such as \ac{pdb} and \ac{mmcif} in favor of structure-specific tools.

The execution graph resolves and runs each subject independently and collects its outcome at a join step.
The default worker policy permits eight concurrent jobs and two retries after the initial attempt, with exponential backoff; tools may override the concurrency limit and execution timeout.
A returned \texttt{ok=False} outcome is terminal, whereas operational exceptions enter the retry path.
Resolution and preflight failures are recorded without execution retries.
Failure records retain the run, subject, stage, error type and message, and attempt count.
Successful raw results are persisted before tool-specific materialization creates normalized metrics or derived entities.
Materialization failures produce separate failure records while preserving the raw output and allowing later results to be processed.
The run summary counts execution outcomes; materialization failures are recorded separately and therefore require inspection when assessing whether all derived records were created.

\subsection{Remote compute jobs}
\label{sec:si-compute}

PocketXMol and the optional PrexSyn compute backend use a shared asynchronous client to submit finite jobs with explicit inputs, resource timeouts, and required output artifacts.
The client waits for a successful terminal state and confirmed output readiness before retrieving results.
Idempotency keys incorporate the workflow run, tool, subject, parameters, and workspace to identify repeated submissions.
If waiting fails or is cancelled, the client attempts to cancel the remote job and confirm its terminal state; a subsequent attempt after timeout or confirmed abandonment receives a fresh submission identity.
This manages the normal job lifecycle but does not guarantee recovery from abrupt process termination or cancellation before submission returns a job identifier.

\clearpage
\section{Integrated scientific tools}
\label{sec:si-tools}

The following catalogue describes the scientific methods and their integration into \farmaco{}, grouped by role in the discovery workflow (\cref{fig:agent_network}).
Boltz-2, OpenFold2, OpenFold3, MolMIM, GenMol, and DiffDock are accessed as \acp{nim}, hosted by NVIDIA.\footnote{\url{https://build.nvidia.com/models?filters=usecase\%3Ausecase\_drug\_discovery}, accessed March 4, 2026.}

\subsection{Target structures and sequence alignments}

\paragraph{OpenFold2}
OpenFold2 is an open-source implementation and extension of AlphaFold-style protein structure prediction models \cite{ahdritzOpenFoldRetrainingAlphaFold22024}.
It reproduces the core architecture of AlphaFold2 while enabling greater transparency, extensibility, and integration with custom training pipelines.

\paragraph{OpenFold3}
OpenFold3 \cite{openfold3-preview,abramsonAccurateStructurePrediction2024} extends the AlphaFold paradigm to model more complex biomolecular systems, including protein–protein, protein–nucleic acid, and protein–ligand interactions.
Inspired by AlphaFold3-style architectures, it integrates multimodal structural inputs and expanded training data to support prediction of biomolecular complexes, enabling broader applications in structural biology and drug discovery.

\paragraph{Rowan protein preparation}
\href{https://docs.rowansci.com/workflows/protein-preparation}{Rowan} \cite{rowanScientific2026} prepares receptor structures for ligand generation, docking, and binding free-energy calculations.
The integrated \texttt{rowan\_prepare\_protein} tool supports repair of missing atoms and residues, terminal capping, protonation at a specified pH, and hydrogen-position optimization.
The integration defaults to PDBFixer repair, ACE/NME terminal caps, and OpenMM protonation at pH~7.4 \cite{eastmanOpenMM8Molecular2023}.
Chain selection and retention of waters, ions, and cofactors are configurable.
Prepared structures are registered in \farmacoatlas{}, with preparation settings, reports, and logs retained for subsequent inspection.

\paragraph{Multiple sequence alignment search}
The \ac{msa} search tool retrieves homologous protein sequences for a target and returns the resulting alignment in A3M format.
It accesses NVIDIA's ColabFold \acs{msa} Search endpoint and allows the sequence databases, expectation-value threshold, and maximum number of returned sequences to be configured.
Alignments are cached using a key derived from the normalized target sequence and search parameters, allowing them to be reused by downstream structure-prediction workflows such as Boltz-2 without repeating the remote search.

\subsection{Binding-site identification}

\paragraph{Pocketeer}
Pocketeer \cite{leguillouxFpocketOpenSource2009} is a lightweight tool for identifying potential ligand-binding pockets on protein structures.
It analyzes the geometric features of protein surfaces to detect cavities that may accommodate small molecules.
\texttt{pocketeer-0.3.0} is installed in the same Python environment as the agentic system.
In addition, a custom utility maps the atoms defining a detected pocket to the corresponding amino acid residues in the protein structure using the Pocketeer pocket \ac{json} output and the original \ac{pdb} file.

\paragraph{P2Rank}
P2Rank \cite{krivakP2RankMachineLearning2018} is a \ac{ml}–based method for predicting ligand-binding sites on protein structures.
It evaluates local structural environments on the protein surface using engineered features and random forest models to estimate the likelihood of ligand binding.
P2Rank is widely used for automated binding-site identification and prioritization in structure-based drug discovery.
\texttt{p2rank-2.5.1} is made accessible to the agentic Python environment via installing the Java-based program in the same Docker container, and building a Python \ac{api} around the \ac{cli} \ac{io}.
An additional utility exposes the predicted pocket residues as structured data.

\subsection{Ligand generation}

\paragraph{MolMIM}
MolMIM is a probabilistic encoder–decoder model for small-molecule generation, trained on SMILES representations using Mutual Information Machine (MIM) learning.
It learns distributions over chemical structures and can generate novel molecules conditioned on target properties or structural constraints.
MolMIM is particularly useful for exploring chemical space in early-stage drug discovery, enabling scaffold generation, property optimization, and \textit{de novo} ligand design.

\paragraph{GenMol}
GenMol \cite{leeGenMolDrugDiscovery2025} is a generative molecular design framework that produces chemically valid small molecules guided by structural or physicochemical constraints.
By combining deep generative modeling with molecular representations such as graphs or \ac{smiles} strings, GenMol can propose novel candidate compounds for drug discovery workflows, supporting tasks such as scaffold exploration, lead diversification, and property-driven optimization.

\paragraph{PocketXMol}
PocketXMol \cite{pengUnifiedModeling3D2026} is a diffusion-based generative model for structure-based drug design that generates small molecules conditioned on a protein binding pocket.
It integrates geometric representations of the pocket with molecular diffusion processes to propose chemically valid ligands that fit the target site.
PocketXMol enables \textit{de novo} ligand generation guided by structural context and binding-site constraints.
The integration uploads a persisted protein structure and submits a remote compute job specifying pocket coordinates, sampling radius, and the requested number of molecules.
The job returns a structured result file and an archive of generated molecular structures, which the client retrieves for persistence in \farmacoatlas{} (\cref{sec:si-compute}).

\paragraph{PrexSyn}
PrexSyn \cite{luoEfficientProgrammableExploration2025} is a framework for exploring synthetically accessible chemical space through reaction-based molecular generation.
It constructs candidate molecules by combining 115 reaction templates with the \texttt{enamine\_rxn115} fragment library containing about 220k building blocks \cite{gaoGenerativeAINavigating2025}, enabling systematic exploration of compounds that are likely to be synthesizable.
This approach supports property-directed analog generation within realistic synthetic constraints.
\texttt{prexsyn-7b1e984} was installed in a Docker container, including the appropriate weights from \url{https://huggingface.co/datasets/luost26/prexsyn-data}.
The integration supports a local \ac{http} service, selected by default, and an optional remote compute backend.
The latter submits the query ligand and sampling parameters as a finite job and retrieves a required structured result file through the shared compute client (\cref{sec:si-compute}).

\subsection{Pose and affinity prediction}

\paragraph{Boltz-2}
Boltz-2 \cite{passaroBoltz2AccurateEfficient2025} is a structural biology foundation model for predicting both biomolecular complex structures and small molecule–protein binding affinities.
Building on AlphaFold-style co-folding, it integrates enhanced data curation, dynamic ensemble training, and controllability features (e.g., distance constraints and template conditioning).
Notably, Boltz-2 approaches the accuracy of \ac{fep} methods for affinity prediction while being over 1000× faster.

\paragraph{DiffDock}
DiffDock \cite{corsoDiffDockDiffusionSteps2023,corsoDeepConfidentSteps2024} is a diffusion-based deep learning model for protein–ligand docking.
Instead of relying on traditional search heuristics and scoring functions, DiffDock models the docking process as a generative diffusion process that directly predicts ligand poses within protein binding pockets.
This approach improves robustness to initial ligand placement and enables accurate blind docking while maintaining competitive computational efficiency.

\subsection{Structural analysis, validation, and free energy}

\paragraph{\acs*{plip}}
\Ac{plip} \cite{schakePLIP2025Introducing2025} is a tool for automated analysis of non-covalent interactions between proteins and bound ligands.
It detects interaction types such as hydrogen bonds, hydrophobic contacts, salt bridges, $\pi$-$\pi$ stacking, and halogen bonds directly from structural data.
\Ac{plip} is commonly used to interpret docking results, analyze binding modes, and characterize protein–ligand complexes.
\texttt{plip-3.0.0} is installed in the main Docker container hosting the agentic system.

\paragraph{PoseBusters}
PoseBusters \cite{buttenschoenPoseBustersAIbasedDocking2024} is a benchmarking and validation framework for evaluating predicted protein–ligand binding poses.
It performs a series of automated sanity checks on three-dimensional structures resulting from docking, including geometric plausibility, chemical validity, steric clashes, and consistency with structural constraints.
\texttt{posebusters-0.6.3} is installed in the same Python environment as the agentic system.

\paragraph{Azulene Studio binding free-energy calculations}
Azulene Studio, provided by Azulene Labs, runs alchemical absolute and relative binding free-energy calculations, complementing faster learned affinity and docking scores with physics-based estimates.
\farm{} exposes two submission workflows.
\texttt{opal\_abfe\_submit} takes 1--10 persisted protein--ligand complex structures and computes an absolute binding free energy for each from a solvent and a complex leg.
\texttt{opal\_rbfe\_submit} takes a protein-only structure, a fixed reference ligand pose~A, and 1--10 candidate poses~B in the receptor frame, and computes $\Delta\Delta G = \Delta G(B) - \Delta G(A)$ for each A$\to$B edge.
The agent builds networks from such edges; no all-pairs network is generated automatically.
Both workflows support protonation-state assignment at a chosen pH and repeated protocols.
\ac{rbfe} equilibration and production lengths are configurable, with defaults of 5~ps and 10~ps, whereas \ac{abfe} lengths are currently fixed at these values as a cost guard.
Because the calculations run for hours, submission returns immediately.
Each job is stored as a durable record keyed by its effective input, so identical requests reuse an existing job rather than being resubmitted.
Separate \texttt{opal\_collect} and \texttt{opal\_cancel} tools check or cancel given jobs once per call.
Completed free energies, uncertainties, and, for \ac{abfe}, standard-state corrections are then persisted as results linked to the target, ligands, and structures in \farmacoatlas{}.

\subsection{Developability, ligand preparation, and compound sourcing}

\paragraph{\acs*{admet}-\acs*{ai}}
\acs{admet}-\acs{ai} \cite{swansonADMETAIMachineLearning2024} is an \ac{ml} platform for predicting \ac{admet} properties of small molecules.
By leveraging large curated datasets and deep learning models, it enables rapid \textit{in silico} estimation of pharmacokinetic and toxicity-related properties.
A Docker container based on \texttt{python:3.12-slim} hosts \texttt{admet-ai-2.0.1}, which exposes predictions via a FastAPI service.

\paragraph{RDKit-based ligand preparation, filtering, and comparison}
Several complementary cheminformatics operations are implemented with RDKit \cite{rdkit} and exposed collectively through the database-backed workflow interface.
The conformer-generation tool converts ligand \ac{smiles} strings into three-dimensional \ac{sdf} structures using ETKDGv3 embedding, removes geometrically redundant conformers by root-mean-square-deviation pruning, and optionally optimizes and ranks the resulting geometries with the Universal Force Field.
The descriptor tool calculates queryable physicochemical and structural properties, including molecular weight, lipophilicity, topological polar surface area, hydrogen-bond counts, rotatable bonds, ring counts, fraction of sp\textsuperscript{3} carbon atoms, formal charge, \ac{qed}, and Lipinski rule-of-five violations.
Structural-alert screening applies selectable RDKit filter catalogues, including \ac{pains}, Brenk, \ac{nih}, ZINC, and ChEMBL-derived rule sets, to flag potentially problematic chemotypes.
For ligand comparison, RDKit shape alignment reports three-dimensional shape, pharmacophore-feature, and combined similarity scores relative to a chosen reference conformer.
Finally, a reaction-profile filter evaluates compounds against bundled or user-authored \ac{smarts}-based rules to determine whether they are suitable building blocks for transformations such as amidation or Suzuki coupling.

\paragraph{Supplier catalogue search}
The supplier-lookup tool searches the ChemDiv and Enamine catalogues for purchasable compounds using exact, similarity, or substructure queries derived from a ligand's \ac{smiles} representation.
A dedicated FastAPI service provides a common interface to the ChemDiv search service and to Enamine collections exposed through SmallWorld.
Returned catalogue identifiers, structures, molecular formulae, molecular weights, availability information, and backend-specific similarity measures are stored with the workflow result, enabling the agent to prioritize candidates that can be acquired experimentally.

\subsection{Data retrieval and format-conversion utilities}

\paragraph{\acs*{mmcif} to \acs*{pdb} Converter}
The \ac{mmcif} to \ac{pdb} converter \cite{zhangBeEMFastFaithful2023} is a utility tool for transforming macromolecular structure files from the modern \ac{mmcif} format into the legacy \ac{pdb} format.
Since many computational chemistry and docking tools still rely on \ac{pdb} inputs, such conversion utilities facilitate compatibility.

\paragraph{\acs*{pdb} Structure and Sequence Retrieval}
This utility provides automated access to macromolecular structures and sequences from the \href{https://www.rcsb.org}{\ac{rcsb}} \ac{pdb}.
Protein structures are downloaded using Biopython’s \texttt{PDBList} interface based on their four-character \ac{pdb} identifiers.
Downloaded structures are cached locally to avoid redundant network requests and improve workflow efficiency.
In addition, the tool retrieves the corresponding protein sequences in FASTA format\footnote{\url{https://blast.ncbi.nlm.nih.gov/doc/blast-topics/\#fasta}, accessed March 4, 2026.} via the \ac{rcsb} \ac{pdb} FASTA endpoint for the specified \ac{pdb} entry.
Both structural data and sequences are stored within the agent working directory to enable downstream computational workflows.

\paragraph{Clean up protein structures}
This utility cleans protein structure files in \ac{pdb} format using Biopython-based parsing and selection \cite{cockBiopythonFreelyAvailable2009}.
It removes non-protein entities such as water molecules and hetero atoms, retaining only standard protein residues.
The cleaned structure is written to a new \ac{pdb} file in the working directory while preserving the original structure.

\paragraph{Protein–ligand structure merging}
This utility combines a protein structure in \ac{pdb} format with a ligand structure from an \ac{sdf} file into a single complex \ac{pdb} file.
Using RDKit \cite{rdkit}, the ligand coordinates are converted into \ac{pdb} format and appended to the protein structure.

\paragraph{\acs*{smiles}-to-\texttt{XYZ} conversion}
This utility generates three-dimensional molecular geometries from \ac{smiles} strings using RDKit.
The resulting structures are written as XYZ coordinate files for use in downstream molecular modeling workflows.

\paragraph{Web search}
\Ac{ai} web search retrieves up-to-date information from the web using a large language model with integrated search capabilities.
It returns synthesized answers based on external sources.

\clearpage
\section{Case-study methods and provenance}
\label{sec:si-case-studies}

\subsection{TBXT retrospective comparison}
\label{sec:si-tbxt}

\subsubsection{Released data and scope of reproducibility}
\label{sec:si-tbxt-methods}

The released TBXT artifact package contains the target sequence, the prepared 7ZKF receptor used by the agent, the co-crystallized reference ligand used to define pocket A$'$, the agent's pocket report, all four tabular workflow stages, derived comparison tables, and scripts that reconstruct the reported summary statistics and figure.
The 40 expert-campaign compounds are identified by opaque internal accession codes; structures were available for 39 compounds when the retrospective analysis was assembled.
Their molecular structures, supplier identifiers, and docking poses remain withheld while experimental validation is in progress.
Accordingly, the released package supports recalculation of the reported docking distributions, descriptor summaries, and precomputed similarity summaries, but not independent redocking or structure-derived descriptor calculation for the purchased set.
We audited the released package to ensure that no accession-to-structure information was exposed.
No file links a \texttt{Matter-\#\#\#\#} accession to a \ac{smiles} string, InChI, or supplier identifier, and no such mapping appears in the package's Git history.
The accession-to-structure key is stored separately and is not included in the released package.
As the TBXT project is part of an ongoing challenge, the identities of the selected compounds remain blinded and will be disclosed after the challenge is completed.

The common retrospective docking analysis was performed outside \farmaco{} with AutoDock Vina 1.1.2 \cite{trottAutoDockVinaImproving2010}.
The 7ZKF receptor (chain A) was stripped of the co-crystallised ligand A1I1T, phosphate ions and crystallographic waters.
Hydrogens were then added and protonation states assigned at pH~7.4 with PDB2PQR 3.7.1 (PROPKA pKa prediction and hydrogen-bond-network optimisation), and the receptor was converted to a rigid PDBQT file with Meeko, using Gasteiger partial charges and AutoDock atom types.
Ligand conformers were generated with RDKit 2026.03.5 (ETKDGv3, random seed 42) and minimised with MMFF94, and the lowest-energy conformer of each compound was converted to PDBQT with Open Babel 3.1.0 using Gasteiger partial charges.
Docking used a 24~\AA{} cubic box centred at $(13.650,\allowbreak -22.970,\allowbreak -6.120)$, exhaustiveness 16, nine requested binding modes and random seed 42, and the top-ranked mode was reported for each compound.
The same prepared receptor and the same settings were applied to all three compound sets.
The pocket report suggested a box of about $22 \times 25 \times 21$~\AA{} (ligand extent plus 8~\AA{} padding per side) for a future run; \farm{} performed no docking in this campaign, and the 24~\AA{} cubic box was used throughout the retrospective analysis.
Only the 7ZKF docking results are summarized in the main text, so only the prepared 7ZKF receptor is included in the released package.

Boltz-2 produced an affinity regression value and a binder-probability output \cite{passaroBoltz2AccurateEfficient2025}.
The two cohorts were scored through different routes.
The 34 agent-selected compounds were scored by \farm{} through the NVIDIA-hosted BioNeMo \ac{nim} endpoint for \texttt{mit/boltz2}, at \texttt{health.api.nvidia.com/\allowbreak v1/\allowbreak biology/\allowbreak mit/\allowbreak boltz2/\allowbreak predict}.
Because the service is hosted, no container image tag or \ac{nim} release number is available and the \ac{api} returns no version string; the \ac{nim} release serving the calls is therefore not known.
The four calls ran between 2026-08-09 13:08:13 and 2026-08-10 00:25:22~\ac{utc}, and every call recorded \texttt{recycling\_steps} 3, \texttt{sampling\_steps} 50, \texttt{sampling\_steps\_affinity} 50, \texttt{diffusion\_samples} 1, \texttt{step\_scale} 1.638 and no molecular-weight correction; no random seed, model checkpoint identifier or code commit was recorded.

The TBXT challenge compounds were selected and scored outside \farm{}, with a conventional workflow that used Rowan for Boltz-2.
Thirty-one of the 39 values come from Rowan co-folding jobs started on 24 and 26~October 2025; the other eight are vendor analogues substituted at procurement and were scored in a later Rowan batch on 10~September 2026 (rowan-python 2.1.1, stjames 0.0.102), together with three control compounds.
Both Rowan batches used the same protein input (the 172-residue sequence of \ac{pdb} 7ZKF chain A), the compound \ac{smiles}, the \ac{msa} server and no potentials; neither recorded a backend model version or seed.

For the plotted affinity scale, the regression value reported as $\log_{10}(\mathrm{IC}_{50}/\mu\mathrm{M})$ was converted as $\mathrm{pIC}_{50}=6-\texttt{affinity\_pred\_value}$.
The pipeline file also retains a backend-native field named \texttt{predicted\_pic50}, which uses a different transformation and is not used for the reported scale.
That transformation is $\texttt{predicted\_pic50}=1.364\,(6-\texttt{affinity\_pred\_value})$, which holds to numerical precision for all 34 agent-selected TBXT compounds and for all ten AmpC shortlist predictions (\cref{sec:si-ampc-boltz2}); the two quantities therefore differ by a factor of 1.364 in their deviation from 6 and must not be pooled or compared across runs that report different fields.

The two-sided Mann--Whitney comparison used the 34 agent-selected and 39 purchased-compound docking scores (\cref{tab:si-tbxt-docking-comparison}).
The reported statistics are unchanged to two decimals when ties and a continuity correction are included in the asymptotic calculation.

\begin{table}[htbp]
      \centering
      \small
      \caption{\textbf{Statistical comparison of TBXT docking scores.}
            Agent-selected and purchased compounds were docked against \acs*{pdb} 7ZKF using the same protocol.
            The docking-score distributions were not significantly different, despite small differences in their best and median scores.}
      \label{tab:si-tbxt-docking-comparison}
      \begin{tabular}{lr}
            \toprule
            \textbf{Measure} & \textbf{Value} \\
            \midrule
            Agent-selected compounds & 34 \\
            Purchased compounds & 39 \\
            Mann--Whitney $U$ & 561.5 \\
            Two-sided $p$ & 0.26 \\
            \bottomrule
      \end{tabular}
\end{table}

All descriptor and chemical-space comparisons are descriptive.
The absence of shared scaffolds was established before blinding from the two structure sets; it cannot be independently recomputed from the released purchased-compound table.

\subsubsection{Boltz-2 scoring of the purchased set, and a calibration caveat}
\label{sec:si-boltz-runs}

The purchased (TBXT Challenge) and agent-selected compounds were scored by different services at different times, so their Boltz-2 values are not on a demonstrably common scale and the comparison is descriptive.
Of the 39 purchased compounds, 31 carry values from the original scoring batch; the other eight are vendor analogues substituted at procurement, after that batch, and were scored later.
Three of the 31 with known original values (Matter-0031, Matter-0034 and Matter-0020) were rescored alongside the eight and none reproduced: 4.644, 5.427 and 6.222 became 4.534, 5.108 and 5.396 ($-0.11$, $-0.32$ and $-0.83$, growing with the value), compressing the controls' range from 1.58 to 0.86 log units.
The eight later values (median 5.43, SD 0.30, range 4.84--5.71) resemble the 31 original values (median 5.43, SD 0.30, range 4.64--6.22), so the shift is visible only in the controls and cannot be read off the eight values themselves.
This is consistent with a change in model or service between batches, but replicates intended to separate drift from run-to-run variation were stopped by the provider, so the model's variance is unmeasured.
We therefore did not rescale the eight values, which are treated as a separate, descriptive batch.
No control was scored on both services, so comparability is untested between the purchased and agent-selected sets and fails within the Rowan batches.

\subsubsection{Complete instruction supplied to \texorpdfstring{\farm{}}{Fármaco}}

The following is the supplied initial natural-language instruction.
The campaign was executed over two consecutive days, with provenance recorded in the \farmacoatlas{}.
A follow-up prompt directed the agent to resume the workflow; the released artifacts contain the full conversations for both execution segments.

\definecolor{tbxtpromptbg}{HTML}{F7FAFC}
\definecolor{tbxtpromptrule}{HTML}{6B4A96}
\lstdefinestyle{tbxtpromptstyle}{
  basicstyle=\ttfamily\small,
  backgroundcolor=\color{tbxtpromptbg},
  frame=leftline,
  framerule=2pt,
  rulecolor=\color{tbxtpromptrule},
  framesep=8pt,
  xleftmargin=12pt,
  breaklines=true,
  breakindent=0pt,
  columns=fullflexible,
  keepspaces=true,
  showstringspaces=false
}

\begin{lstlisting}[style=tbxtpromptstyle]
Prepare the TBXT protein target for structure-based drug design. Find 7ZKf
protein structure from rcsb database for TBXT and identify the most relevant
druggable pocket for this protein.

For this target, retrieve the FASTA sequence and prefer high-quality
ligand-bound experimental structures when available. Use a justified predicted
or homologous structure when needed. Prepare the receptor, identify the most
relevant druggable pocket, and report the evidence used for sequence,
structure, and pocket selection.

Save the target-preparation outputs in their respective formats. The FASTA file
should contain the TBXT protein sequence used for structure selection or
prediction.

Use the reference ligand co-crystallised in the target pocket is provided as
`A1I1T_ref.sdf`. Compute the pocket center as the centroid of that ligand's
heavy atoms and use it as `pocket_coord`.

The pocket file should include the residues lining the pocket, the pocket center
as XYZ coordinates in Angstrom, pocket volume and surface area when available, a
druggability score with the method used to compute it, recommended docking box
dimensions, and key interaction hotspots.

Generate **700-1,000 novel ligands** inside the defined pocket. Save the generated
molecules as csv file.

Filter the generated ligands using these criteria:

- Lipinski rule of five: allow at most one violation
- SA score <= 4.5
- QED >= 0.5
- TPSA <= 140
- RotBonds <= 10
- PAINS: discard any class A, B, or C structural-alert hit
- Brenk: discard any reactive or unstable group
- hERG: discard high-risk predictions
- AMES: discard positive predictions
- Hepatotoxicity: soft flag positive predictions
- CYP3A4 inhibition: soft flag strong inhibition predictions
- BBB penetration: soft flag any positive prediction
- Oral bioavailability: soft flag values below 0.3

Save the ligands that pass the hard filters as `TBXT_filtered.csv`. Include the
canonical SMILES, descriptor values, hard-filter pass/fail status, soft ADMET
flags, and any generated ligand identifiers.

After filtering, calculate binding affinity for the passing ligands. Cluster the
passing ligands into 20 clusters, pick the two ligands with the best affinity
score from each cluster, and save the selected compounds as csv.

Map these selected compounds onto purchasable vendor ligands from ChemDiv and
Enamine using a similarity cutoff of 0.7 and a cap of 50 hits per query. Keep
Enamine and ChemDiv results in separate output files per vendor.
\end{lstlisting}

The agent resolved the request for 700--1,000 molecules as 100 samples per sampling call and invoked ten calls, eight of which registered ligands, collectively producing 705 ligand records.
Of these, 693 had valid descriptors and 656 were unique among the descriptor-valid structures.
The hard-filter cascade retained 59 compounds.
The requested 20 clusters with two representatives each yielded 34 selected compounds because six clusters contained only one retained member.

\subsubsection{Published reference ligands}
\label{sec:si-reference-ligands}

The five published reference ligands are the only compounds in this case study with reported experimental affinities, and they are included so that the two computationally selected sets can be read against a measured baseline rather than only against each other.
They are taken from the study that first described ligand binding at pocket A$'$ \cite{newmanStructuralInsightsHuman2025}, and the apparent dissociation constants in \cref{tab:si-reference} are the \ac{spr} values reported there.
Their structures are public and are released in the artifact package.
Docking scores were obtained with the same common protocol applied to the agent-selected and purchased sets.

These five compounds provide a limited assessment of how reliably docking scores rank measured affinity.
Across the full series, docking score and affinity show positive but statistically non-significant rank correlations (Spearman $\rho=0.67$, $p=0.22$ for 7ZKF; $\rho=0.82$, $p=0.09$ for 7ZK2).
However, these correlations are driven largely by M11, which is approximately 14--43 times weaker than the other compounds and also receives the worst docking score.

Among the four compounds with comparable potency, docking does not recover the experimental ordering.
M10 has the strongest measured affinity ($8\pm3~\mu$M) but ranks only third of five by docking score.
Conversely, M7 receives the best docking score ($-7.40$~kcal\,mol$^{-1}$) but has a measured affinity of approximately $14\pm9~\mu$M.

This result supports our use of docking scores as coarse ranking signals rather than quantitative estimates of affinity.
It also suggests that the small differences between the agent-selected and purchased distributions should not be interpreted as meaningful.
The best scores were $-8.5$ and $-8.4$~kcal\,mol$^{-1}$ for the agent-selected and purchased sets, respectively, while their corresponding medians were $-7.3$ and $-7.5$~kcal\,mol$^{-1}$.
Overall, these computational scores neither establish binding nor predict potency.
\begin{table}
      \centering
      \small
      \caption{\textbf{Published reference ligands for TBXT pocket A$'$.}
      Apparent dissociation constants $K_\mathrm{d}$ are the \acs*{spr} values reported by Newman \textit{et al.} \cite{newmanStructuralInsightsHuman2025}, given as value $\pm$ standard error.
      Docking scores were calculated here under the common retrospective protocol; lower values indicate more favorable predicted binding.
      These are the only compounds in this case study with experimental measurements.}
      \label{tab:si-reference}
      \begin{tabular}{lrrr}
            \toprule
            \textbf{Compound} & \textbf{\acs*{spr} $K_\mathrm{d}$ ($\mu$M)} & \textbf{Vina 7ZKF (kcal\,mol$^{-1}$)} & \textbf{Vina 7ZK2 (kcal\,mol$^{-1}$)} \\
            \midrule
            M3  & $14 \pm 9$  & $-7.10$ & $-5.90$ \\
            M6  & $25 \pm 7$  & $-6.30$ & $-6.00$ \\
            M7  & $14 \pm 9$  & $-7.40$ & $-6.20$ \\
            M10 & $8 \pm 3$   & $-6.50$ & $-6.40$ \\
            M11 & $340 \pm 8$ & $-6.10$ & $-5.80$ \\
            \bottomrule
      \end{tabular}
\end{table}

\subsubsection{Execution provenance}
\label{sec:si-tbxt-run-info}

\farmacoatlas{} records each tool invocation with its parameters and a timestamp, and links every ligand to the sampling run that produced it.
The run table covers the generation, filtering, scoring and sourcing stages, comprising calls to PocketXMol, RDKit, the structural-alert screen, \acs*{admet}-\acs*{ai}, Boltz-2 and the supplier lookup.
Structure retrieval, receptor preparation and pocket detection are not registered as runs, so the earlier part of the workflow is evidenced by its output artifacts rather than by database records; these are described below.

Identifiers are assigned at three levels, each identifying something different.
The archived execution workspaces are named by the identifiers \texttt{eval-35178c3d} for segment~A and \texttt{eval-859d8652} for segment~B, both under the same username; these identify the retained artifact directories and are the identifiers under which the run is archived.
Within the database, each of the 23 tool invocations carries an integer run identifier, and each persisted selection carries a subset identifier such as \texttt{sub\_9cffe7b9}; these are unique within the database file rather than globally.
Each PocketXMol sampling call additionally records its own identifier, \texttt{run\_a577d2db} through \texttt{run\_e77cf458}, which appears in the stored path of every structure that call produced, so an individual molecule can be traced to the specific sampling call that generated it.
The schema does not define a single session-level identifier spanning the whole campaign, and we do not report one; the workspace identifiers above are the closest equivalent and are what should be cited when referring to this archived run.
The values below are taken from that database rather than from file modification times, which reflect when the archive was written out and are not a reliable record of when the work ran.
The complete trace is given in \cref{tab:si-trace} and is included in the released package, where it contains no compound structures and can be redistributed without restriction.
Additionally, PocketXMol tools have been deployed on a \ac{gpu} cluster, so the time recorded here is total run time that includes deploying and sampling PocketXMol.

The campaign ran from 2026-08-09 03:58:36 to 2026-08-10 01:17:01 \ac{utc}, an elapsed span of 21 hours and 18 minutes, across 23 tool invocations.
Within the recorded stages the order is as expected: ten PocketXMol sampling runs, then RDKit descriptor calculation, structural-alert screening and \acs*{admet}-\acs*{ai} prediction, then four Boltz-2 batches, then the two supplier-catalogue searches.
Target preparation and pocket detection precede all of these and are covered separately below.
Because every generated ligand carries a reference to its sampling run, the 705 stage-one records can be traced individually back to raw generator output.
We verified this chain independently by re-parsing the raw structure files: the 705 files written by the eight productive sampling runs map without exception onto the stage-one table, and the twelve files that RDKit cannot parse correspond exactly to the twelve records flagged as having invalid descriptors.

Target preparation and pocket detection left their own artifacts.
P2Rank ran once, reporting version 2.5.1 with a full parameter dump, and placed its rank-1 pocket at $(-7.50, -41.63, -3.81)$ with probability 0.015, distant from the ligand site.
Pocketeer ran under more than one parameter setting, returning a single pocket under its default settings and 29 under a relaxed minimum-size threshold, and the pocket report identifies the selected site as the ligand-guided pocket corresponding to relaxed pocket 18, ranked 9 of 29.
The pocket centre itself was fixed by the instruction rather than by pocket detection: the prompt directed that it be computed as the centroid of the heavy atoms of the supplied reference ligand, and the resulting coordinate recorded in the pocket report, $(13.653, -22.965, -6.120)$, is the value passed to all ten sampling runs.
That agreement can be checked directly against \cref{tab:si-trace}, so although this stage has no run record, the quantity it produced is verifiable.
Its timing can be bounded but not measured: because the sampling runs consume that coordinate, target preparation and pocket detection must have completed before the first database event at 03:58:19 \ac{utc}, which is how they are entered in \cref{tab:si-trace}.

Execution halted after the generation stage and was resumed with a second prompt, distinct from the first, that instructed the agent to continue from filtering onward and explicitly not to regenerate ligands or rerun PocketXMol; both prompts are released in full alongside a second-precision transcript of every tool call between them.
The database was carried forward rather than rebuilt, which we confirmed by comparing the two copies: all 705 ligand records are identical across them, including primary keys and sampling-run references, as is the target record, and the internal subset numbering continues through the restart without resetting.
Provenance is therefore continuous across the restart.
The idle interval between the generation and analysis stages was 8 hours and 16 minutes, comparable to intervals occurring within a single uninterrupted stretch of the campaign, such as the 5 hours and 46 minutes separating two consecutive Boltz-2 batches.
Long waits between queued jobs are characteristic of this workload.

Accordingly, the campaign ran from two prompts, both released in full, with no human decision about chemistry at any stage: the second prompt directed which stage to resume from the first prompt already set.

\begingroup
      \small
      \renewcommand{\arraystretch}{1.15}
      \setlength{\tabcolsep}{3pt}
      \setlength{\LTcapwidth}{\linewidth}
      \begin{longtable}{@{}rc>{\raggedright\arraybackslash}p{0.23\linewidth}>{\raggedright\arraybackslash}p{0.19\linewidth}>{\raggedright\arraybackslash}p{0.35\linewidth}r@{}}
      \caption{\textbf{Tool invocations for the TBXT campaign.}
      The lower block is taken from the \farmacoatlas{} run table.
      The upper block covers target preparation and pocket detection, whose timestamps were recovered from the archived chat transcript.
      The resulting pocket centre, $(13.653, -22.965, -6.120)$, was used for all sampling runs.
      Segment~A precedes the operator restart and segment~B follows it.
      The ligands column reports the number of registered ligands; a dash indicates that no new ligands were added.
      Timestamps are in \acs*{utc}.}
      \label{tab:si-trace}\\
            \toprule
            \textbf{Run} & \textbf{Seg.} & \textbf{Tool} & \textbf{Timestamp} & \textbf{Key parameters} & \textbf{Ligands} \\
            \midrule
      \endfirsthead
            \multicolumn{6}{l}{\textbf{Table~\thetable{} (continued)}} \\
            \toprule
            \textbf{Run} & \textbf{Seg.} & \textbf{Tool} & \textbf{Timestamp} & \textbf{Key parameters} & \textbf{Ligands} \\
            \midrule
      \endhead
            \midrule
            \multicolumn{6}{r}{\textit{Continued on next page}} \\
      \endfoot
            \bottomrule
      \endlastfoot
            \multicolumn{6}{l}{\textit{Target preparation and pocket detection: evidenced by output artifacts}} \\
            \addlinespace[2pt]
            -- & A & Structure input & 2026-08-09 03:55:59 & \texttt{7ZKF.fasta}, \texttt{A1I1T\_ref.sdf} & -- \\
            -- & A & Target registration & 2026-08-09 03:56:07 & \texttt{db\_get\_or\_create\_target} & -- \\
            -- & A & P2Rank pocket prediction & 2026-08-09 03:56:13 & version 2.5.1; rank-1 $p=0.015$, distal & -- \\
            -- & A & Pocketeer pocket detection & 2026-08-09 03:56:13 & 3 calls; default: 1 pocket, relaxed: 29 pockets & -- \\
            -- & A & Pocket definition & 2026-08-09 03:57:49 & centroid computed: $(13.653, -22.965, -6.120)$ & -- \\
            -- & A & Receptor registration & 2026-08-09 03:57:58 & \texttt{7ZKF\_receptor.pdb} prepared receptor & -- \\
            \addlinespace[3pt]
            \midrule
            \multicolumn{6}{l}{\textit{Recorded tool invocations}} \\
            \addlinespace[2pt]
            1 & A & PocketXMol sampling & 2026-08-09 03:58:36 & num\_mols 100, r 11.0, out batch1 & -- \\
            2 & A & PocketXMol sampling & 2026-08-09 03:59:33 & num\_mols 100, r 11.0, out batch1 & 92 \\
            3 & A & PocketXMol sampling & 2026-08-09 04:02:48 & num\_mols 100, r 11.0, out batch2 & 85 \\
            4 & A & PocketXMol sampling & 2026-08-09 04:06:07 & num\_mols 100, r 11.0, out batch3 & 91 \\
            5 & A & PocketXMol sampling & 2026-08-09 04:09:24 & num\_mols 100, r 11.0, out batch4 & 90 \\
            6 & A & PocketXMol sampling & 2026-08-09 04:21:10 & num\_mols 100, r 11.0, out batch6 & 83 \\
            7 & A & PocketXMol sampling & 2026-08-09 04:25:34 & num\_mols 100, r 11.0, out batch5 & 90 \\
            8 & A & PocketXMol sampling & 2026-08-09 04:29:41 & num\_mols 100, r 11.0, out batch7 & 86 \\
            9 & A & PocketXMol sampling & 2026-08-09 04:36:17 & num\_mols 100, r 11.0, out batch8 & 88 \\
            10 & A & PocketXMol sampling & 2026-08-09 04:40:43 & num\_mols 100, r 11.0, out batch9 & -- \\
            11 & B & RDKit descriptors & 2026-08-09 12:56:58 & all descriptors & -- \\
            12 & B & Structural alerts & 2026-08-09 12:57:15 & 4 alert sets & -- \\
            13 & B & \acs*{admet}-\acs*{ai} & 2026-08-09 12:58:05 & -- & -- \\
            14 & B & \acs*{admet}-\acs*{ai} & 2026-08-09 13:01:13 & 6 metrics & -- \\
            15 & B & RDKit descriptors & 2026-08-09 13:04:09 & 9 metrics & -- \\
            16 & B & RDKit descriptors & 2026-08-09 13:04:26 & -- & -- \\
            17 & B & RDKit descriptors & 2026-08-09 13:04:54 & 13 metrics & -- \\
            18 & B & Boltz-2 & 2026-08-09 13:08:13 & -- & -- \\
            19 & B & Boltz-2 & 2026-08-09 18:54:56 & without\_potentials & -- \\
            20 & B & Boltz-2 & 2026-08-09 20:40:08 & without\_potentials & -- \\
            21 & B & Boltz-2 & 2026-08-10 00:25:22 & -- & -- \\
            22 & B & Supplier lookup & 2026-08-10 01:15:43 & backend chemdiv, limit 50 & -- \\
            23 & B & Supplier lookup & 2026-08-10 01:17:01 & backend enamine, limit 50 & -- \\
      \end{longtable}
\endgroup

\clearpage
\subsection{PGK2/PGK1 adaptive selection}
\label{sec:si-pgk2}

\subsubsection{Released data, structural inputs, and execution record}

The PGK2/PGK1 case-study artifact package contains the supplied PGK2 and PGK1 sequences, receptor-provenance and preparation records, the adaptive-run instruction, a human-readable transcript, the lossless raw-message export, the \farmacoatlas{} SQLite database, normalized final-candidate tables, selected DiffDock ligand poses and Boltz-2 predicted structures, the self-contained status report, and the source data and scripts used to construct \cref{fig:pgk2,tab:pgk2}.
The PGK2/PGK1 adaptive-selection execution reported here was initiated by one top-level user instruction on September~19, 2026; the 39 database run records span 04:45:58--12:03:54 \ac{utc} by creation time.
Because the run table records creation rather than completion times, these timestamps delimit the recorded workflow sequence but do not isolate tool runtime from agent reasoning and database operations.

The structural inputs were obtained from CACHE Challenge~\#7, which provides unpublished human PGK2 bound to compound~47 and human PGK1 bound to compound~45 for the identification of PGK2-selective inhibitors~\cite{cacheChallenge7PGK2}.
The challenge page does not assign \ac{pdb} accessions to these complexes; we therefore refer to them by the CACHE working identifiers \texttt{PGK2\_cmp47} and \texttt{PGK1\_cmp45}.
The official coordinate and reflection packages contain PHENIX X-ray-refinement records, with reported high-resolution limits of 1.44~\AA{} for PGK2--compound~47 and 1.45~\AA{} for PGK1--compound~45.
Exact coordinate comparison established how the receptor inputs were derived from the CACHE files.
For each receptor, protein chain~A and \texttt{ATOM} records were retained, explicit hydrogen atoms were removed, and atoms with blank or \texttt{A} alternate-location identifiers were selected, with the retained alternate-location field normalized to blank.
All \texttt{HETATM} records, including the co-crystallized ligands, waters, 3-phosphoglycerate in the PGK2 structure, and sulfate in the PGK1 structure, were removed; the retained protein coordinates were otherwise unchanged.
The CACHE page is cited as the structure source, and the coordinate files are not redistributed with the manuscript artifacts.

\paragraph{Software environment.}
The agent \ac{api} environment contained RDKit 2025.09.5 with Python 3.11.15, \ac{plip} 3.0.0, and OpenBabel 3.1.1.
PocketXMol used commit \texttt{69e53681} with PyTorch 2.0.0+cu117 and RDKit 2023.9.3; its model weights were obtained from Zenodo record \texttt{17801271}.
\acs*{admet}-\acs*{ai} used \texttt{admet\_ai} 2.0.1, Chemprop 2.2.4, PyTorch 2.13.0, and RDKit 2026.3.4.
DiffDock, Boltz-2, and the ColabFold \ac{msa} search were accessed as hosted NVIDIA \ac{nim} services; their exact runtime versions were not retained by the clients.

\paragraph{Resolved parameters and stochasticity.}
DiffDock returned one pose per request (\texttt{num\_poses=1}) with \texttt{time\_divisions=20} and \texttt{steps=18}.
The Boltz-2 requests used \texttt{recycling\_steps=3}, \texttt{sampling\_steps=50}, \texttt{diffusion\_samples=1}, \texttt{step\_scale=1.638}, \texttt{sampling\_steps\_affinity=50}, and \texttt{diffusion\_samples\_affinity=1}, with \texttt{without\_potentials=false}, \texttt{concatenate\_msas=false}, and \texttt{affinity\_mw\_correction=false}.
No templates or pocket-coordinate constraints were supplied to Boltz-2.
The ColabFold \ac{msa} search used \texttt{databases=all}, an E-value cutoff of $10^{-4}$, and a maximum of 500 sequences; the returned alignments contained 217 sequences for each target and used UniRef30\_2302, colabfold\_envdb\_202108, and PDB70\_220313.
PocketXMol did not expose a user-settable seed through the \farm{} tool; the effective seeds recorded for the four generation calls were 8849, 8690, 8796, and 8758.
DiffDock and Boltz-2 accepted no seed field in the submitted requests and were controlled by the hosted services.
\acs*{admet}-\acs*{ai} exposed no seed setting; repeated inference for the checked ligand reproduced the stored endpoint values exactly.

\subsubsection{Stage 1: generation, filtering, docking, and contact gate}

The adaptive instruction specified repeated generation and selection until five Stage~2 passers were obtained or a stopping condition was reached.
Four PocketXMol calls used the PGK2 pocket center $(52.42,43.78,25.95)$~\AA{} and an 8~\AA{} radius, registering 87 and 94 molecules in cycle~1 and 95 and 93 molecules in cycle~2, for 369 molecules in total.
Each registered molecule was canonicalized without tautomer enumeration and evaluated using \ac{pains} catalogues A--C, \ac{qed} $\geq 0.6$, \ac{sa} score $\leq 5$, and an extended Lipinski rule-of-five requiring at least four of five limits: molecular weight $\leq 500$, cLogP $\leq 5$, \ac{hbd} $\leq 5$, \ac{hba} $\leq 10$, and rotatable bonds $\leq 10$.
The persisted hard-filter subsets contained 19 and 15 molecules in cycle~1 and 23 and 19 molecules in cycle~2, yielding 76 passers overall.

The 76 hard-filter passers were docked against PGK2.
For each ligand, the single returned DiffDock pose was merged with the receptor and analyzed by \ac{plip}.
A Stage~1 pass required a recorded hydrogen bond or hydrophobic interaction with PGK2 TYR242; a halogen-only contact was not an accepted gate type, and no manual interaction override entered the reported selection.
The persisted Stage~1 subsets contained 28 ligands from cycle~1 and 30 from cycle~2, yielding 58 ligands for Stage~2.
Thus, the same explicit filter and interaction rules were applied to all four generated batches, and the intermediate subsets and structures remained linked to their originating run records in \farmacoatlas{}.

\subsubsection{Stage 2: target--counter-target comparison and candidate selection}

The 28 and 30 Stage~1 passers from the two cycles were evaluated separately against PGK2 and PGK1 by Boltz-2 co-folding, and PGK1 was additionally evaluated by DiffDock.
The instruction supplied a mapped PGK1 pocket center, whereas the archived DiffDock request records contain \texttt{num\_poses=1} but no pocket-coordinate field.
Boltz-2 was run without a template or pocket constraint, and its PGK2 and PGK1 outputs were retained together with the corresponding confidence, \ac{plddt}, \texttt{binding\_probability}, and affinity fields.

The archive also retains the prompt-requested ligand \ac{rmsd} fields generated by optimal ligand superposition.
A post-run audit found that these calculations did not constitute an independent DiffDock--Boltz-2 pose comparison, so the values are retained as provenance rather than used as structural evidence or selection gates in the manuscript.
All five model-score-threshold passers also had archived numerical \ac{rmsd} values below the 3.5~\AA{} prompt threshold.
The prespecified model-scale selection rule reported in the manuscript was
$\Delta\texttt{predicted\_pic50}=\texttt{predicted\_pic50}_{\mathrm{PGK2}}-\texttt{predicted\_pic50}_{\mathrm{PGK1}}\geq 0.477$.
The PGK2 and PGK1 scores are the archived \texttt{predicted\_pic50} fields and are treated here as model-output scores.
For all 116 archived predictions, these scores equal $1.364\,(6-\texttt{binding\_affinity})$, differing from the normalized \(\mathrm{pIC}_{50}\) convention $6-\texttt{binding\_affinity}$.
The 0.477 threshold was applied on the recorded Boltz-2 score scale and is not interpreted as a threefold affinity difference.

Five ligands (26, 78, 160, 351, and 369) satisfied the model-score threshold, with differences of 0.501--1.076; three were obtained in cycle~1 and two in cycle~2.
Their Stage~1 PGK2 poses contained TYR242 hydrogen bonds for ligands~78, 160, and 369 and hydrophobic contacts for ligands~26 and 351.
The recorded PGK2 \texttt{binding\_probability} values ranged from 0.172 to 0.371 and were reported but not used as a gate.
\acs*{admet}-\acs*{ai} was subsequently run once for these five ligands only.
The resulting target--counter-target and \ac{admet} values are summarized in \cref{tab:pgk2}; the adaptive counts and score comparison are shown in \cref{fig:pgk2}.

\subsubsection{Instruction supplied to \texorpdfstring{\farm{}}{Fármaco}}

The following substantive natural-language instruction generated the adaptive-selection execution analyzed here.
It was submitted as one top-level user message; the equality signs shown here are retained from the source prompt.

\definecolor{pgkpromptbg}{HTML}{F7FAFC}
\definecolor{pgkpromptrule}{HTML}{2C7F77}
\lstdefinestyle{pgkpromptstyle}{
  basicstyle=\ttfamily\small,
  backgroundcolor=\color{pgkpromptbg},
  frame=leftline,
  framerule=2pt,
  rulecolor=\color{pgkpromptrule},
  framesep=8pt,
  xleftmargin=12pt,
  breaklines=true,
  breakindent=0pt,
  columns=fullflexible,
  keepspaces=true,
  showstringspaces=false
}

\begin{lstlisting}[style=pgkpromptstyle]

Set up two protein targets using the uploaded files: PGK2
(PGK2_cmp47.fasta, PGK2_cmp47_receptor.pdb, chain A, human PGK2) and
PGK1 (PGK1_cmp45.fasta, PGK1_cmp45_receptor.pdb, chain A, human PGK1).

=== STAGE 1 (per molecule, PGK2 only) ===
1. Generate molecules with pocketxmol_sample against PGK2, pocket_coord
   [52.42, 43.78, 25.95], radius 8.0 (the co-crystal pocket).
2. Normalize/canonicalize each generated molecule. No tautomer enumeration.
3. Compute descriptors (MolWt, MolLogP, TPSA, NumHDonors,
   NumHAcceptors, NumRotatableBonds, FormalCharge, QED, SA_Score).
4. Apply hard filters: no PAINS structural alerts (catalogs A, B, C) /
   QED >= 0.6 / SA_Score <= 5 / extended Lipinski Ro5 (MW<=500,
   cLogP<=5, HBD<=5, HBA<=10, rotatable bonds<=10 -- satisfy at least
   4 of 5).
5. For molecules passing step 4, run diffdock_predict against PGK2.
6. Build the protein-ligand complex from the rank-1 DiffDock pose only,
   using merge_protein_ligand_structure.
7. Run plip_analyze_complex on the rank-1 complex. A molecule passes
   the Stage 1 gate if PLIP reports an interaction (hydrogen bond,
   ideally via the phenolic hydroxyl, or hydrophobic contact) with
   residue TYR242 on PGK2.

=== STAGE 2 (per Stage-1-passing molecule) ===
1. Run boltz2_predict co-folding the molecule with PGK2.
2. Convert the Boltz-2 mmCIF output to PDB with convert_mmcif_to_pdb.
3. Confirm the Boltz-2 predicted pose independently agrees with the
   DiffDock rank-1 pose: compute the ligand RMSD between the two PGK2
   poses. Gate: RMSD < 3.5 A.
4. Run diffdock_predict against PGK1, pocket_coord
   [26.09, 18.89, 36.95], radius 8.0 (the PGK2 co-crystal pocket,
   mapped onto PGK1 by structural alignment -- PGK1 has no co-crystal
   ligand of its own). Rank-1 pose only.
5. Run boltz2_predict co-folding the same molecule with PGK1. Convert
   to PDB the same way.
6. Compute the PGK1 RMSD using the identical method described in step 3.
   Gate: PGK1 RMSD < 3.5 A (this is a data-quality check, not a
   selectivity signal itself -- we only trust the PGK2-vs-PGK1
   selectivity comparison in step 8 for molecules whose binding-mode
   prediction is independently confirmed reliable on BOTH targets).
7. Record Boltz-2 binding_probability for PGK2. Do not gate on this --
   report it for reference only.
8. Selectivity gate: delta_pIC50 = predicted_pic50(PGK2) -
   predicted_pic50(PGK1). Gate: >= 0.477 (approx. 3-fold selectivity).

A molecule counts as a Stage-2-gate pass only if it clears ALL THREE
gates in steps 3, 6 and 8.

=== KEEP GOING UNTIL WE HAVE ENOUGH ===
Generate batches of about 100 molecules (up to 3 batches) until you have
about 20 that pass Stage 1, then run all of them through Stage 2 together.
Don't run ADMET yet at this point.

If you don't end up with 5 molecules passing everything in Stage 2,
repeat this whole process (a fresh batch of ~20 Stage-1 passers, then
Stage 2 on that batch) up to 2 more times (3 attempts total) -- keep
going automatically without stopping to ask. Stop once you reach 5
molecules passing everything, or after 3 attempts, or after 800 molecules
generated in total, whichever comes first. Report the total pass count,
how many batches/attempts you ran, and the pass rate at each gate. If you
stopped short of 5, ask how to proceed.

=== ONCE YOU'RE DONE ===
For the final passing molecules only: run admet_ai_predict once per
molecule.

Report, per molecule: SMILES, TYR242 interaction type and DiffDock PGK2
confidence, Stage 2 PGK2 RMSD and PGK1 RMSD, Boltz-2
confidence/pLDDT/predicted_pic50/binding_probability for both PGK2 and
PGK1, delta_pIC50, and key ADMET flags (hERG, AMES, DILI, ClinTox, HIA,
BBB). Also report: total molecules generated and the pass rate at each
gate.
\end{lstlisting}

\clearpage
\subsection{AmpC retrospective benchmark}
\label{sec:si-ampc}

\subsubsection{Data and execution record}

The AmpC artifact package contains the source \ac{csv}, user instructions, the conversation export for ten completed turns, per-turn responses and tool-call arguments, a turn index, the \farmacoatlas{} SQLite database, downloaded and generated structures, pocket-detection outputs, and normalized analysis tables.
It also contains the separate Boltz-2 affinity run described in \cref{sec:si-ampc-boltz2}, with its own conversation export, database, raw service responses, and predicted complex structures.
Scripts regenerate the numerical summaries, the reported pose \acp{rmsd}, and the generated-candidate summary figure.
The reported workflow consists of eight scripted prompts and two follow-up questions executed on September 10, 2026 in a single conversation and workspace.
The ten turns account for 2.9~h of summed turn latencies and 39 database workflow runs; these latencies include reasoning and database queries as well as tool execution.
Resource accounting includes these ten turns and the separately recorded successful Boltz-2 completion, giving eleven completed executions per configuration (\cref{sec:si-ampc-nodb}).
Other conversations and failed or empty recovery attempts are excluded from the resource totals.

The dataset contains 1{,}521 compounds and was imported as 1{,}521 ligand records with 9{,}115 metrics.
All compounds have an experimental-outcome entry: 298 numeric $K_\mathrm{i}$ values, 74 \texttt{HB} labels, and 1{,}149 \texttt{NH} labels.
The released dataset does not specify the meaning of the two categorical labels; \texttt{NH} is read here as a negative outcome, while \texttt{HB} denotes a non-quantified outcome that may be positive.
The numeric-$K_\mathrm{i}$ definition of a hit used here excludes both categorical labels from the numerator but retains all tested compounds in the denominator.
Selection-strategy labels are present for 1{,}510 compounds; the 11 unlabelled compounds are excluded from strategy comparisons.
Potency summaries describe only compounds with numeric $K_\mathrm{i}$ values.

\subsubsection{Retrospective enrichment and selection}

The numerical verification script reconstructs the dataset summaries, rank-bin enrichment, selection-strategy comparisons, selected-subset outcomes, interaction fingerprints, and \ac{admet} and generation counts from the source \ac{csv} and database.
Its programmed checks pass.
Agreement with recorded values establishes numerical reproducibility of those checks, but does not validate the interpretation of a metric or establish experimental performance of a prediction.

The rank-dependent enrichment is monotonic under the numeric-$K_\mathrm{i}$ definition of a hit, but not under the broader definition that also counts the \texttt{HB} label, because \texttt{HB} compounds concentrate in the worst-ranked bins:
\begin{center}
\begin{tabular}{lrrr}
\toprule
\textbf{Global rank} & \textbf{Numeric $K_\mathrm{i}$} & \texttt{HB} & \textbf{Non-}\texttt{NH} \\
\midrule
$\le 10^3$ & 37/75 (49.3\,\%) & 2 & 52.0\,\% \\
$10^3$--$10^4$ & 139/375 (37.1\,\%) & 9 & 39.5\,\% \\
$10^4$--$10^5$ & 110/628 (17.5\,\%) & 32 & 22.6\,\% \\
$10^5$--$10^6$ & 4/66 (6.1\,\%) & 7 & 16.7\,\% \\
$10^6$--$10^7$ & 4/143 (2.8\,\%) & 6 & 7.0\,\% \\
$10^7$--$10^8$ & 3/154 (1.9\,\%) & 8 & 7.1\,\% \\
$>10^8$ & 1/80 (1.25\,\%) & 10 & 13.8\,\% \\
\bottomrule
\end{tabular}
\end{center}
Rank bins have inclusive upper bounds, and the non-\texttt{NH} column counts numeric $K_\mathrm{i}$ and \texttt{HB} outcomes together.
The agent reported that the trend is unchanged when \texttt{HB} is counted as a positive, but that comparison used coarser rank quintiles (43.9\,\% to 9.2\,\% from best to worst), which average over the reversal in the two worst bins above.
The reported enrichment therefore depends on the choice of hit definition, which the released dataset does not resolve.

The 450 tested compounds within the campaign's top 10{,}000 global ranks contained 176 of the 298 hits (59.1\,\%).
Across the whole tested set, the three selection categories gave hit rates of 45/174 (25.9\,\%) for \texttt{manual}, 149/776 (19.2\,\%) for \texttt{auto}, and 104/560 (18.6\,\%) for \texttt{auto/manual}, with median $K_\mathrm{i}$ values among quantified hits of 61.0, 172.6, and 139.1~\textmu{}M respectively.
In addition to this overall strategy comparison, the agent compared outcomes within three broad rank strata:
\begin{center}
\begin{tabular}{lrrr}
\toprule
\textbf{Global rank} & \textbf{Manual} & \textbf{Automatic} & \textbf{Joint} \\
\midrule
$\le 10^4$ & 23/47 (48.9\,\%) & 98/258 (38.0\,\%) & 55/145 (37.9\,\%) \\
$10^4$--$10^5$ & 21/81 (25.9\,\%) & 45/253 (17.8\,\%) & 44/290 (15.2\,\%) \\
$>10^5$ & 1/46 (2.2\,\%) & 6/265 (2.3\,\%) & 5/125 (4.0\,\%) \\
\bottomrule
\end{tabular}
\end{center}
Entries are numeric-$K_\mathrm{i}$ hits divided by tested compounds.
``Joint'' denotes the dataset label \texttt{auto/manual}; rank strata have inclusive upper bounds.
These descriptive comparisons do not adjust for differences in chemistry or rank distributions within each stratum.

The prompt described the top-50 selection as blinded, but the experimental outcomes remained available in the database and had already been analyzed in the same conversation.
The executed selection query ordered the imported compounds by ascending global rank, limited the result to 50, and did not reference activity fields.
The agent persisted the subset, generated docked structures, and then queried its experimental outcomes.
The selected subset contained 26 hits (52.0\,\%), including three with $K_\mathrm{i}\le 10$~\textmu{}M, compared with 298 hits among all 1{,}521 tested compounds (19.6\,\%).
The exercise therefore demonstrates execution of a rank-based selection rule, not independent blinded prediction.
All reported hit rates describe the original authors' experimentally tested subset and do not estimate performance across the full 1.7-billion-molecule library.

\subsubsection{Pocket detection and structural comparison}

The receptor was prepared by downloading \ac{pdb} 1L2S and cleaning the structure during the run; identity with the original study's complete receptor-preparation protocol was not established.
P2Rank and Pocketeer were called through a pocket-analysis sub-agent, with results recorded in the conversation and output files rather than database workflow records.
On the prepared receptor, P2Rank assigned scores of 26.96 and 20.15 and probabilities of 0.918 and 0.850 to the chain-B and chain-A catalytic clefts, respectively.
Both pockets contain Ser64, Tyr150, Asn152, Lys315, Thr316, Gly317, and Asn346.
Pocketeer also detected the catalytic region but ranked an inter-chain cavity first.

DiffDock generated five poses per ligand for the top-50 subset and ten per ligand for the four co-crystal ligands.
The latter were docked into the 1L2S-derived receptor; the co-crystal coordinates were used for validation.
The mapping between catalogue identifiers and \ac{pdb} entries agrees with the source study's data-availability statement \cite{liuImpactLibrarySize2025}: Z6615020275/9C84, Z6615017509/9C6P, Z6615017782/9DHL, and Z4462773688/9C81.

The agent calculated ligand \acp{rmsd} in an interactive Python tool.
The archived recomputation script superposes each co-crystal chain onto the corresponding receptor chain using shared C$\alpha$ atoms and Kabsch alignment, then compares ligand coordinates using an RDKit \ac{mcs}.
It selects the lowest \ac{rmsd} among the enumerated mappings and the two chains, requiring coverage of at least 80\,\% of docked heavy atoms.
The substructure search uses unique matches with a maximum of 100 per molecule; it does not guarantee enumeration of all symmetry-equivalent atom mappings.
Crystal ligands were reconstructed from \ac{pdb} coordinates without restoring bond orders from a chemical template.
Consequently, the \ac{mcs} covers 22/24, 21/23, 21/24, and 22/24 heavy atoms for 9C84, 9C6P, 9DHL, and 9C81, respectively.
The resulting rank-1 \acp{rmsd} reproduce as 0.87, 3.27, 2.76, and 5.55~\AA{} to the reported precision.
Only 9C84 had a rank-1 pose within 2~\AA{} of its crystal reference; considering all ten poses per ligand did not change which ligand met this threshold (\cref{tab:ampc}).
These are partial-heavy-atom \ac{mcs} comparisons, not standardized full-ligand symmetry-corrected \acp{rmsd}.

PoseBusters \cite{buttenschoenPoseBustersAIbasedDocking2024} was run on all 40 co-crystal-ligand poses using its dock configuration.
Each rank-1 result records 25 of 30 checks as passing.
The failed entries include missing-reference or cofactor/water-related checks as well as minimum-distance-to-protein checks; the aggregate pass fraction should therefore not be interpreted as a fraction of physically valid poses.

\Ac{plip} \cite{schakePLIP2025Introducing2025} was applied to the experimental complexes and to complexes formed by merging each rank-1 docked ligand with the prepared receptor.
Crystal fingerprints are unions of contacts from both chains; chain identity is discarded when forming residue/type and residue-only sets.
The docked complexes contain one ligand copy, so the reported Jaccard values compare that ligand's contacts with an aggregate crystal reference.
For example, restricting the crystal reference to the corresponding chain changes the residue/type Jaccard from 0.56 to 0.83 for 9C84 and from 0.42 to 0.56 for 9C6P.
Crystal fingerprints also include water-mediated contacts, whereas the prepared docking complexes lack those waters.
Docked poses frequently contacted Ser64, Lys67, Asn152, and Ala318.
Recurring docked contacts are not necessarily recovered experimental contacts: the docked 9C81 ligand contacts Ser64 and Lys67, which are absent from its crystal fingerprint.

\begin{table}
      \centering
      \small
      \caption{\textbf{Pose and interaction recovery for the four AmpC co-crystal ligands.}
      Ligands were docked into a receptor prepared from \acs*{pdb} 1L2S; the co-crystal structures served as validation references.
      \acsp*{rmsd} were computed over the maximum common substructure after protein superposition, covering 21--22 of 23--24 ligand heavy atoms, and reproduced to the reported precision.
      Jaccard indices compare \acs*{plip} contacts of the rank-1 docked pose with the union of contacts from both crystal chains, using residue-and-interaction-type or residue alone and omitting chain identity.
      $K_\mathrm{i}$ denotes the inhibition constant; Best of 10 is the lowest ligand \acs*{rmsd} among ten docked poses.
      Type Jaccard compares residue-and-interaction-type contacts, whereas Residue Jaccard compares contacted residues.
      See \cref{sec:si-ampc} for mapping and reference-selection details.}
      \label{tab:ampc}
      \begin{tabular}{lrrrrrr}
            \toprule
            \textbf{Co-crystal} & \textbf{$K_\mathrm{i}$ ($\mu$M)} & \textbf{Original rank} & \textbf{Rank-1 \acs*{rmsd} (\AA{})} & \textbf{Best of 10 (\AA{})} & \textbf{Type Jaccard} & \textbf{Residue Jaccard} \\
            \midrule
            9C84                & 1.31                                 & 6{,}607                & 0.87                         & 0.87                        & 0.56                  & 0.56                     \\
            9C6P                & 0.86                                 & 2{,}216                & 3.27                         & 3.06                        & 0.42                  & 0.50                     \\
            9DHL                & 0.95                                 & 8{,}574                & 2.76                         & 2.57                        & 0.36                  & 0.64                     \\
            9C81                & 322.96                               & 1{,}544{,}474          & 5.55                         & 5.55                        & 0.21                  & 0.23                     \\
            \bottomrule
      \end{tabular}
\end{table}

\subsubsection{\acs*{admet} triage and generative follow-up}
\label{sec:si-ampc-generation}

\acs*{admet}-\acs*{ai} predictions, RDKit descriptors, \ac{pains}/Brenk alerts, and catalogue matches were used to prioritize the 298 numeric-$K_\mathrm{i}$ hits.
The 71 reported Enamine matches come from an exact query against \texttt{REALDB-25Q3-9.4B}.
A missing match in that query does not establish that a compound is unavailable; the original study states that its tested compounds could be purchased from Enamine \cite{liuImpactLibrarySize2025}.
Predicted \ac{dili} scores were $\ge 0.8$ for 241 hits and $<0.5$ for only four.
None of the nine compounds combining $K_\mathrm{i}\le 50$~\textmu{}M, a clean alert profile, and an exact catalogue match had a predicted DILI score below 0.5.
\Ac{admet} scores are model predictions and were not calibrated or experimentally validated for these compounds.
The reported AMES and ClinTox values were retrieved from complete result payloads.
For the active-hit \ac{admet} run, the requested name \texttt{Ames} did not match the returned name \texttt{AMES}, and ClinTox was not requested, so these values are absent from that run's metric rows despite being present in its result payloads.

Nine hits with $K_\mathrm{i}\le 2$~\textmu{}M were submitted as MolMIM seeds with requested parameters of 20 molecules per seed and minimum similarity 0.6.
Four seed runs contributed 20 molecules each to the 80-molecule set.
The returned \texttt{molmim\_score} is labelled \ac{qed} in the successful result payloads and equals the independently stored RDKit \ac{qed} for all 80 molecules.
The agent nevertheless aliased this field as similarity in its queries and used it as a similarity threshold and ranking contribution; one composite score also included a separate \ac{qed} contribution.
Thus, the numerical values reproduce, but their interpretation as seed similarity is incorrect and the reported MolMIM shortlist is not evidence of successful similarity-based prioritization.
The requested generation constraint and the returned \ac{qed} score must be distinguished.

The four methods shared property, structural-alert, and \ac{admet} assessments, but their prioritization rules were not identical.
Shape/feature comparison to the 9C6P ligand conformer and exact supplier lookup were additionally performed for PocketXMol and PrexSyn.
The combined property window used in the generation summaries was molecular weight 250--450, logP 0--4, and \ac{tpsa} 50--140~\AA$^2$; none of the 25 GenMol molecules met all three criteria.
Of 100 requested PocketXMol samples, 85 were marked successful, two incomplete, and 13 bad; the reported downstream counts include both successful and incomplete samples.
PrexSyn returned 79 hits across nine seeds, including the seed itself as the top hit for each query; the agent retained 20 non-identical analogues for subsequent analysis.
The counts with predicted DILI below 0.75 were 56/80 for MolMIM, 24/25 for GenMol, 23/87 for PocketXMol, and 0/20 for the retained PrexSyn set.
Some PocketXMol candidates had lower predicted DILI scores than the retained PrexSyn analogues, but generally showed weaker feature similarity to the 9C6P reference ligand.
The screening columns in \cref{fig:ampc}a apply each criterion independently to the complete analyzed set for each method.
Cell values give counts, and color encodes within-set percentages.
The alert screen requires a clean PAINS/Brenk profile; the other columns use the combined property window, predicted logS $\ge -4$, hERG $<0.5$, DILI $<0.75$, and Ames $<0.5$.
The set sizes and selection rules differ across methods.

PocketXMol used a sampling centre of $(23.5,5.5,14.5)$~\AA{} and a radius of 10~\AA{}.
This centre is 6.9~\AA{} from the chain-A catalytic-residue centroid and 8.6~\AA{} from the P2Rank pocket centre.
Both reference points lie within the sampling sphere, but the effect of the offset on generated molecules was not tested.
No generated compound was synthesized or assayed, and predicted \ac{admet} properties, shape/feature similarity, and structural-alert status do not establish AmpC activity.

\subsubsection{Boltz-2 affinity of the generated shortlist}
\label{sec:si-ampc-boltz2}

A shortlist of ten generated compounds---six MolMIM molecules, two PrexSyn analogues, and two PocketXMol samples---was persisted as a subset in the recorded conversation and submitted to Boltz-2 for co-folding and affinity prediction against \ac{pdb} 1L2S.
The predictions reported here were produced on September 18, 2026 in a separate conversation, in a fresh workspace and a fresh database, with the ten structures supplied in the prompt as explicit \ac{smiles} strings.
The verification script confirms that these ten strings are exactly the members of the shortlist subset persisted in the recorded conversation, and that the new database contains one target, ten ligands, one workflow run, ten results, and forty metric rows and nothing else.
The run is thus a completion of the intended step rather than a continuation of the recorded workflow; it is indexed separately from the original ten turns and included in the resource totals in \cref{sec:si-ampc-nodb}.

\Cref{fig:ampc}b plots the converted affinity score $6-\texttt{affinity\_pred\_value}$ against the predicted binding probability for each shortlisted compound.
Compound labels match the identifiers in \cref{tab:ampc-boltz2}.
All ten predictions succeeded.
A single \ac{msa} of 200 sequences was computed once and reused for all ten pairs, and each prediction used one diffusion sample, one affinity sample, no molecular-weight correction, and no pocket conditioning, so every reported metric is the value of that single sample rather than an average.

\begin{table}
      \centering
      \small
      \caption{\textbf{Boltz-2 affinity predictions for the ten shortlisted generated compounds against \acs*{pdb} 1L2S.}
      Compounds are ranked by predicted affinity.
      $v$ is the returned \texttt{affinity\_pred\_value}; $\mathrm{pIC}_{50}=6-v$ is the scale used throughout this work, and \texttt{predicted\_pic50} is the backend-native field, equal to $1.364\,(6-v)$.
      Binding prob. is the predicted binding probability, and Confidence is the returned model confidence score.
      The table also reports the \acs*{iptm}.
      The generator column identifies each compound's origin in the recorded workflow and is not part of the Boltz-2 output.
      These are model outputs for compounds that were never synthesized or assayed; they are not measured potencies and do not establish AmpC activity.}
      \label{tab:ampc-boltz2}
      \begin{tabular}{rllrrrrrr}
            \toprule
            \textbf{Rank} & \textbf{Compound} & \textbf{Generator} & \textbf{$v$} & \textbf{$\mathrm{pIC}_{50}$} & \texttt{predicted\_pic50} & \textbf{Binding prob.} & \textbf{Confidence} & \textbf{\acs*{iptm}} \\
            \midrule
            1  & 1718 & PrexSyn    & 0.543 & 5.457 & 7.443 & 0.213 & 0.915 & 0.761 \\
            2  & 1721 & PrexSyn    & 0.777 & 5.223 & 7.124 & 0.410 & 0.966 & 0.898 \\
            3  & 1663 & PocketXMol & 0.781 & 5.219 & 7.118 & 0.363 & 0.958 & 0.921 \\
            4  & 1572 & MolMIM     & 0.898 & 5.102 & 6.959 & 0.160 & 0.885 & 0.648 \\
            5  & 1527 & MolMIM     & 0.973 & 5.027 & 6.857 & 0.248 & 0.948 & 0.831 \\
            6  & 1522 & MolMIM     & 1.438 & 4.563 & 6.223 & 0.277 & 0.972 & 0.937 \\
            7  & 1543 & MolMIM     & 1.438 & 4.563 & 6.223 & 0.217 & 0.908 & 0.700 \\
            8  & 1568 & MolMIM     & 1.641 & 4.359 & 5.946 & 0.328 & 0.936 & 0.796 \\
            9  & 1664 & PocketXMol & 1.969 & 4.031 & 5.499 & 0.201 & 0.927 & 0.803 \\
            10 & 1582 & MolMIM     & 2.281 & 3.719 & 5.072 & 0.215 & 0.964 & 0.897 \\
            \bottomrule
      \end{tabular}
\end{table}

The regression output is returned in a quantized representation; compounds 1522 and 1543 receive the identical value 1.4375, so ranks 6 and 7 are tied.
The reported \ac{iptm} equals the reported ligand \ac{iptm} for every complex.
Three of the eighty numeric cells in the agent's own summary table are one unit high in the final decimal, because a four-decimal intermediate was rounded a second time; the verification script identifies exactly those three cells and reproduces every other value from the raw service responses.

An \href{https://github.com/marcelmbn/farmaco-artifacts/blob/main/case-studies/ampc/environment-manifest.json}{environment manifest} records the recoverable request settings, artifact hashes, and historical build settings.
The recorded workflow, separate Boltz-2 completion, and no-database ablation used \texttt{gpt-5.5}.
The execution code revisions, exact remote-service versions, and random seeds were not preserved.
The source \ac{pdb} coordinate data are available under CC0~1.0\footnote{\url{https://www.rcsb.org/pages/policies}}.

\subsubsection{No-database ablation}
\label{sec:si-ampc-nodb}

The ten prompts of the recorded workflow were replayed on September 18, 2026 against a configuration in which no \ac{sql} tools are exposed.
In that configuration the agent receives the same scientific tools in a raw form that returns \ac{json} directly, keeps every intermediate result in its context, and writes output files to the chat's working directory.
The prompt texts are identical to the recorded series except for one line: the instruction to load the dataset into the database was replaced by an instruction to load it into the working environment, because the database does not exist in this configuration.
The dataset \ac{csv} was supplied as a file attachment on the first prompt rather than referenced by path.
The run used one continuous conversation, as the recorded workflow did.

\begin{table}
      \centering
      \small
      \caption{\textbf{Resource comparison for the ten original AmpC turns and successful Boltz-2 completions.}
      Model calls are the number of language-model invocations recorded in the trace for that turn, including those made by sub-agents.
      Costs are the recorded totals in USD; latency is the server-side turn time in seconds, defined in \cref{sec:si-ampc}.
      The Boltz-2 completion is indexed as turn 12 with the database and turn 11 without it; each total covers eleven completed executions and excludes failed and empty recovery attempts.
      A dash denotes an unavailable model-call count for the database-backed completion and its expanded total.
      The task labels include \acs*{admet} triage.
      Each configuration was executed once.}
      \label{tab:ampc-nodb}
      \begin{tabular}{llrrrrrr}
            \toprule
            & & \multicolumn{3}{c}{\textbf{With database}} & \multicolumn{3}{c}{\textbf{No database}} \\
            \cmidrule(lr){3-5}\cmidrule(lr){6-8}
            \textbf{Turn} & \textbf{Task} & \textbf{Calls} & \textbf{USD} & \textbf{s} & \textbf{Calls} & \textbf{USD} & \textbf{s} \\
            \midrule
            1  & Dataset ingestion          & 11  & 0.25 & 84    & 5  & 0.14 & 48   \\
            2  & Docking rank vs.\ activity & 14  & 0.37 & 112   & 5  & 0.27 & 77   \\
            3  & Selection strategy         & 5   & 0.19 & 42    & 2  & 0.13 & 38   \\
            4  & Retrospective selection    & 34  & 0.78 & 1153  & 35 & 2.03 & 1128 \\
            5  & Pocket detection           & 15  & 1.07 & 501   & 5  & 1.22 & 218  \\
            6  & Interaction profiles       & 82  & 5.55 & 3897  & 22 & 4.76 & 907  \\
            7  & \ac{admet} triage          & 30  & 2.42 & 1100  & 16 & 2.62 & 627  \\
            8  & Generative follow-up       & 48  & 2.55 & 1049  & 7  & 1.46 & 422  \\
            9  & Which algorithm            & 1   & 0.21 & 190   & 1  & 0.25 & 119  \\
            10 & PocketXMol and PrexSyn     & 63  & 3.76 & 2258  & 5  & 1.09 & 368  \\
            \midrule
            \multicolumn{2}{l}{Ten-turn subtotal} & 303 & 17.16 & 10387 & 103 & 13.97 & 3952 \\
            -- & Boltz-2 completion         & --  & 0.50 & 309   & 3   & 0.77 & 1191 \\
            \midrule
            \multicolumn{2}{l}{\textbf{Total}} & \textbf{--} & \textbf{17.66} & \textbf{10696} & \textbf{106} & \textbf{14.74} & \textbf{5143} \\
            \bottomrule
      \end{tabular}
\end{table}

Including the successful Boltz-2 completions, the database and no-database configurations used 16{,}479{,}165 and 15{,}464{,}913 total input-plus-output tokens, respectively.
The archived exports do not provide the input, cache-read, and output breakdown needed to report those categories separately.
The original ten-turn subtotals are 16{,}222{,}621 and 14{,}519{,}173 tokens, with 303 and 103 model calls; the no-database Boltz-2 completion adds three model calls, whereas the database-backed completion's model-call count is not recorded in the exported index.
Of the 230 tool calls in the original ten-turn database workflow, 186 address the database (106 read-only queries, 38 workflow invocations, 23 workflow-tool descriptions, and 19 subset selections by description), and each is a separate model round trip.
The last two categories have no counterpart in the no-database configuration.

Each configuration was executed once, so the figures carry no estimate of run-to-run variation.
The agent also selected different DiffDock sampling parameters in the two runs without being instructed to: the recorded workflow requested five poses at the service defaults, whereas the no-database run requested one pose with ten time divisions and six diffusion steps.
The reduced budget lowers both the runtime and the volume of returned \ac{json} on the three turns that dock or analyse poses, which flatters the no-database configuration on the interaction-profile and analogue-search turns.
It does not affect the direction of the retrospective-selection result, where the no-database configuration was more expensive despite the cheaper sampling.

The no-database run also completed the Boltz-2 affinity turn that failed in the recorded workflow, using ten raw affinity calls in a single turn.
The original database-backed attempt failed because of a remote service timeout; its successful completion used a fresh conversation, workspace, and database with the ten ligand \ac{smiles} supplied explicitly, whereas the no-database completion continued its existing conversation.
The expanded totals therefore sum eleven recorded completed executions per configuration with different conversation histories for the final step; they exclude the failed attempt, the empty recovery turn, and elapsed time between conversations.
They do not estimate unrecorded recovery costs.

\subsubsection{Instructions for the reported workflow}

The eight scripted prompts each repeat the same dataset preamble followed by one task block.
The preamble is reproduced once below, followed by the eight task blocks and the two follow-up messages included in the reported workflow, in the order they were sent.

\begin{lstlisting}[style=tbxtpromptstyle]
# Discovery pipeline

This case study is based on a published AmpC beta-lactamase virtual-screening
campaign. The study docked an ultra-large make-on-demand compound library
against AmpC and experimentally tested a prioritized subset of compounds. The
main scientific question was whether docking rank, library scale,
compound-selection strategy, and experimental testing scale affect hit
discovery. The original docking receptor was based on PDB 1L2S. Final
experimental cocrystal structures are available for selected hits: 9C81, 9C6P,
9C84, and 9DHL. A CSV file from the study is available and contains the
experimentally tested compounds and associated screening metadata.
Dataset file: `../assets/ampc_beta-lactamase.csv`
\end{lstlisting}

\begin{lstlisting}[style=tbxtpromptstyle]
## Prompt 1: Dataset ingestion and orientation

A CSV file from a published AmpC beta-lactamase virtual-screening campaign is
available. Load the dataset into the Farmaco database. Inspect the imported
records and summarize what scientific information the dataset contains.
Identify the key fields available for retrospective benchmarking. Report the
number of compounds, available identifiers, docking-related fields,
experimental activity fields, and selection-strategy fields.

## Prompt 2: Docking rank versus activity

Analyze whether the original docking scores and ranks predict experimental
activity. Report hit rates across score and rank regions. Summarize whether
better-ranked compounds are more likely to be active.

## Prompt 3: Selection strategy analysis

Compare the experimental outcomes of the different compound selection
strategies. Analyze whether manually selected, automatically selected, or
jointly selected compounds differ in hit rate or potency. Report the key
findings concisely.

## Prompt 5: Blinded follow-up selection

Using only information that would have been available before experimental
testing, select 50 compounds from the AmpC virtual-screening CSV for
hypothetical follow-up. Do not use experimental activity or Ki values during
selection. Explain the selection strategy briefly. After selection, evaluate
the selected compounds against the experimental outcomes. Generate docked
structures for the 50 newly selected compounds.

## Prompt 6: Pocket detection and analysis

Using the AmpC target structures, identify the likely ligand-binding pocket.
Run pocket detection and compare the predicted pocket residues with the known
AmpC active-site region. Report whether the expected beta-lactamase active site
is recovered.

## Prompt 7: Interaction-profile analysis

Using the AmpC benchmark data and the final experimental cocrystal structures
9C81, 9C6P, 9C84, and 9DHL, analyze the ligands with available structures. Use
1L2S as the original docking receptor and the final cocrystal structures only
as experimental validation references. Compare docked poses against
crystallographic poses where possible. Run pose-quality checks and
protein-ligand interaction analysis. Assess whether the predicted poses recover
the key AmpC active-site interaction patterns. Compare interaction fingerprints
between docked and crystal poses. Identify cases where docking recovered the
binding mode well and cases where activity was recovered but the pose was
incorrect. Report ligand RMSD, pose-quality flags, recurring active-site
contacts, and key failure modes.

## Prompt 8: ADMET-based hit triage

Using the AmpC benchmark data, identify experimentally active compounds and
rank them by developability. Use experimental activity only to define the hit
set. Within the hit set, rank compounds using predicted ADMET properties,
structural alerts, ligand properties, chemical diversity, and supplier
availability. Prioritize compounds that combine activity with favorable
predicted solubility, permeability, metabolic stability, low toxicity risk,
clean structural-alert profiles, and practical availability. Report the top
follow-up candidates and explain the main ADMET or structural liabilities that
deprioritized other hits.

## Prompt 9: Generative follow-up design

Starting from the strongest AmpC hits in the CSV, propose follow-up analogs.
Use molecular generation workflows and filter the generated compounds by ADMET,
structural alerts, and similarity to known hits. Report a small prioritized set
of suggested analogs and explain the design rationale.
\end{lstlisting}

The prompt series is numbered 1 through 9 with no prompt 4.
The two follow-up messages were sent after the scripted series, in this order:

\begin{lstlisting}[style=tbxtpromptstyle]
What algorithm did you use to generate more candidates?
\end{lstlisting}

\begin{lstlisting}[style=tbxtpromptstyle]
Can you try other generative algorithms such as PocketXMol and PrexSyn to
generate even more compounds and evaluate them in the same way?
\end{lstlisting}

\clearpage
\subsection{KpThiL binding-site analysis, generation, and affinity/ADMET follow-up}
\label{sec:si-kpthil}

\subsubsection{Binding-site analysis}
\farm{} defined the substrate-binding pocket using the two bound \ac{tps} molecules, one in each protomer.
The two sites had contact-residue Jaccard indices of $0.73$ at 4~\AA{} and $0.83$ at 6~\AA{}.
Chain~A was selected as the representative site while preserving the full dimer context, since partner-chain residues THR123, THR124, and ASP122 lie within 4--6~\AA{} of the ligand.
Substrate recognition involves a polar cluster observed in both chains.
Key contacts include GLU263, which forms the closest interaction with the aminopyrimidine N4A atom at approximately 2.6~\AA{}; the ASN53 backbone carbonyl, which contacts the ring nitrogens; HIS54, which lines the phosphate region through NE2 at approximately 2.9~\AA{}; and ASP215, which also contributes to phosphate recognition.
The surrounding contact shell comprises PHE55, LEU56, LEU168, HIS190, and ASP47.
P2Rank identifies the site as pocket~2, with a closest-pocket distance of approximately 2.6~\AA{}, while Pocketeer recovers symmetric pockets 4 and 7 together with pocket~5 spanning the \ac{tps} contact region.
Because \ac{atp} is absent from the structure, putative \ac{atp}-binding subsites, including P2Rank pocket~1 with a reported probability of 0.95, are labelled as inferred and treated as secondary.

\subsubsection{Execution record and follow-up prompts}
The initial pocket analysis and 50-compound catalogue selection comprised 16 user turns.
A separate seven-turn continuation began with the initial handoff report supplied as context.
A subsequent one-turn affinity/ADMET follow-up used a fresh database and explicit input files.
All three complete conversation exports and working directories, including three independent database snapshots, are released in \texttt{case-studies/kpthil/}.
The key execution prompts in the continuation were, in order:

\begin{lstlisting}[style=tbxtpromptstyle]
Can you run PocketXMol as suggested?
\end{lstlisting}

\begin{lstlisting}[style=tbxtpromptstyle]
Run the recommended next step.
\end{lstlisting}

\begin{lstlisting}[style=tbxtpromptstyle]
Can you run PrexSyn to find synthesizable analogs?
\end{lstlisting}

\subsubsection{Generation and molecular checks}
The agent imported and cleaned the local experimental structure, then ran PocketXMol using the chain~A \ac{tps} centroid $(-22.689,-6.431,10.576)$~\AA{} and a 10~\AA{} radius.
All ten generated molecules passed \ac{pains} screening and had valid RDKit descriptors.
An initial descriptor request failed because it used the unsupported metric name \texttt{LogP}; a subsequent request using \texttt{MolLogP} succeeded.
Each generated conformer passed all 12 PoseBusters checks under the \texttt{mol} configuration.
These checks assess molecular integrity and internal geometry; they do not validate receptor-relative binding poses or recovery of the observed \ac{tps} interactions.

\subsubsection{Docking and synthesis-route exploration}
DiffDock was attempted for three prioritized PocketXMol molecules, but all three predictions failed because the then-current SQL workflow required a canonical \ac{pdb} identifier in \texttt{target.pdb}, which was absent for the unpublished local structure.
PrexSyn was queried for five analogues per generated molecule and succeeded for nine of the ten queries, returning 45 analogue hits with proposed synthesis routes; the remaining query failed after a remote connection abort.
These analogues were not experimentally tested or incorporated into the original 50-compound catalogue selection, and no successful docking ranking was obtained.

\subsubsection{Standalone affinity and ADMET follow-up}
\label{sec:si-kpthil-ranking}
The standalone task \texttt{dimeric\_ThiL\_enzyme\_followup} was executed once on October 19, 2026, in the database-enabled configuration using the complete prompt archived under \texttt{input/prompts/ranking\_followup/01.md}.
Its opening instruction was:

\begin{lstlisting}[style=tbxtpromptstyle]
Using the supplied protein and candidate molecules, produce a new ranking for prospective KpThiL
ligand testing based on Boltz-2 predictions and ADMET properties.
\end{lstlisting}

The attached files were \texttt{candidates.csv}, the unchanged experimental \texttt{KpThiL\_refine\_011.pdb}, and \texttt{target.fasta}.
The CSV contained the earlier 50 catalogue compounds, ten PocketXMol molecules, and 45 PrexSyn hits, with source labels and historical vendor identifiers but without prior ranks or scores.
No prior database or conversational history was imported.
All 105 records parsed successfully and remained distinct after RDKit stereochemistry-aware canonicalization.
The target sequence represented the observed 323-residue construct, reconciling the experimental chains by residue number.
The run completed in 49.2~min; its saved telemetry records 52 \ac{llm} calls with \texttt{gpt-6-sol} and USD~0.643 in \ac{llm} costs, excluding scientific-service costs.
The full export is archived in \texttt{log/ranking\_followup/}, and the full workspace is archived in \texttt{data/ranking\_followup/} within \texttt{case-studies/kpthil/}.

\farm{} calculated RDKit descriptors, screened the full structural-alert catalogue, and ran \acs*{admet}-\acs*{ai} for all 105 candidates \cite{swansonADMETAIMachineLearning2024}.
All descriptor and \ac{admet} requests succeeded; 71 candidates triggered at least one alert across the broader catalogue, including one PrexSyn hit, \texttt{prexsyn\_002\_02}, with a \ac{pains}-B flag.
The broader alert counts can include overlapping matches from different catalogues and do not imply that all 71 compounds fail \ac{pains}.
The agent retained full \ac{admet} outputs and considered solubility, Caco-2 and PAMPA permeability, CYP inhibition, clearance, and toxicity predictions, including hERG, Ames, and \ac{dili}.

Boltz-2 complex and affinity predictions succeeded for all 105 ligands \cite{passaroBoltz2AccurateEfficient2025}.
Each request contained one protein sequence and one ligand, with three recycling steps, 50 structure-sampling steps, one structure sample, 50 affinity-sampling steps, one affinity sample, and molecular-weight affinity correction disabled.
The experimental dimer was not a co-folding input.
The canonical converted affinity score was $\mathrm{pIC}_{50}=6-\texttt{affinity\_pred\_value}$ and spanned 3.56--6.66.
The raw \ac{nim} \texttt{affinity\_pic50} field, which uses a separate scale, is retained in the response JSONs and audit table but was not used for this conversion.
The ranked CSV also expresses these predictions as $10^{6-\mathrm{pIC}_{50}}$~\textmu{}M; these are model-derived potency proxies, not measured enzyme inhibition or binding affinities.
These micromolar proxies were calculated from the unrounded stored scores; the $\mathrm{pIC}_{50}$ values in \cref{tab:kpthil-ranking} are rounded to two decimal places.

\subsubsection{Pose checks and heuristic ranking}
\label{sec:si-kpthil-poses}
For each predicted complex, the agent used Kabsch alignment on 315 shared chain-A C$\alpha$ positions and applied the same transformation to the ligand.
Eight positions in the 323-residue sequence (1--6, 17, and 18) lacked experimental chain-A C$\alpha$ coordinates and were excluded from the alignment.
The median receptor C$\alpha$ root-mean-square deviation was 1.22~\AA{}.
It classified a pose as a \ac{tps}-pocket fit when its ligand centroid was less than 9~\AA{} from the experimental \ac{tps} centroid, at least 30\% of its heavy atoms were within 5~\AA{} of any \ac{tps} heavy atom, and it contacted at least two chain-A residues within 4.2~\AA{}.
It counted the ligand heavy atoms whose nearest experimental protein heavy atom in either chain was less than 2~\AA{} away as potential clash atoms.
This count is neither the number of overlapping atom pairs nor a force-field clash energy.
Nine poses passed the pocket heuristic: six had at most three clash atoms and three had more than three; only \texttt{catalogue\_003} had zero.
The remaining 96 poses were classified outside the \ac{tps} pocket by this rule.
The strongest converted Boltz-2 affinity score, $\mathrm{pIC}_{50}=6.66$ for \texttt{prexsyn\_003\_02}, corresponded to a ligand centroid 26.87~\AA{} from the \ac{tps} centroid.

The agent chose and revised the thresholds and score during execution; they were not prespecified screening criteria.
The final ranking sorted first by pocket/clash tier and then by a weighted score combining converted affinity, binder probability, ligand interface confidence, pocket proximity, clash counts, predicted solubility and permeability, toxicity/CYP scores, broader structural alerts, and catalogue status.
The three tiers, in order, were \ac{tps}-pocket fits with at most three potential clash atoms, pocket fits with more than three, and other poses; the weighted score ordered candidates within each tier.
A final adjustment applied additional penalties for hERG, Ames, or \ac{dili} scores of at least 0.8.
The full score and final adjustment are documented in the artifact README and preserved execution history.
Morgan-fingerprint similarities were reported relative to the initial score order; they did not contribute to the score and were not recomputed after the final reranking.
The final rank order therefore represents the agent's exploratory prioritization rather than a validated or prospectively fixed selection model.

\begin{table}
    \centering
    \caption{\textbf{First five candidates from the final KpThiL ranking, all of which passed the \acs*{tps}-pocket heuristic.}
    Converted Boltz-2 $\mathrm{pIC}_{50}$, binder probability, and hERG and \acs*{dili} scores are computational predictions, not experimental measurements or calibrated probabilities for these candidates.
    Clash atoms are ligand heavy atoms whose nearest experimental protein heavy atom was less than 2~\AA{} away.
    The ranking combines pocket/clash tiers with a weighted score, including additional penalties for hERG, Ames, or \acs*{dili} scores of at least 0.8.
    Rank~1 was designated a \acs*{tps}-like reference by the agent; ranks~3--5 require synthesis or availability checks.}
    \label{tab:kpthil-ranking}
    \small
    \begin{tabular}{rlrrrrr}
        \toprule
        Rank & Candidate & $\mathrm{pIC}_{50}$ & Binder probability & Clash atoms & hERG & \acs*{dili} \\
        \midrule
        1 & \texttt{catalogue\_003} & 4.66 & 0.226 & 0 & 0.047 & 0.013 \\
        2 & \texttt{catalogue\_001} & 3.56 & 0.152 & 1 & 0.074 & 0.017 \\
        3 & \texttt{pocketxmol\_002} & 5.50 & 0.152 & 1 & 0.860 & 0.976 \\
        4 & \texttt{pocketxmol\_003} & 4.80 & 0.098 & 2 & 0.610 & 0.809 \\
        5 & \texttt{prexsyn\_004\_05} & 4.42 & 0.207 & 2 & 0.863 & 0.403 \\
        \bottomrule
    \end{tabular}
\end{table}

The agent recommended \texttt{catalogue\_003} (historical Enamine identifier \texttt{CSSS00020675773}) as a \ac{tps}-like reference and \texttt{catalogue\_001} (\texttt{CSSS00015206161}) as a related comparator, with model-derived potency proxies near 22 and 274~\textmu{}M, respectively.
The comparator \texttt{catalogue\_001} had the weakest converted affinity prediction in the set; its second-place rank reflects the tiered multi-criterion heuristic, not a claim of stronger affinity than lower-ranked candidates.
The highest-ranked novel generated molecule, \texttt{pocketxmol\_002}, had a proxy near 3.2~\textmu{}M but binder probability 0.152 and predicted hERG and \ac{dili} scores of 0.860 and 0.976.
The optional PrexSyn candidate \texttt{prexsyn\_004\_05} had a proxy near 38~\textmu{}M and hERG score 0.863 (\cref{tab:kpthil-ranking}).
The report found no convincing novel purchasable candidate for testing around 10--20~\textmu{}M and proposed a small reference/comparator-focused biochemical set, with the generated candidates remaining exploratory.
Catalogue availability, synthesis feasibility, binding, and inhibition were not experimentally verified.

The new Boltz-2 poses did not undergo PoseBusters molecular-geometry validation; the earlier ten-molecule PoseBusters results concern different generated conformers.
Partner-chain contacts and clashes were assessed only after alignment to the experimental dimer and are not interactions supported by a dimeric Boltz-2 prediction.
The run performed no \ac{abfe} or \ac{rbfe} calculations.
The preserved agent script \texttt{rank\_build.py} produces the initial score order; a separate REPL adjustment produced the final tiered CSV.
The offline script \texttt{scripts/reproduce\_paper\_numbers.py} replays both steps in a temporary directory, requires a byte-identical final CSV, and independently recomputes pocket/contact/clash checks from the saved coordinates.
The archived raw outputs, including the agent's final report and ranking, are retained unchanged.

\clearpage
\subsection{TYK2 relative binding free energy}
\label{sec:si-tyk2}

\subsubsection{Single-edge and network benchmark sessions}
\label{sec:si-tyk2-benchmark}

\paragraph{Data and execution record.}

The original single-edge and network analysis used the two first-turn instructions and the lossless conversation exports of both sessions.
The exports cover every turn, tool call, and tool result, together with per-turn costs and call arguments and the \farmacoatlas{} SQLite databases as the sessions left them.
The databases hold the Azulene Studio job records and results, including replica-exchange transition statistics.
The analysis also used the uploaded protein and ligand files, the DiffDock poses and Boltz-2 outputs of the comparison turn, the benchmark reference data, and normalized analysis tables.
The numerical comparisons below are retained from that analysis.
The currently released \texttt{case-studies/tyk2/} package covers the later interactive campaign described in \cref{sec:si-tyk2-design}; it does not contain the two September~29 sessions.
Both sessions ran on September 29, 2026 with \texttt{gpt-5.5}.
The protein structure (288 residues, ACE/NME-capped, with crystallographic waters) and the 13 posed ligands were taken from the TYK2 system of an Azulene Studio validator set provided by Azulene Labs, which draws on the OpenFE/OpenFF protein--ligand benchmark \cite{hahnBestPracticesConstructing2022,baumannLargescaleCollaborativeAssessment2025}.
The uploaded files are byte-identical between the two sessions.
Main-text ligand labels abbreviate the benchmark identifiers: EJM31 corresponds to \texttt{lig\_ejm\_31}, with the same convention used for the other EJM ligands.
Experimental $K_\mathrm{i}$ values and their reported errors \cite{liangLeadIdentificationNovel2013,liangLeadOptimization4Aminopyridine2013}, and the 18-edge \ac{lomap} reference network, were not uploaded.
Experimental $\Delta\Delta G_{A\to B} = RT\ln(K_{\mathrm{i},B}/K_{\mathrm{i},A})$ is evaluated at 298.15~K, with uncertainties propagated from the reported $K_\mathrm{i}$ errors.

\paragraph{Sessions and settings.}

\begin{table}
      \centering
      \small
      \caption{\textbf{User turns and Azulene Studio settings of the two TYK2 sessions.}
      Costs are recorded \acs*{llm} costs in USD and exclude Azulene Studio and other service costs.
      Cached input tokens are included in the token totals.}
      \label{tab:si-tyk2}
      \begin{tabular}{lrr}
            \toprule
                                                & \textbf{Single edge}           & \textbf{Network}                 \\
            \midrule
            User turns                          & 4                              & 3                                \\
            Tool calls                          & 29                             & 25                               \\
            \ac{llm} calls                      & 38                             & 24                               \\
            Total tokens                        & 344,656                        & 396,404                          \\
            \ac{llm} cost (USD)                 & 0.73                           & 0.95                             \\
            Azulene Studio jobs submitted / completed     & 1 / 1                          & 6 / 5                            \\
            Equilibration / production per replica & 0.5~ns / 2~ns (prescribed) & 5~ps / 10~ps (tool default)      \\
            Protocol repeats                    & 1 (prescribed)                 & 3 (tool default)                 \\
            pH (protonation assigned)           & 7.4 (agent choice)             & 7.0 (tool default)               \\
            \bottomrule
      \end{tabular}
\end{table}

In the single-edge session, \farm{} was asked to compute \texttt{lig\_jmc\_23}$\to$\texttt{lig\_jmc\_27}, a fluorine-to-chlorine change on a \textit{cis}-cyclopropane, using the prescribed settings in \cref{tab:si-tyk2}.
The agent inspected the uploaded files, registered the target, ligands, and poses, submitted one job, and returned its identifier without waiting.
The follow-up prompts were ``Can you check the status?'', ``Did the calculation finish?'', and ``Can you compare to related metrics by DiffDock and Boltz-2?''.
The two status requests returned \texttt{running} and then $\Delta\Delta G = -0.28$~kcal\,mol$^{-1}$, weakly favouring \texttt{lig\_jmc\_27}; the single repeat provided no uncertainty estimate.
For the comparison, the agent stored the sequence extracted from the uploaded structure on the target record, generated one RDKit conformer per ligand, and ran DiffDock against the uploaded protein and Boltz-2 from sequence and \ac{smiles} as subset workflows.
Neither method used the supplied binding poses.
DiffDock reported best-pose confidences of 0.305 and 0.492 for \texttt{lig\_jmc\_23} and \texttt{lig\_jmc\_27}.
Boltz-2 predicted \(\mathrm{pIC}_{50}\) values of 6.797 and 6.613, calculated as $6-v$ from the raw \texttt{affinity\_pred\_value} $v$, and binding probabilities of 0.750 and 0.664.
The \(\mathrm{pIC}_{50}\) difference corresponds to an energy-equivalent estimate of $+0.25$~kcal\,mol$^{-1}$ at 298.15~K in favour of \texttt{lig\_jmc\_23}.
The agent noted that DiffDock confidence is not an affinity estimate and judged the two ligands approximately tied.
The experimental difference, $+0.42 \pm 0.25$~kcal\,mol$^{-1}$ in favour of \texttt{lig\_jmc\_23} ($K_\mathrm{i}$ of 2.5 versus 5.1~nM), is consistent with that judgement.
The \ac{rbfe} estimate has the opposite sign but an absolute error of 0.70~kcal\,mol$^{-1}$, whereas the Boltz-2 difference matches the experimental direction.

In the network session, the follow-up prompts were ``Can you submit the first 6 most interesting calculations?'' and ``Check if the calculations are ready.''.
The agent's initial network comprised 12 edges and its optional set seven.
Six of the initial and three of the optional edges also occur in the reference \ac{lomap} network.
The agent connected the acyclic and cyclic amide series through \texttt{lig\_ejm\_43}--\texttt{lig\_ejm\_46}--\texttt{lig\_jmc\_23}, whereas the reference network uses hub edges from \texttt{lig\_ejm\_31} and \texttt{lig\_ejm\_42}.
The submitted star contained no cycles, so none of the proposed cycle-closure checks could be evaluated.
The \texttt{lig\_ejm\_42}$\to$\texttt{lig\_ejm\_55} job failed with ``Propagating replica 7 at state 7 resulted in a NaN!''.
For the five completed jobs, the database records the free-energy results and replica-exchange statistics but not Azulene Studio's per-job simulation archives, which were not downloaded; these archives exceeded Azulene Studio's 50~MB storage limit and were published without their checkpoint files.
Across the five completed edges, the root-mean-square error with respect to experiment is 1.39~kcal\,mol$^{-1}$, and four of the five estimates lie within twice their reported uncertainty of the experimental value.
The \ac{abfe} workflow was not exercised in these sessions.

\subsubsection{Interactive analogue-design campaign}
\label{sec:si-tyk2-design}

\paragraph{Inputs and execution record.}

The released \texttt{case-studies/tyk2/} package preserves the complete 28-turn conversation of chat \texttt{92ecead9-a815-4d1a-b857-e4d3b7d5817a}, recorded on October 2--6, 2026 with \texttt{gpt-6-sol}.
It includes all user instructions, delivered responses, tool-call arguments and results, per-turn costs and trace identifiers, and the complete 152-file working directory.
The unchanged \farmacoatlas{} database contains 63 ligand records, 113 structures, 67 workflow runs, 294 results, 1,370 metrics, and three Azulene Studio jobs; ligand-record counts include duplicate generative outputs and do not count unique proposed compounds.
Normalized tables and a standard-library script reproduce the candidate selections, Boltz-2 comparisons, directed \ac{rbfe} results, preparation settings, and conversation totals, and verify the retained evidence hashes and database file references.

The three initial files were \texttt{protein.pdb}, \texttt{seed.csv}, and \texttt{seed\_001.sdf}.
They are byte-identical to the initial assets of the TYK2 adaptive lead-optimization prompt package derived from the Azulene Labs validator bundle.
The seed is a renamed \texttt{lig\_ejm\_31}; identifying receptor headers and historical ligand properties had been removed before upload.
The supplied receptor has a 288-residue protein chain, terminal caps, and two waters; the ligand pose was supplied separately.
No reference ligands or experimental activity table were revealed in this recorded session, and no generated candidate was experimentally tested.
The first instruction was:

\begin{lstlisting}[style=tbxtpromptstyle]
I've uploaded a TYK2 protein structure and a starting ligand with its proposed binding pose.
Can you develop a small series of improved TYK2 inhibitors from this starting point?

Explore your own design ideas, balancing predicted potency with drug-like properties and synthetic feasibility.
Keep published TYK2 compounds and activity data out of this design phase.
Select up to five promising candidates and explain the chemical hypotheses behind them.
Keep this first assessment exploratory and inexpensive, leaving binding free-energy calculations for a later stage.

Return your proposed compounds, reasoning, and the evidence supporting your initial shortlist.
\end{lstlisting}

\paragraph{Design, screening, and receptor preparation.}

The agent's initial eight analogue proposals were followed by 24 MolMIM output records from two seeds, 17 GenMol outputs, and 11 successfully persisted PocketXMol structures from 12 requested samples.
The agent excluded GenMol's small fragments from the TYK2 investigation set and selected two non-identical PrexSyn analogues alongside manual designs and MolMIM and PocketXMol outputs.
The consolidated 11-member subset contains ligand IDs 2, 3, 4, 5, 13, 15, 24, 26, 56, 62, and 63; structural triage reduced the follow-up set to 2, 5, 15, 24, and 63.
Boltz-2 initially failed for the five-subject seed/analogue screen and a subsequent single-subject retry with a hosted CUDA error; later runs completed for the seed and all 11 consolidated candidates.
Ligand~15, from MolMIM run~16, had a reported Boltz-2 affinity score of 9.255 and binding probability 0.828, compared with 8.866 and 0.805 for the seed.
The archived \texttt{predicted\_pic50} field equals $1.364\,(6-\texttt{binding\_affinity})$ for all 12 successful predictions and is not the normalized \(\mathrm{pIC}_{50}\) scale $6-\texttt{binding\_affinity}$ used elsewhere in the manuscript.
The latter gives 6.785 for ligand~15 and 6.500 for the seed; the analysis tables retain the raw regression value, archived score, and normalized scale separately.
\Cref{tab:tyk2-design} reports the normalized $\mathrm{pIC}_{50}$ values.
The agent compared the archived scores; neither scale constitutes a measured inhibition constant.

The submitted candidate poses were structures~77, 76, and~80 for ligands~2, 5, and~15, respectively, generated by \texttt{ligand\_template\_pose} using seed structure~3.
All three passed 22/22 standard PoseBusters checks against prepared receptor~101.
The matched core was constrained to the seed frame; zero mapped-core displacement is a consequence of construction and does not independently validate binding.
Rowan preparation run~53 failed after its default Boltz-2 missing-residue method \cite{passaroBoltz2AccurateEfficient2025} introduced an unrecognized terminal \texttt{UNK} residue; its saved report records 13.86 credits charged.
Run~54 succeeded on cleaned structure~61 using PDBFixer, ACE/NME caps, and OpenMM protonation at pH~7.4 \cite{eastmanOpenMM8Molecular2023}, producing water-free structure~101 and charging 4.22 credits.
The original waters were HOH A1318 and A1325; the agent identified a modeled bridge involving A1325, the seed carbonyl, and Ser985 and explicitly described water removal as a protocol assumption.

\paragraph{Submissions, results, and limitations.}

The user reviewed the receptor, seed and candidate files, simulation lengths, and protonation settings before authorizing the seed$\to$2 and seed$\to$5 submissions on October~3.
Both jobs used receptor~101, seed pose~3, pH~7.4, \texttt{assign\_protonation\_states=false}, 0.5~ns equilibration, 5~ns production per replica, three repeats, and \texttt{keep\_dirs=true}.
Their results were collected on October~5.
The seed$\to$15 submission initially failed with \ac{http}~402, ``Credits exhausted''; a subsequent attempt reused the rejected local record without contacting Azulene Studio.
After a change to the submission safeguard, the same local job acquired a remote identifier and completed; the result was collected on October~6.
The database retains all three completed jobs and full raw scalar results, including replica-exchange transition statistics; the conversation retains the rejected submission and retry history.

\begin{table}
      \centering
      \small
      \caption{\textbf{Recorded resources and protocol of the TYK2 design session.}
      Costs are recorded \acs*{llm} costs in USD and exclude Azulene Studio, Rowan, and other scientific services.
      Total tokens include cached input tokens.
      Simulation lengths are the effective submitted settings, rather than tool defaults; false indicates that protonation was not reassigned.}
      \label{tab:si-tyk2-design}
      \begin{tabular}{lr}
            \toprule
            \textbf{Quantity} & \textbf{Recorded value} \\
            \midrule
            User turns & 28 \\
            Tool calls & 264 \\
            \ac{llm} calls & 467 \\
            Total tokens & 17,454,238 \\
            Cached input tokens & 13,686,292 \\
            \ac{llm} cost (USD) & 9.82 \\
            Consolidated / pose-supported candidates & 11 / 5 \\
            Completed Azulene Studio jobs & 3 \\
            Equilibration / production per replica & 0.5~ns / 5~ns \\
            Protocol repeats & 3 \\
            pH / protonation reassignment & 7.4 / false \\
            \bottomrule
      \end{tabular}
\end{table}

For each job, Azulene Studio reported that its full result bundle exceeded the 50,331,648-byte storage limit and omitted two complex-phase \texttt{simulation.nc} files from the downloadable archive.
The agent did not download these archives; the release retains scalar outputs, transition statistics, and omission metadata, but does not contain simulation trajectories.
The reported uncertainties do not quantify pose, force-field, or water-model error, and convergence was not established.
The third result, $+0.018\pm0.099$~kcal\,mol$^{-1}$, supports similar predicted seed/ligand~15 affinity only within this computational setup, pending experimental measurement.
The final recommendation was a matched-pair comparison for experimental testing; improved experimental potency was not established.
The \ac{abfe} workflow was not exercised in this session.

\clearpage
\section{Multi-task benchmark}
\label{sec:si-benchmark}

\subsection{Scope and experimental design}

We evaluated three agent configurations across 22 computational drug-discovery tasks: \farmaco{} with the \farm{} database and workflow interface (mode A; labelled F\'armaco in the figures), the same domain agent with the underlying scientific capabilities exposed through plain tool calls (mode B), and the standalone Codex coding agent (mode C).
All 198 runs used \texttt{gpt-6-sol} as their execution model; the Codex runs used \texttt{codex-cli 0.162.1}.
Each task used the same scientific instructions and supplied input files across configurations, and the agents chose their methods unless the task specified one, as in the 48-ligand Boltz-2 task.
Codex worked in the provisioned \farm{} software environment and could write and execute code and access the supplied scientific-service endpoints.
The two domain-agent configurations could also use model-backed web search.

Each task and configuration has three runs, with three separate judge sessions per run.
The benchmark contains 66 runs per configuration across all 22 tasks, giving 198 agent runs and 594 judge sessions in total.

\subsection{Task coverage}

\Cref{tab:si-benchmark-tasks} lists the scientific questions and principal input or output scope.
The task suite spans data analysis, molecular filtering and comparison, conformer generation, sequence and structure analysis, ligand generation, binding-pose assessment, and synthesis-route exploration.
The staged AmpC prioritization task carries candidates through property and alert filtering, a diverse cohort of at most 50 compounds for \ac{admet} assessment, at most ten compounds for affinity and pose prediction, and at most five nominations, followed by a tighter lipophilicity limit using the completed assessments.
Its separate reference-binding-mode follow-up starts a fresh agent context from a copy of the corresponding prioritization run's workspace, including its selected compounds and assessed structures.
Resource use for the follow-up covers its own invocation.

\clearpage
{\small
\renewcommand{\arraystretch}{1.15}
\begin{longtable}{@{}>{\raggedright\arraybackslash}p{0.12\linewidth}p{0.33\linewidth}p{0.48\linewidth}@{}}
    \caption{\textbf{Coverage of the 22-task benchmark.}
    All configurations receive the same instructions and original inputs for a task; the fresh-context AmpC follow-up instead inherits its own configuration's preceding workspace.
    Counts indicate the requested task scope.}
    \label{tab:si-benchmark-tasks}\\
    \toprule
    \textbf{System} & \textbf{Task} & \textbf{Principal scope} \\
    \midrule
    \endfirsthead
    \multicolumn{3}{l}{\textit{\Cref{tab:si-benchmark-tasks}, continued}}\\
    \toprule
    \textbf{System} & \textbf{Task} & \textbf{Principal scope} \\
    \midrule
    \endhead
    \bottomrule
    \endfoot
    AmpC & Hit ADMET comparison & Predict and compare properties and toxicity risks for 50 compounds. \\
    AmpC & Docking and pose assessment & Assess binding predictions for 20 compounds, with at most three poses each. \\
    AmpC & Hit triage & Apply potency, property, and structural-alert criteria to the supplied screening table. \\
    AmpC & Binding-pocket detection & Identify pockets in PDB 1L2S and compare them with the active site. \\
    AmpC & Screening analysis & Relate docking scores, ranks, and selection strategies to measured hit rates. \\
    AmpC & Reference binding-mode comparison & Reassess prior nominations against the catalytic-site reference ligand. \\
    AmpC & Staged hit prioritization & Carry linked candidate cohorts through filtering, ADMET, binding assessment, and revised nominations. \\
    AmpC & Synthesis-supported analogues & Search from ten seeds for up to 50 analogues with proposed routes and select up to 20. \\
    Building blocks & Amidation partner selection & Classify primary-amine building blocks for amide coupling and chemical compatibility. \\
    KpThiL & Structure prediction & Make at most one structure prediction from the supplied protein sequence. \\
    KpThiL & Substrate recognition & Compare the two bound thiamine-phosphate sites in the supplied dimer. \\
    Ligands & Conformer ensembles & Generate up to five minimized conformers for each of 50 compounds. \\
    Ligands & 3D shape and pharmacophores & Rank 49 compounds against aspirin using shape and pharmacophore overlays. \\
    PGK2 & Pocket-conditioned generation & Generate 30 molecules from the supplied pocket and select up to ten. \\
    PGK2 & Candidate property screening & Assess 369 generated compounds against property and alert criteria. \\
    STK17B & Binding prediction & Rank 48 ligands with Boltz-2 and analyze interactions for the top five. \\
    STK17B & Kinase-domain conservation & Retrieve at most 100 homologues and assess a multiple sequence alignment. \\
    STK17B & Structure and sequence & Compare deposited sequences and resolved chains in PDB 6ZJF. \\
    TBXT & Co-folding poses & Predict protein--ligand complexes for ten compounds and assess their poses. \\
    TBXT & Scaffold elaboration & Generate 30 elaborations retaining the supplied scaffold and select up to ten. \\
    TYK2 & Reference-ligand comparison & Compare 12 candidates with one supplied reference using structures and properties. \\
    TYK2 & Analogue binding-mode hypotheses & Assess up to three reference-guided poses for each of 12 analogues. \\
\end{longtable}
}

\clearpage

\subsection{Evidence-based judging}

Each run was assessed in three fresh standalone judge sessions using \texttt{gpt-6.1-sol} with high reasoning effort.
The judge received the original task, input-file names and unmodified inputs, the final response, a copy of the run's saved workspace, and any task-specific rubric.
The same evaluation instructions and task-specific criteria were used across configurations.
The judge could inspect saved files, molecular structures, tables, and databases and recompute checks in a separate working directory.
Task-specific rubrics emphasized issues such as alignment depth, scaffold retention, conformer validation, cohort membership, and distinguishing pose confidence from affinity.
The criteria accepted scientifically valid alternative methods and credited evidence-supported diagnoses of failures.

The judge assigned separate 0--10 scores for task completion, scientific correctness, evidence consistency, and overall quality, together with a pass, partial, or fail verdict.
The judge assigned the overall score directly from its assessment of the complete delivered result.
A pass indicated work acceptable to a domain expert, partial indicated useful but incomplete or flawed work, and fail indicated an unsatisfactory result.
For each run we used the median of its three judge scores, calculated separately for the overall and component scores.
Mode-level scores are means of these per-run medians.
Verdict percentages use all 198 judge sessions per configuration.

\subsection{Resource accounting and uncertainty}

Execution cost covers the agent's model calls in US dollars, including calls made inside scientific tools.
We also report search-adjusted cost by subtracting the charges for model calls made by the domain agents' search tool.
Total tokens accumulate input and output tokens over model calls, including cached input, whereas peak context is the largest input-token count of any single model call, including cached input.
The cached input share for each run is the number of cache-read tokens divided by total input tokens, including cached input; we report the median of these run-level shares.
For modes A and B, retry delays attributed to provider-side issues in model API calls were removed from the reported runtimes by subtracting 300~s per inferred stalled attempt.
Codex was measured as a black-box service, and its reported runtime is used directly.

We summarize latency, cost, cumulative tokens, and peak context by their medians because their distributions are skewed.
For each configuration, we estimated uncertainty by randomly drawing 22 tasks with repeat selections allowed, keeping all three runs of every selected task, and recalculating the mean or median.
We repeated this process 5,000 times and used the central 95\% of the resulting values as the confidence interval.

\subsection{Released records and reproduction}

The \href{https://github.com/marcelmbn/farmaco-artifacts/tree/main/case-studies/benchmark}{\texttt{case-studies/benchmark/} package} provides the task prompts, task-specific rubrics, original attachments, final responses, execution records, cost summaries, and judge assessments.
Its \texttt{analysis/runs.csv} contains per-run scores and resource totals, and \texttt{analysis/summary.csv} contains their aggregate statistics.
Cached input shares are calculated from each run's \texttt{costs.json}, using \texttt{unified} token counts for modes A and B and \texttt{codex} token counts for mode C.
The supplied analysis scripts reproduce the summary statistics and figures from the per-run table.

\subsection{Aggregate results}

\begin{figure}
    \centering
    \includegraphics[width=\linewidth]{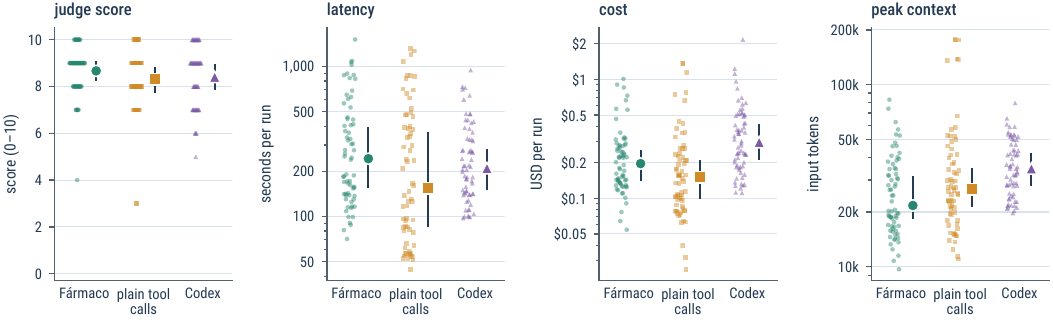}
    \caption{\textbf{Benchmark quality and resource use across agent configurations.}
    Small markers show individual runs; larger markers show the mean overall judge score or median latency, model cost, or peak context.
    Vertical bars show 95\% confidence intervals estimated by resampling tasks.
    Circles denote F\'armaco, squares denote plain tool calls, and triangles denote Codex.
    The latency, cost, and peak-context axes are logarithmic.
    Peak context is the largest recorded input-token count for one model call, including cached input.}
    \label{fig:si-benchmark-overview}
\end{figure}

\Cref{tab:si-benchmark-summary,fig:si-benchmark-overview} summarize the benchmark results.
The mean overall scores were 8.67 for F\'armaco, 8.30 for plain tool calls, and 8.41 for Codex, with overlapping task-bootstrap confidence intervals.
The corresponding pass shares were 151/198 (76.3\%), 123/198 (62.1\%), and 122/198 (61.6\%) judge sessions.
Partial verdicts numbered 47, 69, and 76, respectively; all six fail verdicts belonged to the plain-tool-call configuration.
Using Codex for the judge sessions may favor the methods or presentation of mode C outputs, introducing a potential evaluator bias into these comparisons.

\begin{figure}
    \centering
    \includegraphics[width=\linewidth]{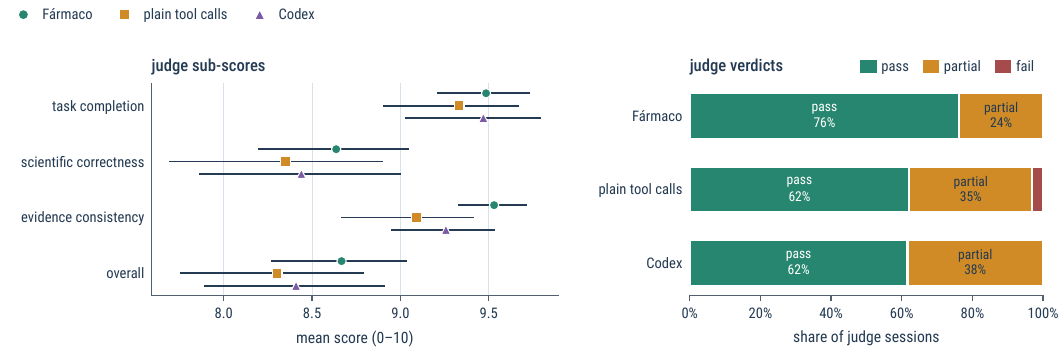}
    \caption{\textbf{Judge sub-scores and verdicts.}
    The left panel shows mean task-completion, scientific-correctness, evidence-consistency, and overall scores, using each run's median over three judge sessions.
    Bars show 95\% confidence intervals from bootstrap resampling of tasks.
    Circles denote F\'armaco, squares denote plain tool calls, and triangles denote Codex.
    The right panel counts 198 judge sessions per configuration, with displayed percentages rounded to whole numbers.
    Pass denotes work accepted as delivered, partial denotes useful but incomplete or flawed work, and fail denotes an unsatisfactory result.
    The judge assigned the overall score directly from its assessment of the complete delivered result.}
    \label{fig:si-benchmark-judge}
\end{figure}

\begin{figure}
    \centering
    \includegraphics[width=\linewidth]{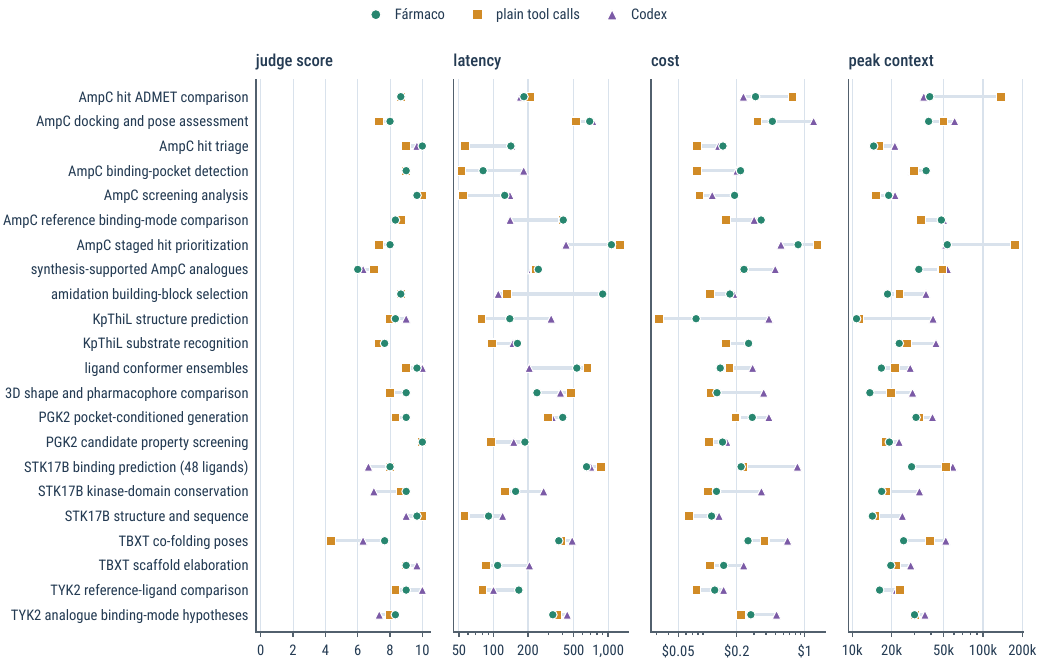}
    \caption{\textbf{Per-task comparison of quality, latency, cost, and peak context.}
    Each row is one of the 22 scientific tasks.
    Markers show the mean overall judge score or median resource metric over the runs of a task and configuration; grey lines span the configurations' values.
    Circles denote F\'armaco, squares denote plain tool calls, and triangles denote Codex; overlapping markers are drawn in that front-to-back order.
    Every task--configuration pair has three runs.
    The latency, cost, and peak-context axes are logarithmic; cost is in US dollars per run and peak context is in input tokens, with k denoting thousands.}
    \label{fig:si-benchmark-per-task}
\end{figure}

\begin{figure}
    \centering
    \includegraphics[width=\linewidth]{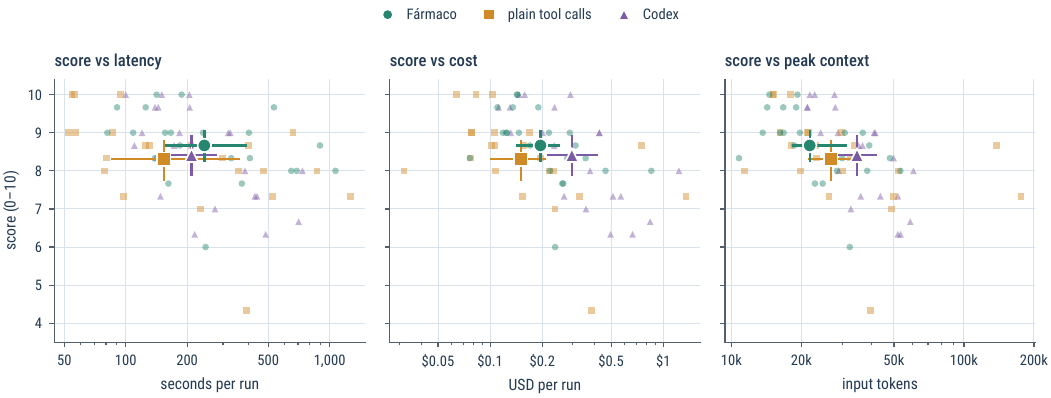}
    \caption{\textbf{Overall judge score versus latency, model cost, and peak context.}
    Small markers show the mean overall score and median resource metric for one task and configuration.
    Large markers show the configuration-wide mean score and median resource metric; horizontal and vertical lines show their respective 95\% task-bootstrap confidence intervals.
    Circles denote F\'armaco, squares denote plain tool calls, and triangles denote Codex.
    Resource axes are logarithmic; peak context counts the largest model input including cached input.}
    \label{fig:si-benchmark-tradeoff}
\end{figure}

F\'armaco had the smallest median peak context, 21,718 input tokens, compared with 26,731 for plain tool calls and 34,674 for Codex.
Median cumulative token use was 176,523, 234,014, and 566,426, respectively.
Median recorded model cost was \$0.195, \$0.150, and \$0.296 per run; search-adjusted costs were \$0.149, \$0.142, and \$0.296.
Plain tool calls had the lowest median latency (154~s) and model cost.
\Cref{fig:si-benchmark-per-task} shows variation between scientific tasks, while \cref{fig:si-benchmark-judge,fig:si-benchmark-tradeoff} separate the judge components and the score--resource comparisons.

\begin{table}
    \centering
    \small
    \caption{\textbf{Aggregate benchmark results.}
    Scores are means of per-run median overall scores, with 95\% confidence intervals from task resampling.
    Pass counts use judge sessions; all other resource summaries are medians over agent runs.
    Costs cover the agent's model calls in US dollars (USD); search-adjusted cost subtracts the domain agents' search-model charges.
    Peak context counts input tokens in the largest model call, including cached input; cumulative tokens sum input and output over calls.}
    \label{tab:si-benchmark-summary}
    \begin{tabular}{@{}p{0.40\linewidth}rrr@{}}
        \toprule
        & \textbf{\farm{}} & \textbf{Plain tool calls} & \textbf{Codex} \\
        \midrule
        Agent runs & 66 & 66 & 66 \\
        Mean overall score & 8.67 & 8.30 & 8.41 \\
        95\% confidence interval & [8.27, 9.03] & [7.76, 8.79] & [7.89, 8.91] \\
        Pass judge sessions & 151/198 & 123/198 & 122/198 \\
        Median latency (s) & 243 & 154 & 210 \\
        Median model cost (USD) & 0.195 & 0.150 & 0.296 \\
        Median search-adjusted cost (USD) & 0.149 & 0.142 & 0.296 \\
        Median cumulative tokens & 176,523 & 234,014 & 566,426 \\
        Median cached input share (\%) & 82.2 & 90.6 & 91.8 \\
        Median peak context (input tokens) & 21,718 & 26,731 & 34,674 \\
        \bottomrule
    \end{tabular}
\end{table}

\subsection{Task-level results and judge detail}

\clearpage
\subsection{Worked example: \acs*{3d} shape and pharmacophore comparison}
\label{sec:si-shape}

The \texttt{molecular\_shape\_feature\_compare} task compares 49 compounds with aspirin in a supplied 50-compound library.
Each configuration generated conformers, calculated shape and pharmacophore overlay scores, and returned a ranked table.
This example uses the same nine runs included in the multi-task benchmark.
\Cref{tab:shape} summarizes their execution accounting and judge scores.

\subsubsection{Execution and resource use}

In each database-backed run, the agent used two subset-scoped workflows for conformer generation and reference-based comparison.
The workflows persisted structures and metrics, and the agent assembled the ranking through \ac{sql} queries.
The requested conformer sampling was one conformer per compound in one run and ensembles of up to eight or ten conformers in the other two.
The plain-tool-call configuration performed batch RDKit calculations in its Python tool; one run also used individual conformer-generation and comparison tools for codeine.
Codex wrote and executed RDKit scripts in the provisioned software environment.

The database-backed configuration had the lowest mean latency and cumulative token use for this task.
Its mean model cost was \$0.129, compared with \$0.136 for plain tool calls and \$0.418 for Codex.
The mean cached share of input tokens was 73.9\%, 93.2\%, and 95.4\%, respectively.
These costs include the model calls made inside the database workflows.

\begin{table}
    \centering
    \small
    \caption{\textbf{Execution efficiency and result quality for the shape and pharmacophore benchmark task.}
    Values are means $\pm$ sample standard deviations over the three benchmark runs per configuration.
    Token counts include cached input, and model costs cover all recorded model calls in US dollars (USD).
    Each run's overall score is the median of its three judge sessions; pass counts use all nine sessions per configuration.}
    \label{tab:shape}
    \begin{tabular}{lrrr}
        \toprule
        & \textbf{\farm{}} & \textbf{Plain tool calls} & \textbf{Codex} \\
        \midrule
        Latency (s) & $218 \pm 71$ & $483 \pm 211$ & $380 \pm 14$ \\
        Model calls & $24.0 \pm 0.0$ & $14.3 \pm 4.9$ & $49.0 \pm 11.1$ \\
        Input tokens ($\times 10^3$) & $113 \pm 10$ & $221 \pm 94$ & $1{,}107 \pm 331$ \\
        Output tokens ($\times 10^3$) & $5.4 \pm 0.4$ & $6.4 \pm 2.7$ & $11.3 \pm 1.6$ \\
        Total tokens ($\times 10^3$) & $119 \pm 9$ & $227 \pm 97$ & $1{,}118 \pm 333$ \\
        Cached share of input (\%) & $73.9 \pm 4.8$ & $93.2 \pm 0.8$ & $95.4 \pm 2.2$ \\
        Model cost (USD) & $0.129 \pm 0.007$ & $0.136 \pm 0.061$ & $0.418 \pm 0.075$ \\
        Overall judge score & $9.0 \pm 0.0$ & $8.0 \pm 0.0$ & $8.0 \pm 0.0$ \\
        Pass judge sessions & 9/9 & 7/9 & 4/9 \\
        \bottomrule
    \end{tabular}
\end{table}

\clearpage
\subsubsection{Ranking agreement and conformer assessment}

All nine final responses contain complete rankings of the 49 candidate compounds with component and combined scores.
The database-backed and plain-tool-call runs combined shape and color scores by addition.
Codex used their mean in two runs and their sum in the third.
We compared the delivered rank orders with the independently recomputed RDKit reference ranking, which samples up to 30 conformers per compound against ten aspirin conformers.
\Cref{tab:si-shape} reports rank correlation and top-10 overlap for the current benchmark runs.

Mean Spearman rank correlations with the reference were $0.941 \pm 0.020$, $0.946 \pm 0.002$, and $0.906 \pm 0.123$ for F\'armaco, plain tool calls, and Codex.
The database-backed runs placed metronidazole first, while plain tool calls and Codex placed salicylic acid first, matching the reference's top-ranked compound.
The reference measures agreement with a more extensively sampled overlay ranking.
The judges identified incomplete conformer-generation records in the database workflows, incomplete method records in plain-tool-call runs, and strained codeine geometries in several plain-tool-call and Codex results.
The judges also identified inverted specified stereocenters in the codeine conformer delivered by one Codex run.

\begin{table}
    \centering
    \small
    \caption{\textbf{Agreement of the benchmark shape rankings with the independently sampled reference.}
    Correlation and overlap are means $\pm$ sample standard deviations over three runs per configuration.
    Top-10 overlap counts shared compounds, and top-rank agreement counts runs that select the reference's first-ranked compound.}
    \label{tab:si-shape}
    \begin{tabular}{lrrr}
        \toprule
        & \textbf{\farm{}} & \textbf{Plain tool calls} & \textbf{Codex} \\
        \midrule
        Compounds ranked per run & 49 & 49 & 49 \\
        Spearman rank correlation & $0.941 \pm 0.020$ & $0.946 \pm 0.002$ & $0.906 \pm 0.123$ \\
        Top-10 overlap & $7.7 \pm 0.6$ & $7.7 \pm 0.6$ & $8.0 \pm 1.0$ \\
        Top-rank agreement & 0/3 & 3/3 & 3/3 \\
        \bottomrule
    \end{tabular}
\end{table}

\clearpage
\subsection{Additional STK17B ranking experiment at larger scale}
\label{sec:si-stk17b}
\label{subsec:stk17b}

\begin{figure}
      \centering
      \includegraphics[width=0.7\linewidth]{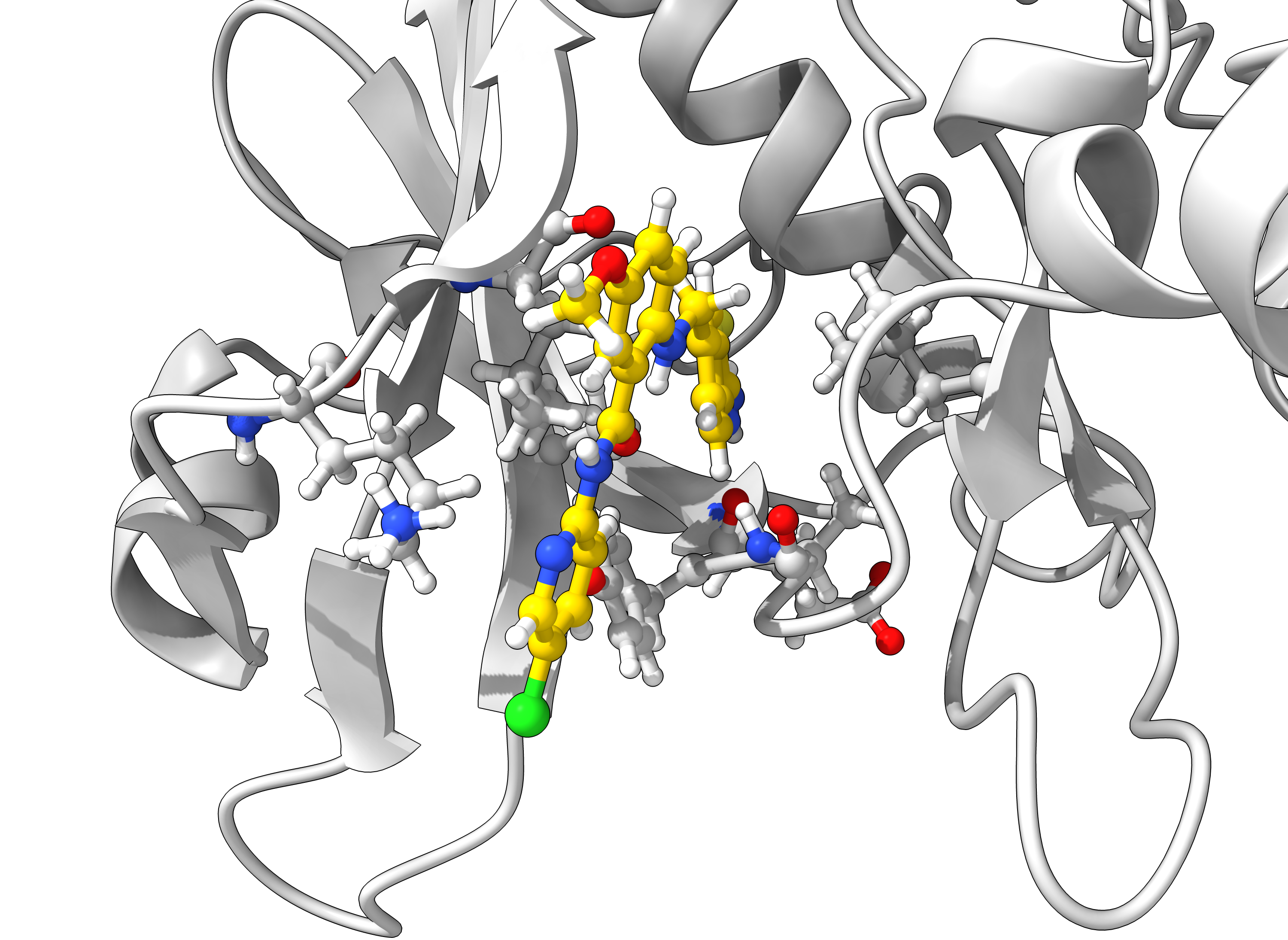}
      \caption{
            \textbf{Predicted binding pose of the top-ranked STK17B ligand.}
            Ligand carbons are shown in yellow, protein carbons and cartoon in grey, and heteroatoms by element.
            Selected interacting residues are shown in ball-and-stick representation.
            Visualization prepared with UCSF ChimeraX \cite{pettersenUCSFChimeraXStructure2021}.
      }
      \label{fig:stk17b_pose}
\end{figure}

This additional experiment extends the 48-ligand STK17B ranking task in the multi-task benchmark (\cref{sec:benchmark}) to the full 256-ligand input library.
\farm{} was instructed to load a \ac{csv} file of 256 reductive-amination ligands, predict their binding to STK17B, also known as DRAK2\footnote{DAP kinase-related apoptosis-inducing protein kinase 2.}, using \ac{pdb} entry 6ZJF with Boltz-2 \cite{passaroBoltz2AccurateEfficient2025}, return a ranked table with binding metrics, and profile the top five predicted poses with \ac{plip}, paying particular attention to hinge-region backbone hydrogen bonds.
In the database-backed run, all five top-ranked poses showed a predicted backbone hydrogen bond to the hinge residue ALA111.
The highest-ranked pose also showed a contact with GLU109.
\Cref{fig:stk17b_pose} shows this predicted complex.

We used this library-scale task to compare agent execution across configurations, focusing on tool calls, token consumption, \ac{llm} cost, and operator intervention.
The same initial instruction was given to the plain-tool-call configuration, in which the \farm{} database is disabled and the underlying services are exposed as plain per-ligand tool calls.
Both configurations used \texttt{gpt-5.5}, completed all 256 predictions, returned a ranking, and performed the top-five interaction analysis.
With the database, the agent imported the ligands, resolved the target, and submitted all 256 protein--ligand pairs through one subset-scoped workflow call.
The workflow persisted the results, metrics, and predicted complexes without returning individual prediction payloads to the model's context.
The agent then queried the results, assembled the ranking, and completed the interaction analysis within a single user turn.
The plain-tool-call configuration issued 256 individual prediction calls and required six user turns.
It stopped before completing the library three times, completed the predictions on the fourth turn, and required two further prompts to acknowledge and then complete the outstanding ranking and interaction analysis.
Both configurations cached the \ac{msa} and issued one alignment search.

The resource accounting in \cref{tab:stk17b} shows that the principal overhead in the plain-tool-call configuration was input processing.
Despite making fewer \ac{llm} calls, the plain-tool-call configuration repeatedly processed a growing conversation containing individual prediction payloads, whereas the database-backed workflow persisted results for selective retrieval.
The resulting increase in cumulative input tokens substantially exceeded the increase in output tokens, indicating that the additional usage primarily reflected processing conversation history rather than generating responses.
Prompt caching moderated the associated cost, but did not eliminate the difference between configurations.
Fewer \ac{llm} calls therefore did not translate into lower token usage or cost.
The agent selected different Boltz-2 sampling settings across configurations, so the comparison concerns agent execution overhead rather than prediction quality.

\begin{table}
      \centering
      \small
      \caption{\textbf{Agent execution for the 256-ligand STK17B task.}
            Both configurations received the same initial instruction and completed the predictions, ranking, and top-five interaction analysis.
            User turns include the initial instruction and follow-up prompts.
            Token counts are cumulative across \acs*{llm} calls; cached input tokens are included in input and total tokens.
            Costs are recorded \acs*{llm} costs in USD and exclude scientific service costs.
            Boltz-2 sampling settings differed between configurations, as described in the text.}
      \label{tab:stk17b}
      \begin{tabular}{lrr}
            \toprule
                                                & \textbf{\farm{} database} & \textbf{Plain tool calls} \\
            \midrule
            User turns                          & 1                         & 6                         \\
            \ac{llm} calls                      & 51                        & 40                        \\
            Tool calls                          & 29                        & 284                       \\
            Input tokens ($\times 10^3$)        & 447                       & 4,376                     \\
            Cached input tokens ($\times 10^3$) & 351                       & 3,732                     \\
            Output tokens ($\times 10^3$)       & 20                        & 51                        \\
            Total tokens ($\times 10^3$)        & 467                       & 4,427                     \\
            \ac{llm} cost (USD)                 & 1.23                      & 6.61                      \\
            \bottomrule
      \end{tabular}
\end{table}

\end{document}